\newif\ifisTR

\isTRtrue

\documentclass{article} 

\usepackage{fullpage}

\usepackage{microtype}
\usepackage{graphicx}
\usepackage{booktabs} 
\usepackage[dvipsnames]{xcolor}

\definecolor{mydarkblue}{rgb}{0,0.08,0.45}
\usepackage[colorlinks,citecolor=mydarkblue,urlcolor=mydarkblue,linkcolor=mydarkblue]{hyperref}

\usepackage{placeins}
\usepackage{algorithm}
\usepackage{algorithmic}
\usepackage{subcaption}
\usepackage{tcolorbox}
\usepackage{nicefrac}  
\usepackage{enumitem}
\usepackage{hyperref}
\usepackage{multirow}
\usepackage{colortbl}
\usepackage{xspace}
\usepackage{wrapfig}

\usepackage{amsmath,amsfonts,bm}

\def\eqref#1{equation~(\ref{#1})}

\def\1{\bm{1}}

\def\vtheta{{\bm{\theta}}}

\DeclareMathAlphabet{\mathsfit}{\encodingdefault}{\sfdefault}{m}{sl}
\SetMathAlphabet{\mathsfit}{bold}{\encodingdefault}{\sfdefault}{bx}{n}

\DeclareMathOperator*{\argmin}{arg\,min}

\usepackage{amsmath}
\usepackage{amssymb}
\usepackage{mathtools}
\usepackage[export]{adjustbox}
\usepackage{amsthm}
\usepackage{xcolor}
\usepackage{tikz}

\usepackage[capitalize,noabbrev]{cleveref}

\theoremstyle{plain}

\theoremstyle{definition}

\theoremstyle{remark}

\definecolor{mygray}{gray}{0.85}
\definecolor{LightBlue}{cmyk}{0.06, 0.03, 0.01, 0.0}

\newcommand\gb{ \rowcolor{gray!15}}

\usepackage[textsize=tiny]{todonotes}

\usepackage[round,semicolon,sort]{natbib}

\newcommand{\ourmethod}{\texttt{MD tree}\xspace}

\renewcommand{\cite}[1]{\citep{#1}}

\begin{document}

\title{MD tree: a model-diagnostic tree grown on loss landscape}
\date{}

\author{%
  Yefan Zhou$^{1}\footnote{First two authors contributed equally.}$, 
  Jianlong Chen$^{2}\footnotemark[1]$, 
  Qinxue Cao$^{3}$, 
  Konstantin Schürholt$^{4}$,   
  Yaoqing Yang$^{1}$ \\
  $^1$ Dartmouth College\\
  $^2$ Zhejiang University\\
  $^3$ University of Illinois Urbana-Champaign \\
  $^4$ University of St. Gallen
}

\maketitle

\begin{abstract}
This paper considers ``model diagnosis'', which we formulate as a classification problem. 
Given a pre-trained neural network~(NN), the goal is to predict the source of failure from a set of failure modes (such as a wrong hyperparameter, inadequate model size, and insufficient data) without knowing the training configuration of the pre-trained NN.
The conventional diagnosis approach uses training and validation errors to determine whether the model is underfitting or overfitting.
However, we show that rich information about NN performance is encoded in the optimization loss landscape, which provides more actionable insights than validation-based measurements.
Therefore, we propose a diagnosis method called \ourmethod based on loss landscape metrics and experimentally demonstrate its advantage over classical validation-based approaches.
We verify the effectiveness of \ourmethod in multiple practical scenarios: (1) use several models trained on one dataset to diagnose a model trained on another dataset, essentially a few-shot dataset transfer problem;
(2) use small models (or models trained with small data) to diagnose big models (or models trained with big data), essentially a scale transfer problem.
In a dataset transfer task, \ourmethod achieves an accuracy of 87.7\%, outperforming validation-based approaches by 14.88\%. 
Our code is available at \href{https://github.com/YefanZhou/ModelDiagnosis}{https://github.com/YefanZhou/ModelDiagnosis}.
\looseness-1
\end{abstract}

\vspace{-1mm}
\section{Introduction}

There is a notable gap in the literature in systematically diagnosing the reason for the underperformance of a pre-trained neural network (NN).
For example, it is often difficult to access proprietary datasets, detailed configuration parameters, and complete training methodologies to develop these models.
When a trained model underperforms, the conventional approach to troubleshooting through retraining with alternative settings thus becomes impractical.
Therefore, we should develop diagnostic methods that do not involve complete datasets, retraining processes, and explicit configuration details. Can we say anything about why the model is underperforming in this setting? 
Answering this question requires a comprehensive analysis of the individual-trained model to uncover the root causes of failure and identify targeted strategies for improvement. The critical question in this context is the following:

\begin{figure}[!t]
    \centering
    \includegraphics[width=0.8\linewidth]{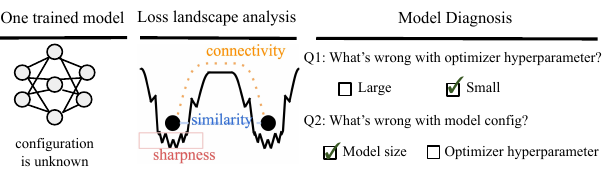} 
    \caption{\textbf{(Overview of our model-diagnostic framework using \ourmethod).} 
    This framework is designed to analyze and diagnose NNs where the training configuration is unknown.
    By examining the loss landscape structure of a given trained model, \ourmethod can identify potential failure sources of suboptimal performance.}~\label{fig:system-design} \vspace{-8mm}
\end{figure} 
\vspace{2mm}
\emph{
How can we methodically diagnose the root causes of the underperformance of a model without the need for detailed configuration specifics, retraining processes, and extensive reliance on the dataset?\looseness-1
}

\paragraph{Canonical Problems in Model Diagnosis.} In this paper, we define the problem of ``model diagnosis'' and present a systematic framework for evaluating and approaching it. 
The goal is to identify the causes of underperformance in trained models without relying on configuration specifics, retraining processes, or extensive use of validation datasets.
See Figure~\ref{fig:system-design} for an overview of the framework.
We categorize the sources that lead to model failure into four:
(1) small optimizer hyperparameter (e.g., small batch size); (2) large optimizer hyperparameter; (3) inadequate model size; and (4) insufficient data amount.
The four sources of failure are inspired by the \emph{statistical physics} viewpoint of learning~\cite{seung1992statistical,WRB93,DKST96,EB01_BOOK,martin2017rethinking, yang2021taxonomizing, zhou2023three}.
According to prior works in this area~\cite{martin2017rethinking,yang2021taxonomizing}, the most important optimizer hyperparameters are the \emph{temperature-like parameters} that characterize the magnitude of noise introduced in stochastic training.
These can be as common as the learning rate and batch size.
Similarly, these works use \emph{load-like parameters} to refer to the size of the data relative to the size of the model. They use temperature-like and load-like parameters together to model the peculiar generalization properties of deep learning, an approach known as the \emph{very simple deep-learning model}~\cite{martin2017rethinking,yang2021taxonomizing}.
Therefore, our framework considers temperature-like hyperparameters (mainly batch size in this paper) and load-like parameters (data amount and model size)~\footnote{Note that large data/model size can also hurt~\cite{nakkiran2021deep} due to double descent~\cite{belkin2019reconciling,bartlett2020benign,hastie2022surprises,mei2022generalization}, but we do not consider large data/model size as primary failure sources here, as appropriate regularization can often mitigate double descent~\cite{nakkiran2020optimal}.}. Controlling these parameters is also considered important in recent large-scale studies, such as the neural scaling law~\cite{kaplan2020scaling}.

More concretely, we decompose the four sources into two canonical binary classification problems. That is, given a trained model, without knowing the training configurations, \ourmethod aims to answer the following questions:

\begin{itemize}[noitemsep,topsep=0pt, after=,before=]
    \item[Q1] Can we determine whether the optimizer hyperparameter (e.g. batch size) used to train the model is large or small, compared to the optimal hyperparameter?
    \item[Q2] Can we determine the more severe failure source between inadequate model size and wrong optimizer hyperparameter? Which one should be addressed first?
\end{itemize}

In Section~\ref{sec:related_work}, we discuss why our approach to defining and addressing model diagnosis is orthogonal or complementary to previous work.

\paragraph{Good Features for Model Diagnosis.}
The key to answering these questions is to extract informative features from trained models associated with different failure sources.
Conventional metrics, as discussed in \citet{raschka2018model},
typically diagnose a trained model by analyzing training and validation error (loss), which can be used to (roughly) determine whether the model is overfitting or underfitting.
However, if we want a fine-grained model diagnosis, such as answering questions like Q1 and Q2, the validation-based metrics have several limitations.
First, we show that validation-based metrics provide limited information on the root causes of underperformance.
For example, while large training errors may indicate underfitting, they do not distinguish between insufficient model size and suboptimal optimizer hyperparameters.
Second, a detailed visualization of the model diagnosis shows that validation-based metrics exhibit complex relationships with failure sources, resulting in complicated nonlinear classification boundaries in the binary classification problems Q1 and Q2.
These boundaries not only lead to inaccurate diagnoses but also transfer poorly across different groups of models.

Again, inspired by the statistical physics viewpoint of learning, this research uses metrics from the NN loss landscape to extract useful features for model diagnosis.
Prior works~\cite{yang2021taxonomizing,zhou2023three} identify several useful loss landscape metrics, including sharpness~\cite{yao2020pyhessian}, connectivity~\cite{garipov2018loss, nguyen2020wide} and similarity~\cite{kornblith2019similarity}, in determining the distinct \emph{regimes} to characterize the test accuracy of NNs, where each regime represents a range of configurations where the loss landscape properties remain homogeneous.
Our work, however, takes one step further and uses these features for model diagnosis, aiming to answer fine-grained questions such as Q1 and Q2, as shown in Figure~\ref{fig:system-design}.\looseness-1

\paragraph{A Simple Model for Model Diagnosis.} 
Our framework employs an interpretable tree-based method called \ourmethod to address Q1 and Q2. This method uses a divide-and-conquer way to partition the NN configuration space into distinct regimes using loss landscape metrics.
Each identified regime is then associated with a particular source of failure, allowing for interpretable diagnostic predictions.
Unlike traditional decision trees (DTs) that rely on information gain for branching, \ourmethod adopts a hierarchy inspired by the statistical physics viewpoint. That is, we prioritize the metrics shown in previous work to capture the sharp transitions of the NN regimes by placing these metrics at shallower levels of the tree.

We evaluate \ourmethod in few-shot learning scenarios. Few-shot means that a few trained NNs are used as a ``training set'' to develop a function that maps loss landscape metrics to diagnostic predictions.
We evaluate the diagnostic ability of this classifier under two transfer learning scenarios: \emph{dataset transfer}, where the NNs to be diagnosed are models trained on different datasets, possibly with label noise, and \emph{scale transfer}, where the training involves diagnosing small models, while the testing involves diagnosing large ones.\looseness-1

\paragraph{Main Findings.}
Our empirical analysis uses datasets of pre-trained models with 1690 different configurations, including different model sizes, data amounts, and optimization hyperparameters. 
Our analysis shows that \ourmethod, which uses loss landscape metrics, can effectively diagnose the sources of model failures and significantly outperform the validation-based method.
Our key observations to support this main claim include: 
\begin{itemize}[noitemsep,topsep=0pt,after=,before=]
    \item[\textbf{O1}] \ourmethod uses loss landscape metrics to distinguish the ``regime'' of the models: the models within the same regime are homogeneous and share the same failure source (e.g., the failure is caused by a small optimizer hyperparameter value). 
    Also, most of the boundaries between the regimes are sharp, so the models with different failure sources are linearly separable by applying simple thresholding to the loss landscape metrics. \looseness-1 
    \item[\textbf{O2}] The boundaries determined by \ourmethod have high transferability across different groups of models (e.g., transferring from models trained on clean data to those with label noise).
    For example, \ourmethod trained with small-scale models (e.g., models with 0.01M parameters) can be used to diagnose failures in models with up to 44.66M parameters and can achieve an accuracy of 82.56\%.
    Therefore, \ourmethod's decision thresholds determined by a few known sample models can be effectively transferred to diagnose other unseen models.
    \item[\textbf{O3}] The tree hierarchy of \ourmethod, inspired by recent studies from \citet{yang2021taxonomizing}, is beneficial to our model diagnosis problem. The effectiveness of the hierarchy is demonstrated by \ourmethod's better performance compared to standard DTs using the same set of loss landscape metrics. Specifically, in a few-shot setting with 12 training samples, \ourmethod outperforms standard DT by 12.71\%.
    \looseness-1
\end{itemize}

In summary, we introduce \ourmethod, 
a useful model diagnostic tool that uses previous research on loss landscape metrics to provide actionable insights on model failures.
Our key contributions are as follows:
\begin{itemize}[noitemsep,topsep=0pt,leftmargin=*,after=,before=]
    \item We introduce a model diagnosis method called \ourmethod, which can identify the specific failure source within ML pipelines (optimizer, model size, data amount) that affects a model. 
    In cases where multiple failure sources are at play, \ourmethod can determine which failure source is most detrimental to performance.\looseness-1
    \item We show that \ourmethod can be applied to a range of practical transfer learning cases, including 
    (i) diagnosing models trained with noisy data given models trained with clean data;
    (ii) diagnosing models trained with large-scale parameter/data given trials on small-scale models; 
    (iii) diagnosing models evaluated on out-of-distribution (OOD) test data~\cite{hendrycks2018benchmarking} or trained on class-imbalanced training data~\cite{cui2019class} given normal models; (iv) diagnosing Transformer models given ResNet models.
    \item Moreover, \ourmethod outperforms validation-based methods in diagnostic accuracy by a significant margin of 14.88\%. 
    In Appendix~\ref{sec:one-step-change}, we apply \ourmethod to a novel task of determining a ``one-step configuration change'' to improve test performance. We show that \ourmethod can lead to superior CIFAR-10 test accuracy improvement compared to validation-based methods.\looseness-1
\end{itemize}
\section{Related Work}\label{sec:related_work} 

\subsection{Model Diagnosis}\label{sec:related_work_modeldiag}
Model diagnostics are widely studied for linear regression models~\cite{weisberg2005applied,chatterjee2009sensitivity}. These approaches typically validate several assumptions, such as the normality of the residuals, the linearity of the relationship between the explanatory and response variables, and the presence of outliers or influential data points. These diagnostics provide ``actionable'' insights and allow us to choose to remove or handle outlier points, transform features, or include or remove features to better explain the data. However, for NNs, applying linear regression diagnostics is not straightforward. Nonetheless, model diagnostics for deep NNs is a growing field with ongoing efforts to understand model failures and behaviors. This section discusses these efforts and shows that our unique perspective on model diagnosis is orthogonal to previous studies.

The work most closely related to ours in the literature uses theoretically principled approaches to measure NNs. One such approach is \emph{Heavy-tailed Self-regularization (HT-SR) theory}~\cite{martin2017rethinking, martin2018implicit_JMLRversion, martin2020predicting_NatComm, MM21a_simpsons_TR, yang2022evaluating,zhou2024temperature}, which analyzes weight matrices to predict and explain model performance. This line of work also connects to \emph{generalization metrics}~\cite{McAllester1999PACBayesianMA,keskar2017large,bartlett2017spectrally,jiang2019fantastic, dziugaite2020search, baek2022agreement, jiang2022assessing, kim2023fantastic, pmlr-v202-andriushchenko23a}.
Previous research on generalization metrics has primarily focused on predicting test performance trends and designing regularizers during training. In contrast, our study develops metrics to diagnose sources of model failure and predict optimal ways to improve model configuration.\looseness-1

Recent research on LLMs often uses the \emph{self-diagnosis} ability to evaluate output quality~\cite{schick2021self, chen2023exploring, fu2023gptscore, ji2023exploring, zheng2024judging}.
Although these methods provide useful information, many lack transparency, making their diagnostic processes difficult to interpret. In contrast, our work uses theoretically motivated metrics and easy-to-interpret tree models for diagnostics.
Research in \emph{mechanistic interpretability} aims to understand the learned circuits in models \cite{wang2023interpretability, conmy2023towards, chughtai2023toy} and to probe the learned features or concepts \cite{gurnee2023finding}. 
Other studies \cite{ilyas2022datamodels} focus on \emph{data attribution}, attributing model behavior to training samples, and extending this approach to large-scale models \cite{park2023trak} and analyzing training algorithms \cite{shah2023modeldiff}. 
Instead of concentrating on specific model components (e.g., mechanistic interpretability) or individual training data samples (e.g., data attribution), our work aims to diagnose failures arising from generic sources in machine learning pipelines, such as training hyperparameters, data quantity, and model size.
While \emph{model editing}~\cite{decao2021editing, mitchell2022fast, meng2022locating} is also related to diagnosis, it primarily focuses on downstream control and knowledge editing, which is beyond the scope of our diagnostic approach. \looseness -1

\subsection{Learning Curve Prediction} \label{sec:related_work_lc}
Learning curves, as seen in \citet{banko-brill-2001-scaling, hestness2017deep, sun2017revisiting, DBLPcurve, Viering2022}, measure a model’s generalization performance relative to training data size. Recent research has focused on neural scaling laws (NSLs)~\citep{kaplan2020scaling, alabdulmohsin2022revisiting, hoffmann2022training, caballero2023broken, muennighoff2024scaling}, particularly in LLMs. 
These laws indicate that language model performance scales with model size, dataset size, and training computation in a power-law manner.
However, NSLs are typically used to predict optimal resource allocation before or during the early stages of training, assuming known training configurations to generate initial points on the curves. 
In contrast, our diagnosis framework focuses on post-training analysis, where a single pre-trained model is available, and the exact training configuration is unknown.
Our paper also focuses on developing transferable metrics that enable robust model diagnostics when hyperparameter tuning on the original training data is not feasible.
Furthermore, the simple power-law relationship in NSLs may overlook complex behaviors that arise from the joint influence of load and temperature parameters. For example, recent work shows that a model's performance exhibits such complex behaviors when both load and temperature change simultaneously, potentially leading to multi-regime patterns~\cite{yang2021taxonomizing, zhou2023three} and double descent~\cite{bartlett2020benign, nakkiran2021deep, mei2022generalization, hastie2022surprises, caballero2023broken}, beyond simple power-law-like patterns. Our work incorporates load and temperature parameters to handle these multi-regime patterns effectively.

\subsection{Statistical Physics of Learning}
Our framework builds on the statistical physics of learning~\cite{seung1992statistical, WRB93, engel2001statistical, zdeborova2016statistical, bahri2020statistical} and considers factors such as data amount and model size. From the perspective of statistical physics, load-like parameters characterize the quantity and/or quality of data relative to model size. Temperature-like parameters, on the other hand, characterize the magnitude of noise introduced during stochastic training. Several recent papers~\cite{martin2017rethinking, yang2021taxonomizing, zhou2023three} use load and temperature parameters to analyze multi-regime patterns in NNs.

\section{Preliminaries and Background}\label{sec:preli}
\paragraph{NN Training.}
We consider training a NN $f$, with trainable parameters
$\vtheta \in \mathbb{R}^p $, on a training dataset comprising $n$ datapoint/label pairs, using an optimizer parameterized by a hyperparameter $t$.
This paper mostly considers \emph{temperature-like} hyperparameters $t$, such as the learning rate or batch size.
We denote the error of NNs (evaluated on a particular dataset) by \( \mathcal{E} \) and the loss by \( \mathcal{L} \). The error evaluated using the training and validation sets is thus denoted as \( \mathcal{E}_\text{tr} \) and \( \mathcal{E}_\text{val} \), respectively, while the loss is denoted by \( \mathcal{L}_\text{tr} \) and \( \mathcal{L}_\text{val} \).\looseness-1

\paragraph{Loss Landscape Metrics.}
Three types of metrics were introduced in~\citet{yang2021taxonomizing} to quantify the local and global geometric structures of loss landscapes, and they were used to study ``phase transitions'' in the hyperparameter space.
In particular, these metrics are: ``connectivity'' metrics such as mode connectivity~\citep{garipov2018loss,draxler2018essentially}, ``similarity'' metrics such as Centered Kernel Alignment~\citep{kornblith2019similarity}, and ``local sharpness'' metrics such as the Hessian trace and largest Hessian eigenvalue~\citep{yao2020pyhessian}.
The mode connectivity ($\mathcal{C}$) quantifies how well different local minima are connected to each other in the loss landscape.
The Hessian trace ($\mathcal{H}_t$) and largest Hessian eigenvalue ($\mathcal{H}_e$) contain the curvature information of the loss landscape, which could be used to quantify the local sharpness. Since $\mathcal{H}_t$ and $\mathcal{H}_e$ have been shown to provide similar information in determining the regime~\cite{yang2021taxonomizing}, we mainly use $\mathcal{H}_t$ to measure sharpness.
CKA similarity ($\mathcal{S}$) captures the similarity between the outputs of trained models.
These metrics are precisely defined in the Appendix~\ref{app:loss-land}.\looseness-1

\section{Constructing Model Diagnosis Framework} 
In this section, we build the framework of \ourmethod. Section~\ref{sec:failure_sources} presents definitions of the diagnosis problems. Section~\ref{sec:diag-method} presents \ourmethod method. Section~\ref{sec:few-shot-method} presents different model metrics and the few-shot diagnosis method.
Section~\ref{sec:eval-setup} presents the evaluation setup, including the diagnostic datasets and transfer setups.

\subsection{Defining the Diagnosis Problem} \label{sec:failure_sources}

Given a training algorithm $\mathcal{A}$, a NN $f_0$ trained with a specific setting of $p$ parameters, $n$ data samples, and optimizer hyperparameter $t$ can be represented as $f_0 = \mathcal{A}(p, n, t)$. 
In the context of real-world constraints, we set the maximum allowable values for data amount, model size, and the range for optimizer hyperparameters as $n_{\max}$, $p_{\max}$ and $[t_{\min}, t_{\max}]$. Our goal is to identify the ``failure source'' $m$ of a trained model $f_0$, which is the hyperparameter primarily responsible for its suboptimal performance. The impact of the failure source is measured using a metric we call \emph{room for improvement} ($\text{RFI}$).\looseness-1

\paragraph{Failure Sources.} We consider four types of failure sources:
\begin{itemize}[noitemsep,topsep=0pt,leftmargin=*,after=,before=]
    \item Failure $m^{\downarrow}_{t}$: the optimizer hyperparameter ($t$) is smaller than the optimal choice.
    \item Failure $m^{\uparrow}_{t}$: the optimizer hyperparameter ($t$) is larger than the optimal choice.
    \item Failure $m_{p}$: the number of parameters ($p$) is too small.\looseness-1
    \item Failure $m_{n}$: the amount of data ($n$) is too small.
\end{itemize}
The above taxonomy defines the set of failure sources considered in this paper:
$\mathcal{M} = \{m^{\downarrow}_{t}, m^{\uparrow}_{t}, m_{p},  m_{n}\}$.

\paragraph{Room for Improvement ($\text{RFI}$).} The $\text{RFI}$ of a failure source $m$ is defined as the gap (in validation error) between the current configuration $(p, n, t)$ and the optimal configuration when changing \emph{the single hyperparameter} in $(p, n, t)$ that corresponds to the failure source $m$. For example, the $\text{RFI}$ of failure source $m_{p}$ is defined as
\begin{align}\label{eqn:RFI}
\begin{split}
    \text{RFI}(m_{p}, f_0) &= \mathcal{E}_\text{val}(f_0) - \mathcal{E}_\text{val}(f^*),  \\ \text{where}   \quad
    f^{*} &= \mathcal{A}(q^*, n, t), \, \\  q^* &= \argmin_{q \in [p, p_{\max}]} \mathcal{E}_\text{val}(\mathcal{A}(q, n, t)).
\end{split}
\end{align}
where the last two lines indicate finding the optimal number of model parameters larger than or equal to the current value $p$ but smaller than or equal to the maximum allowable model size $p_{\max}$. The $\text{RFI}$ for other failure sources are defined similarly and presented in Appendix~\ref{app:rfi-def}.\looseness-1 

\paragraph{Canonical Diagnosis Questions.} Using $\text{RFI}$, we define the diagnosis questions Q1 and Q2. 
We define the \emph{RFI gap} as the $\text{RFI}$ difference between two failure sources, denoted as $G$.
For example, the $\text{RFI}$ gap between large and small optimizer hyperparameters is
$G(m^{\uparrow}_{t}, m^{\downarrow}_{t}) = \text{RFI}(m^{\uparrow}_{t}) - \text{RFI}(m^{\downarrow}_{t})$.
Then, Q1 can be defined as the following binary classification problem:
\vspace{-1mm}
\begin{equation}\label{eqn:Q1}
    \text{Q1}: f_0 \rightarrow \{G(m^{\uparrow}_{t}, m^{\downarrow}_{t}) > 0, G(m^{\uparrow}_{t}, m^{\downarrow}_{t}) < 0   \}.
\end{equation}
\vspace{-1mm}
Similarly, Q2 can be defined as
\vspace{-1mm}
\begin{equation}\label{eqn:Q2}
    \text{Q2}: f_0 \rightarrow \{G(m_{p}, m_{t}) > 0, G(m_{p}, m_{t}) < 0 \},
\end{equation}
where $G(m_{p}, m_{t}) = \text{RFI}(m_{p}) - \text{RFI}(m_{t})$ and 
$\text{RFI}(m_{t}) = \text{max}\{\text{RFI}(m^{\uparrow}_{t}), \text{RFI}(m^{\downarrow}_{t}) \}$. We write $G(m, m')$ simply as $G$ when the two failure sources $m, m'$ are clear. In Appendix~\ref{app:rationale-binary-class}, we explain why $G=0$ is not considered as a third class in the problem.

\paragraph{Diagnosis Objective.}  
Given a sample (which is a pre-trained model $f_0$), one can use~\eqref{eqn:RFI} to define the RFI and then define the binary label using~\eqref{eqn:Q1} or~\eqref{eqn:Q2}.
The goal is to learn a function that maps a pre-trained model $f_0$ to the binary label in $\{G(m,m')>0, G(m,m')<0\}$, using only features extracted from $f_0$, without calculating $G(m,m')$, and no retraining is allowed.\looseness-1

\subsection{Defining \ourmethod}\label{sec:diag-method}
\begin{wrapfigure}{r}{0.55\textwidth}
    \centering
    \includegraphics[width=\linewidth]{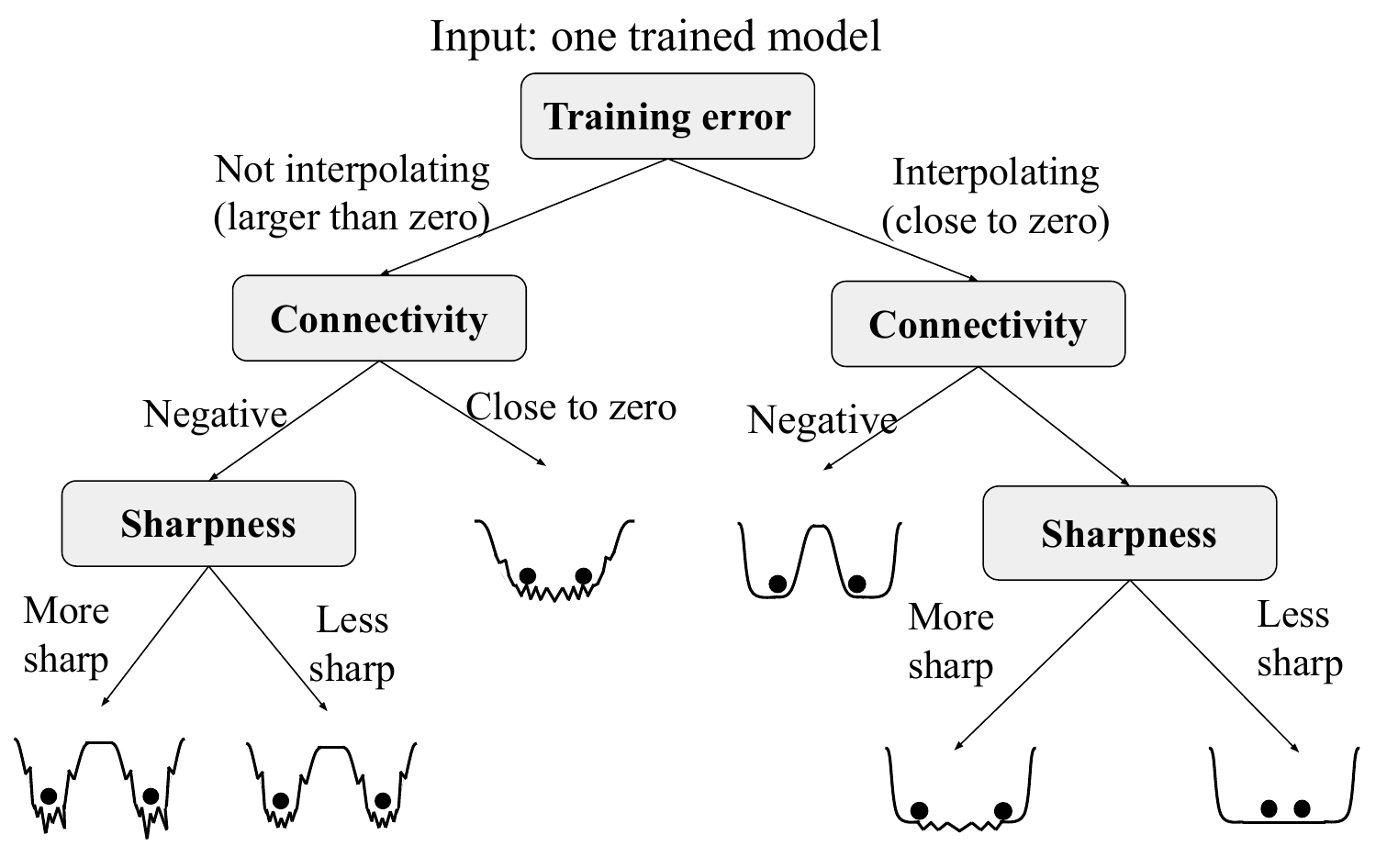}~\vspace{-3mm}
    \caption{\textbf{(MD tree based on loss landscape structure of trained models).} A DT using loss landscape metrics to determine distinct regimes of model configurations such that each regime has the same root cause of failure.
    The tree hierarchy is fixed, while the decision threshold is trained in a few-shot manner.
    Part of the tree hierarchy is selected using ideas from~\citet{yang2021taxonomizing}, which suggests a ``multi-regime'' structure of the hyperparameter space.
    }~\label{fig:md-tree}  \vspace{-6mm}
\end{wrapfigure}

We now introduce \ourmethod, which is a tree-based method to predict whether $G(m,m')$ is greater than 0.
The main idea of \ourmethod is to use a divide-and-conquer way to diagnose failure sources using loss landscape metrics.
See Figure~\ref{fig:md-tree} for the tree structure.
We aim to partition the hyperparameter space into multiple regimes, where the models in the same regime roughly share the same failure source.
\citet{yang2021taxonomizing} use these loss landscape metrics to determine the NN regimes and predict test performance.
Our tree construction borrows the regime partition from this work to grow a partial tree and then completes it using more experimentally useful branch partitions.
The tree is constructed hierarchically, prioritizing metrics that have been shown to lead to sharp transitions in the hyperparameter space, such as training error and mode connectivity. The similarity measure tends to have a smooth transition and is given a lower priority.
Therefore, the tree is constructed in the following order: training error at the root, connectivity at the first depth level, and finally sharpness or similarity.
This order emphasizes the importance of the interpolation threshold~\cite{nakkiran2021deep} and connectivity~\cite{yang2021taxonomizing, zhou2023three} because they can be used to measure sharp and clear transitions in NN behaviors. 
Thus, the tree structure is fixed. \ourmethod only optimizes the thresholds of the metrics at each internal node sequentially from top to bottom. 
Initial values and search ranges are provided for each threshold, and the bounded Brent method \cite{Brent1973} is used to optimize these thresholds to maximize training accuracy.
The hyperparameters are provided in Appendix~\ref{sec:app-diag}.\looseness-1

\subsection{Defining Baseline Methods}\label{sec:few-shot-method}

We introduce baseline diagnosis methods for solving Q1 and Q2, which cover the model metrics that are used as features and the diagnosis functions. 

\paragraph{Model Metrics.}
To predict failure sources based on a trained model $f_0$, we define a feature vector $\mathbf{P}$, which comprises several metrics derived from $f_0$.
We study loss landscape metrics and two conventional baselines:
\begin{itemize}[noitemsep,topsep=0pt,leftmargin=*,after=,before=]
    \item {\bf Loss landscape}: combining the local and global loss landscape metrics defined in Section \ref{sec:preli} with training error. The components of the feature vector are $\mathbf{P} = [ \mathcal{E}_\text{tr}, \mathcal{C}, \mathcal{H}_t, \mathcal{S} ]$. \looseness-1 
    \item {\bf Validation}: training error/loss, and validation error/loss, formally $\mathbf{P} = [\mathcal{E}_\text{val}, \mathcal{E}_\text{tr}, \mathcal{L}_\text{val}, \mathcal{L}_\text{tr}]$. \looseness-1
    \item {\bf Hyperparameter}: the hyperparameters of trained NN that are not related to the diagnosis question. For example, when predicting whether $t$ is large or small ($m^{\uparrow}_{t}$ vs $m^{\downarrow}_{t}$), $t$ is not considered a metric. However, the parameter amount $p$ and the data amount $n$ are considered.
\end{itemize}

\paragraph{Diagnosis Function.}
We adopt a standard DT as a baseline classifier to map the feature vector $\mathbf{P}$ to the binary decision $\{G > 0, G < 0\}$.
Recall that $G$ denotes the RFI gap between two failure sources.
We adopt the standard implementation of DT in~\citet{hastie01statisticallearning}, with details provided in Appendix~\ref{sec:app-diag-hyper}.

\subsection{Experimental Setup}\label{sec:eval-setup} 
\paragraph{Collections of Pre-trained Models.} We introduce the datasets used to evaluate \ourmethod, primarily obtained from the collections of pre-trained models studied by \citet{yang2021taxonomizing}. We release these collections for future research on model diagnosis\footnote{\href{https://github.com/YefanZhou/ModelDiagnosis}{https://github.com/YefanZhou/ModelDiagnosis}}. The collection of models, denoted as $\mathcal{F}$, includes various ResNet models trained on CIFAR-10 with differing number of parameters ($p$), data amounts ($n$), and optimizer hyperparameters ($t$, batch size). This collection comprises a total of 1690 configurations, each with five runs (models) using different random seeds. 
Additionally, we have another set, $\mathcal{F}^{\prime}$, which includes 1690 configurations trained with the same varying parameters but trained with 10\% label noise. For details, see Appendix~\ref{sec:coll-model}. \looseness-1

\paragraph{Few-shot Setup.}
We consider a few practical model diagnostic problems, where the training set for \ourmethod is small (few-shot).
We elaborate on how the corresponding training sets are constructed here, with more details about the training sample labeling provided in Appendix~\ref{sec:case-study}.
\begin{itemize}[noitemsep,topsep=0pt,leftmargin=*,after=,before=]
    \item \textbf{w/ dataset transfer}: For Q1 (which is to determine whether the hyperparameter $t$ is large or small), the training set consists of models randomly sampled from $\mathcal{F}$ for a fixed parameter count and data amount $(p, n)$. For Q2 (determining model size versus optimizer), the training set consists of models randomly sampled from $\mathcal{F}$ for a fixed data amount. For both cases, the test set is $\mathcal{F}^{\prime}$.
    \item \textbf{w/ scale transfer}: 
    The training set includes models trained with data amounts below certain thresholds (e.g., 5K data points) or with parameter counts below specific limits (e.g., 0.04M). The test set is $\mathcal{F}^{\prime}$, the same as dataset transfer. 
    Note that using $\mathcal{F}^{\prime}$ as the test set in the scale transfer study inherently includes dataset transfer.
    This design ensures consistency in our test set $\mathcal{F}^{\prime}$ across the different transfer studies, allowing for comparability of results and ensuring that the test set does not include any training samples from $\mathcal{F}$. \looseness-1
\end{itemize}

In addition to the two primary cases, we consider three additional transfer scenarios. We first examine OOD generalization, where pre-trained models handle distribution shifts during testing. Second, we investigate class-imbalanced training, where pre-trained models are initially trained on long-tailed datasets. Finally, we explore diagnosing unseen model architectures, specifically focusing on Vision Transformers, while the training set includes ResNet models. The details are presented in Appendix~\ref{sec:broader-transfer-sce-setup}.

\section{Is the Optimizer Hyperparameter Large or Small}\label{sec:adjust-temp} 
In this section, we focus on Q1, which is to determine whether an optimizer hyperparameter (e.g., batch size) is too large or too small. This is formulated as a binary classification problem defined in~\eqref{eqn:Q1}. In Section~\ref{sec:q1-few-shot}, we evaluate \ourmethod on the dataset transfer and the scale transfer tasks. In Appendix~\ref{sec:one-step-q1},
we further study \ourmethod on a one-step configuration change task. In Appendix~\ref{sec:broader-transfer-sce-results}, we study three additional transfer scenarios.

\begin{figure}[!th]
    \centering
    \includegraphics[width=0.7\linewidth,keepaspectratio]{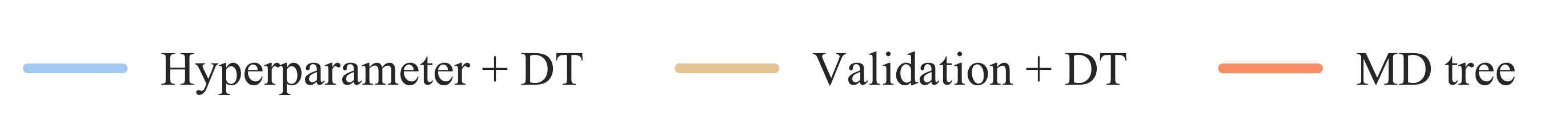} \\
    \begin{subfigure}{0.33\linewidth}
    \includegraphics[width=\linewidth,keepaspectratio]{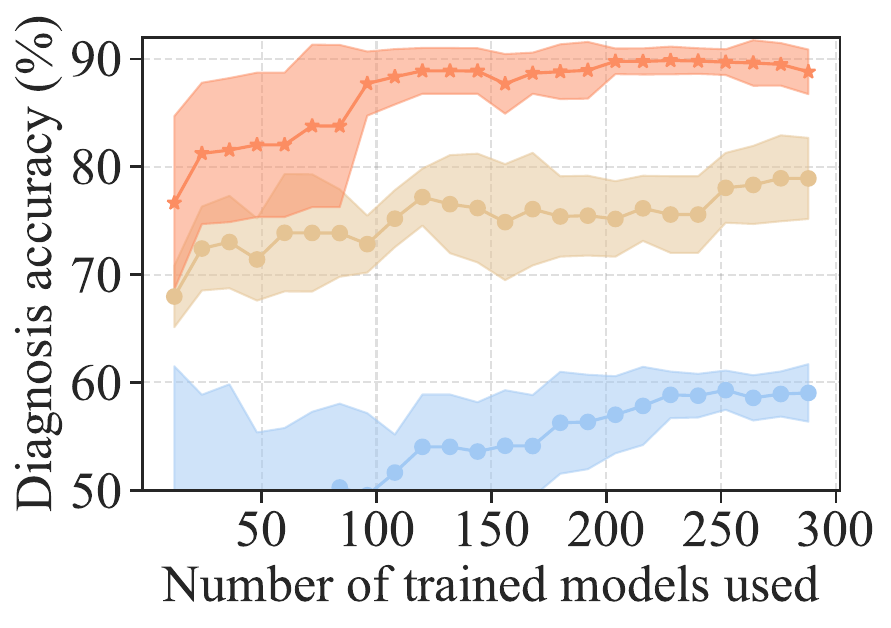} \hspace{2mm}
    \caption{Dataset transfer}~\label{fig:temp-fail-random}
    \end{subfigure} 
    \begin{subfigure}{0.32\linewidth}
        \includegraphics[width=\linewidth]{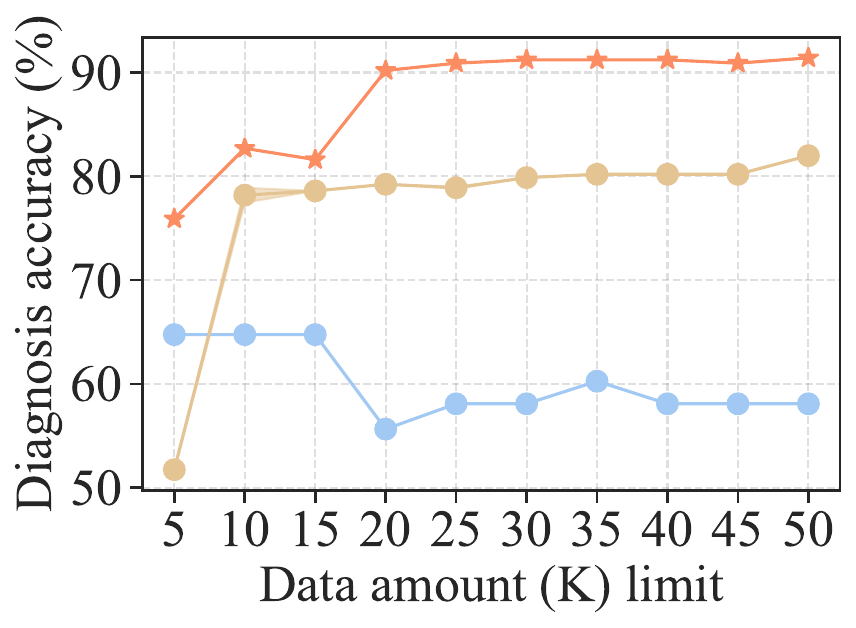}
        \caption{Scale transfer: limiting model's training data amount}~\label{fig:temp-fail-data}
    \end{subfigure} 
    \begin{subfigure}{0.325\linewidth}
        \includegraphics[width=\linewidth]{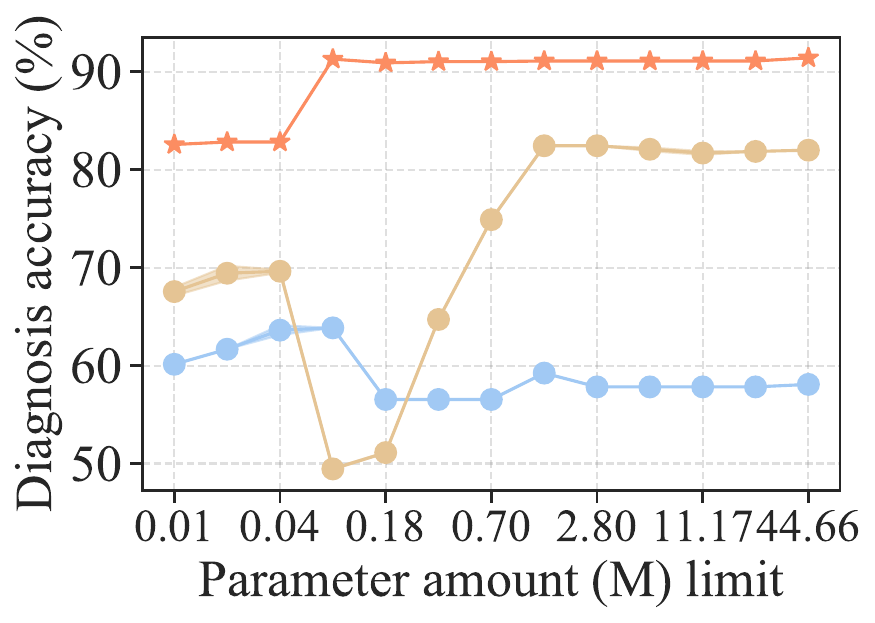}
        \caption{Scale transfer: limiting model's parameter amount}~\label{fig:temp-fail-para}
    \end{subfigure} 
    \caption{\textbf{(Comparing \ourmethod to baseline methods on Q1 tasks with dataset and scale transfer).}
    The $y$-axis indicates the diagnosis accuracy.
    (a) The $x$-axis indicates the number of pre-trained models used for building the training set. 
    (b) The $x$-axis indicates the maximum amount of training (image) data for training models in the training set.
    (c) The $x$-axis indicates the maximum number of parameters of the models in the training set.\looseness-1
  }~\label{fig:temp-fail} \vspace{-6mm}
\end{figure}

\begin{figure*}[!th]
    \centering
    \begin{minipage}[b]{0.48\linewidth}
        \includegraphics[width=\linewidth]{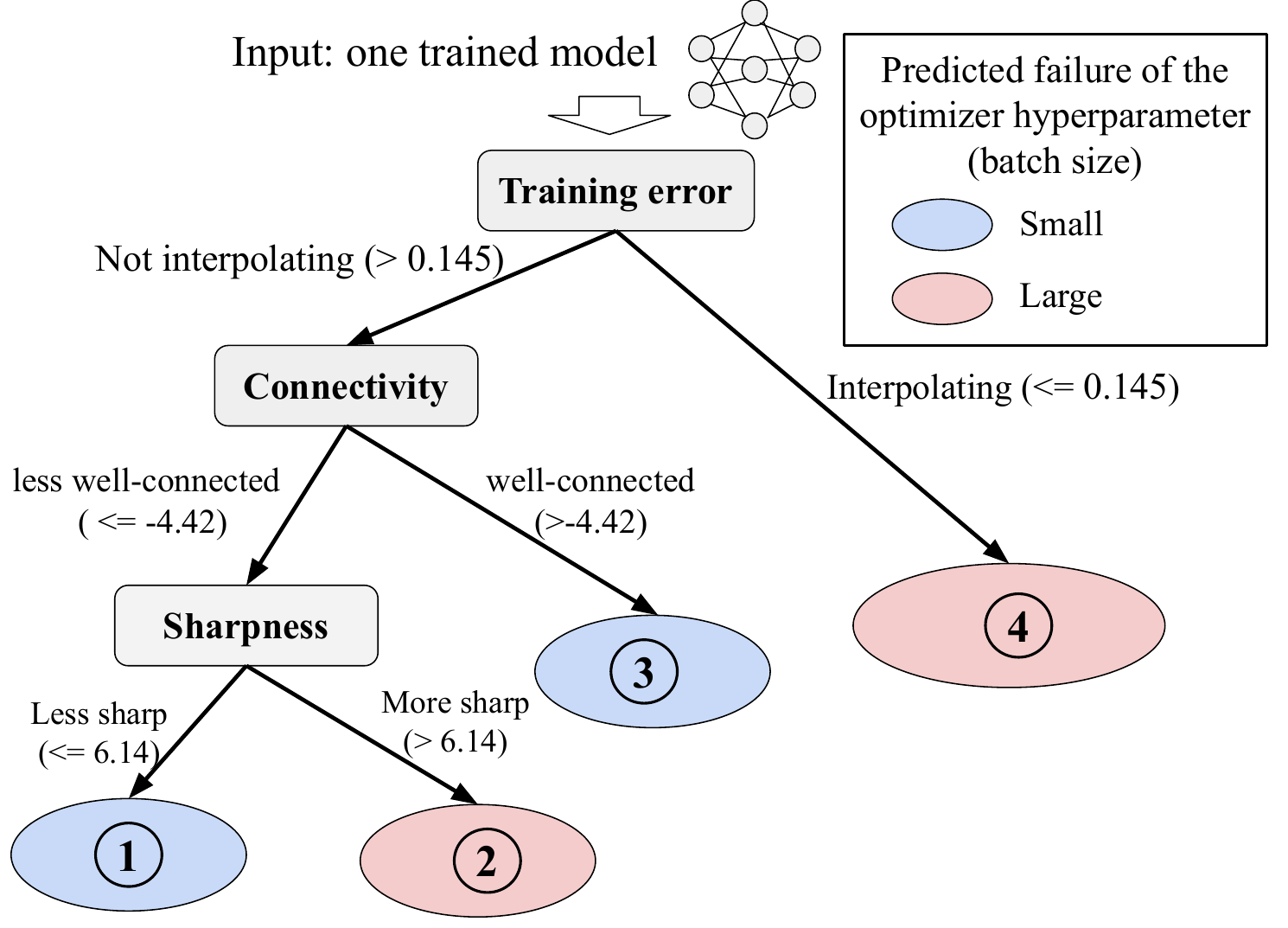}
        \vspace{5mm}
        \subcaption{MD tree for Q1}\label{fig:zero-shot-temp-tree}
    \end{minipage}
    \begin{minipage}[b]{0.48\linewidth}
    \centering
    \begin{subfigure}{0.435\linewidth}
        \includegraphics[width=\linewidth]{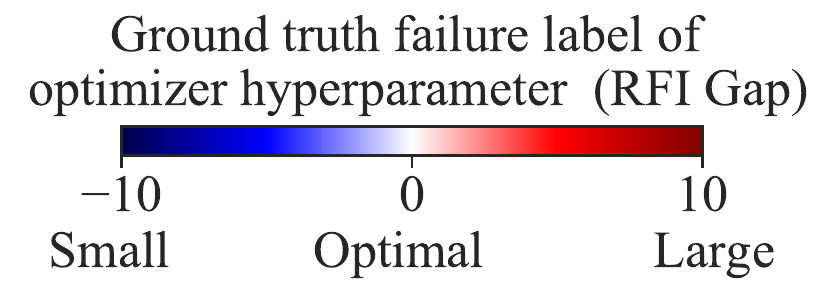}
    \end{subfigure} 
    \begin{subfigure}{0.55\linewidth}
        \includegraphics[width=\linewidth]{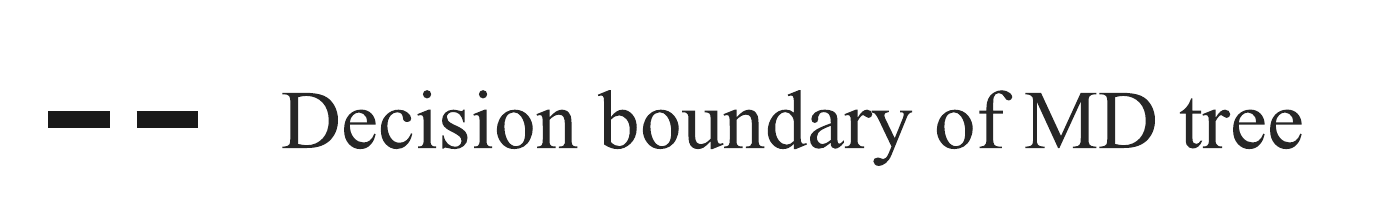}
    \end{subfigure}
    \\
    \begin{subfigure}{0.48\linewidth}
        \includegraphics[width=\linewidth]{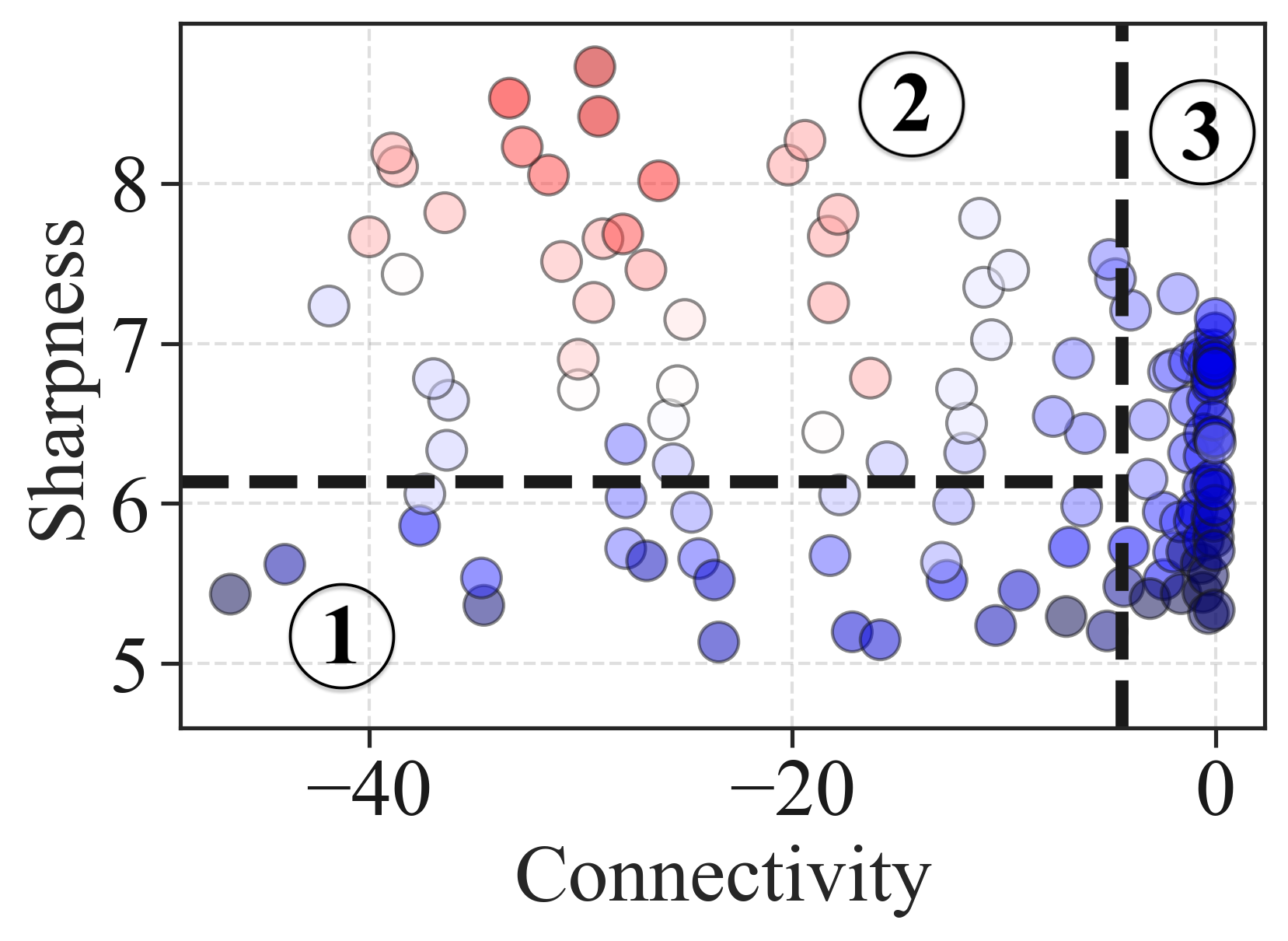}
        \caption{Training (w/o label noise)}\label{fig:temp-vis-a} 
    \end{subfigure} 
    \begin{subfigure}{0.48\linewidth}
        \includegraphics[width=\linewidth]{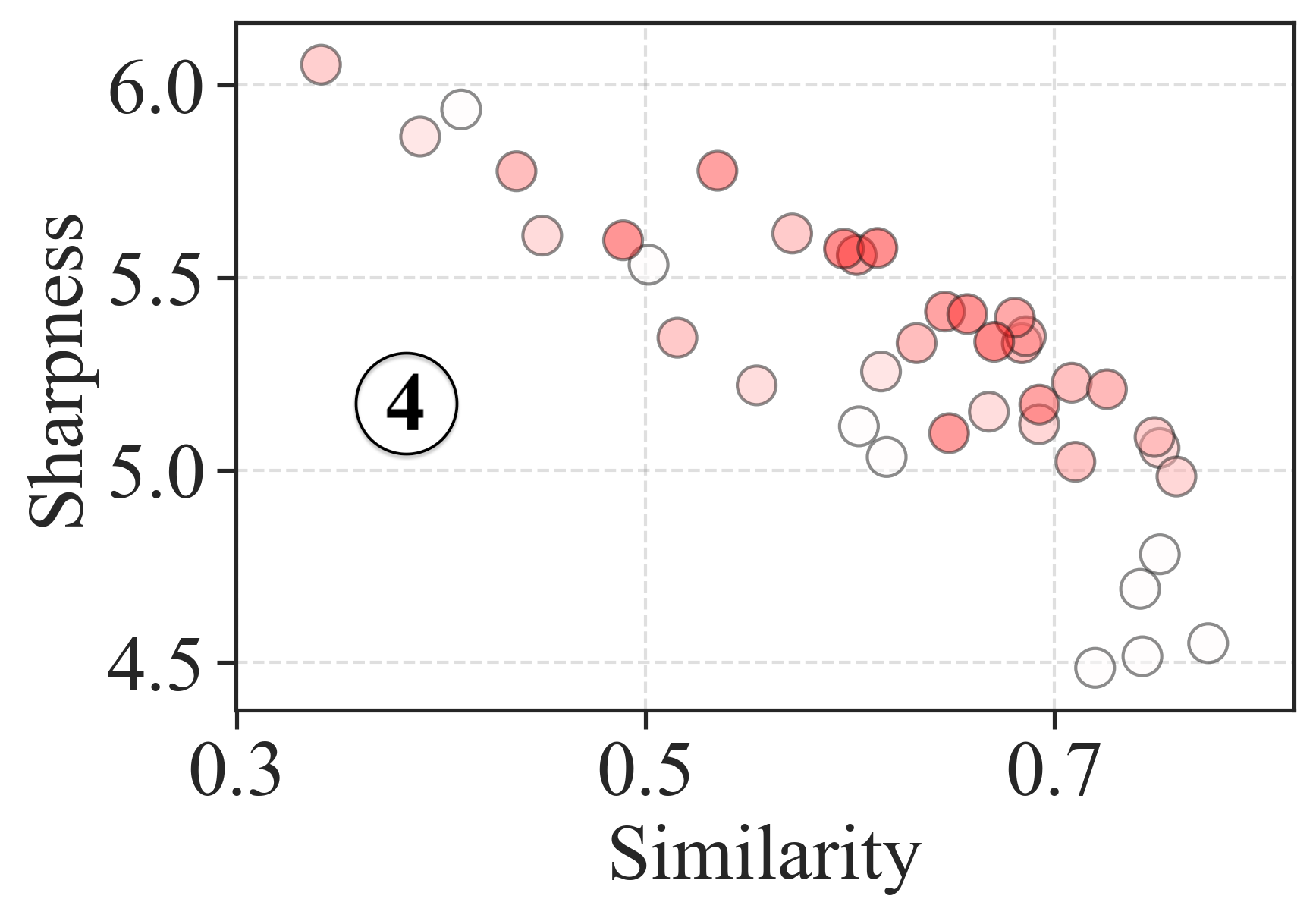}
        \caption{Training (w/o label noise)}\label{fig:temp-vis-b} 
    \end{subfigure} \\
    \begin{subfigure}{0.48\linewidth}
        \includegraphics[width=\linewidth]{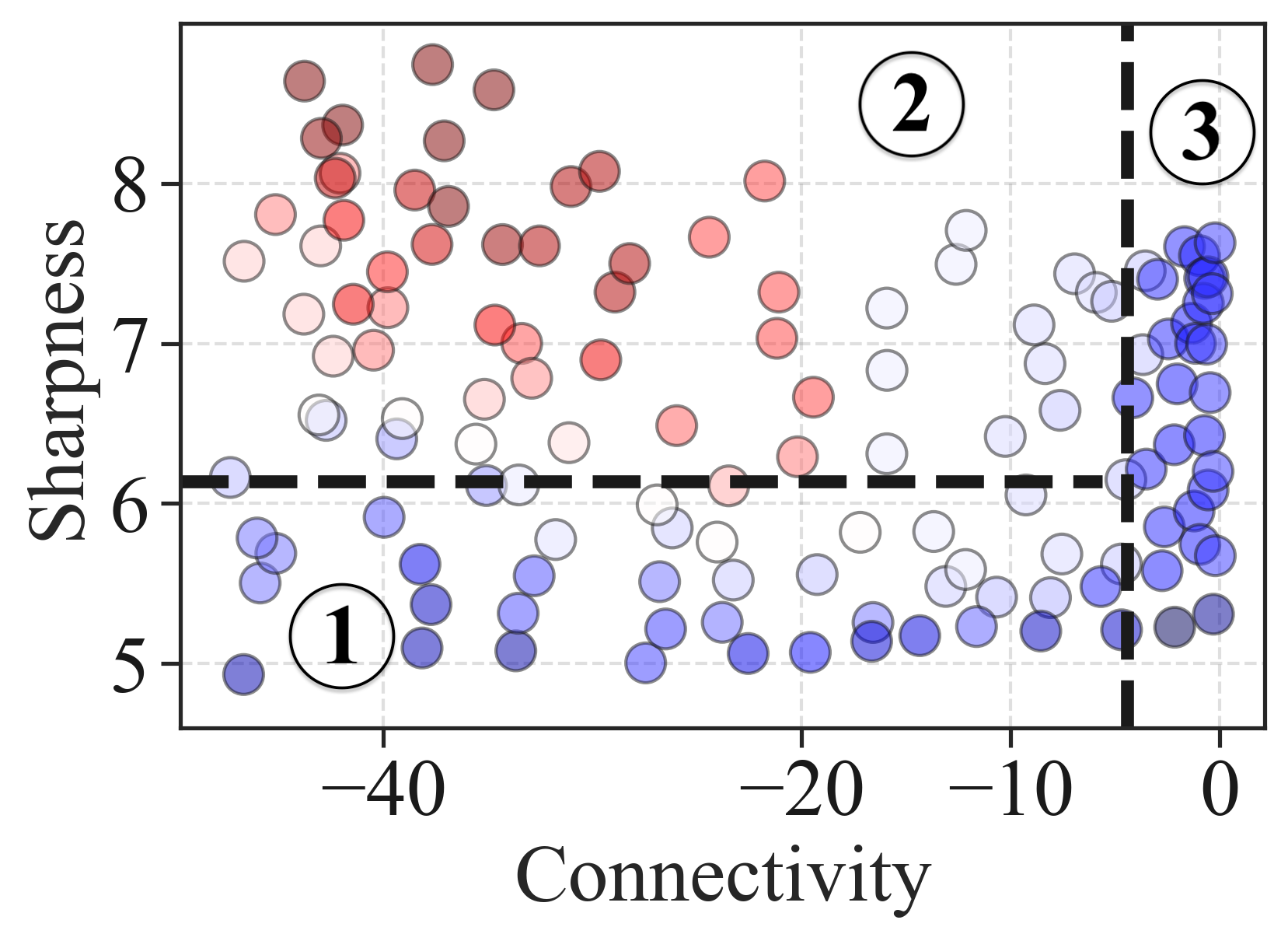}
        \caption{Test (w/ label noise)  } \label{fig:temp-vis-c} 
    \end{subfigure} 
    \begin{subfigure}{0.48\linewidth}
        \includegraphics[width=\linewidth]{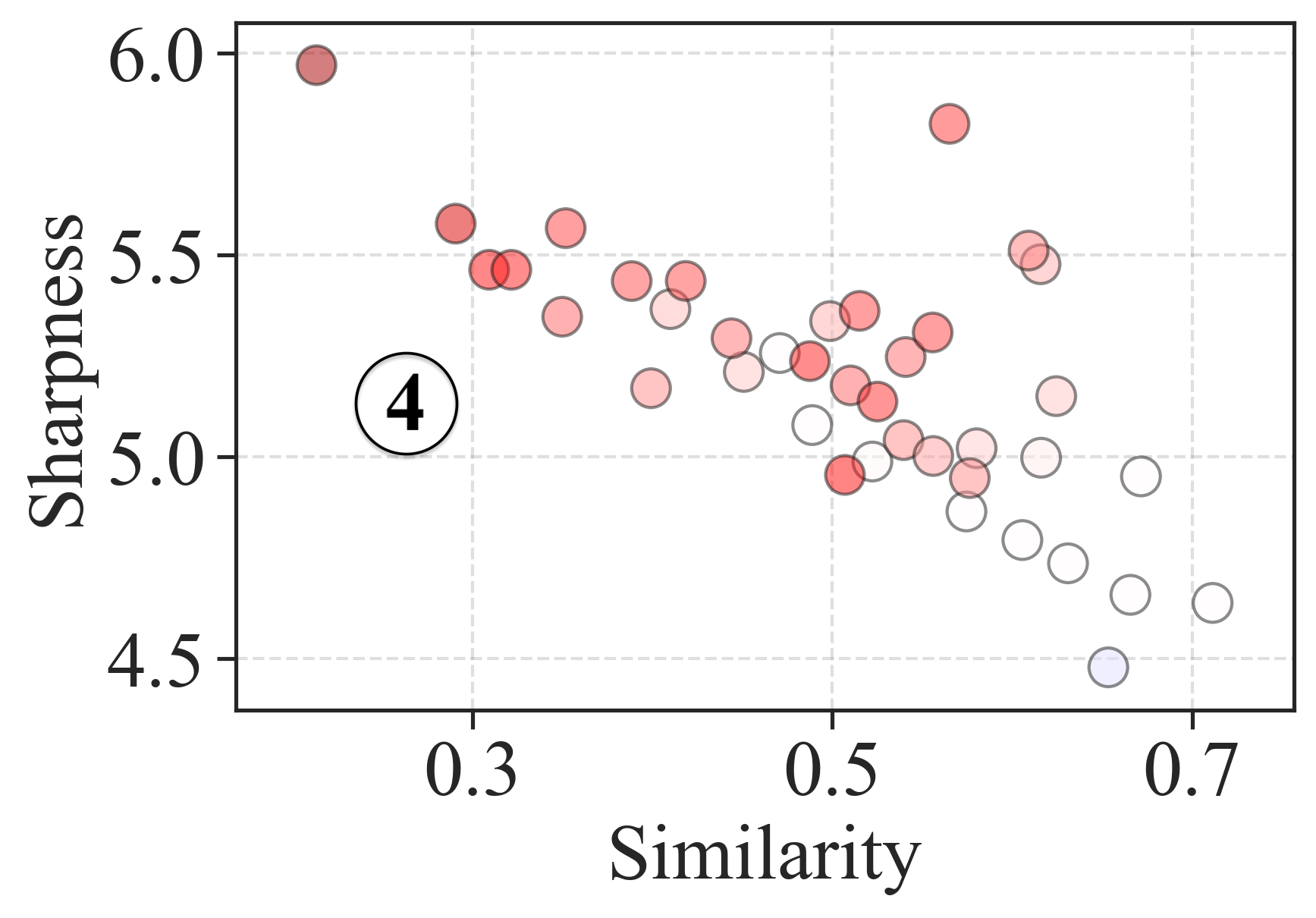}
        \caption{Test (w/ label noise)} \label{fig:temp-vis-d} 
    \end{subfigure} 
    \end{minipage} 
    \caption{\textbf{(Visualizing MD tree and its diagnosis results for Q1).}
    \emph{Left}: structure of the tree defined in Section~\ref{sec:diag-method}. The color of the leaf node indicates the predicted class by \ourmethod.
    The threshold values are learned from the training set.
    \emph{Right}: 
    The first row represents training samples, and the second row represents test samples.
    Each colored circle represents one sample (which is one pre-trained model configuration), and the color represents the ground-truth label: blue means the hyperparameter is too small, while red means too large.
    The black dashed line indicates the decision boundary of \ourmethod.
    Each numbered regime on the right corresponds to the leaf node with the same number on the tree.
    The samples in~\ref{fig:temp-vis-a} and those in~\ref{fig:temp-vis-b} are separated by training error. The same applies to~\ref{fig:temp-vis-c} and~\ref{fig:temp-vis-d}.\looseness-1
    }~\label{fig:zero-shot-temp}  \vspace{-5mm}
\end{figure*}

\subsection{Evaluating Diagnosis Methods}\label{sec:q1-few-shot}
We evaluate \ourmethod by comparing it with two baseline methods: normal DT with {\bf Validation} or {\bf Hyperparameter} as features, defined in Sections~\ref{sec:few-shot-method}.
In Figure~\ref{fig:temp-fail}, we report the diagnosis accuracy in the few-shot setting on two tasks: dataset transfer and scale transfer.
We repeat all experiments in Figure~\ref{fig:temp-fail} for five runs and report the mean and standard deviation.
Figure~\ref{fig:temp-fail-random} shows the diagnosis accuracy versus the size of the training set, for which the training data is randomly subsampled, which leads to a large standard deviation.
Figure~\ref{fig:temp-fail-data} and Figure~\ref{fig:temp-fail-para}, on the other hand, show the results of model diagnosis when the training set is restricted to the subset of models trained with limited data or model size. We do not perform any subsampling in this subset. Therefore, the training set is deterministic, and the diagnostic accuracy has low standard deviations. In Appendix~\ref{sec:corr-result-baseline}, we conduct an ablation study comparing our method with baselines using normal DT with combined Validation and Hyperparameter features.

\paragraph{\ourmethod provides accurate and interpretable diagnosis.} Figure~\ref{fig:temp-fail-random} shows that \ourmethod reaches around 87.7\% accuracy when trained using 96 samples.
When trained with 96 samples, \ourmethod outperforms normal DTs with Validation metrics by 14.88\%. When trained with 12 samples, \ourmethod can still reach an accuracy of 76.64\%.

Further, in Figure~\ref{fig:zero-shot-temp}, we analyze \ourmethod by visualizing how the samples are categorized into each leaf node.
On the left side, we show the \ourmethod structure along with the metric thresholds used to split each internal node. The color of each leaf node represents the predicted labels.
On the right side, we illustrate the classification boundaries (black dashed lines) established by \ourmethod.
Each regime corresponds to a leaf node on the left side, and the color of each circle represents the actual label of the failure source. 
In Figures~\ref{fig:temp-vis-a} and \ref{fig:temp-vis-b}, we see that the regimes (numbered from~\textcircled{\raisebox{-0.9pt}{1}} to~\textcircled{\raisebox{-0.9pt}{4}}), separated by \ourmethod's black boundaries, mainly consist of circles of uniform color, matching the color of the corresponding leaf node in Figure~\ref{fig:zero-shot-temp-tree}.
The same observation holds for test samples shown in Figure~\ref{fig:temp-vis-c} and~\ref{fig:temp-vis-d}.
This indicates that \ourmethod accurately classifies samples into the correct categories.
More importantly, the diagnosis results from \ourmethod are interpretable. For example, it can predict that a model's batch size is too large because that leads to poor connectivity and sharper local minima (regime~\textcircled{\raisebox{-0.9pt}{2}}), which is indeed an issue with large-batch training discussed in~\citet{keskar2017large}.\looseness-1

\paragraph{Loss landscape metrics are more effective than validation metrics in diagnosing model failures.} 
Comparing the visualization of \ourmethod in Figure~\ref{fig:zero-shot-temp} and validation metrics in Figure~\ref{fig:zero-shot-temp-test} further explains why \ourmethod outperforms the latter in diagnosis accuracy.
Validation-based metrics result in nonlinear classification boundaries between failure sources, with significant discrepancies between models in the training set (Figure~\ref{fig:zero-shot-temp-test-a}, w/o label noise) and models in the test set (Figure~\ref{fig:zero-shot-temp-test-c}).
On the contrary, in Figure~\ref{fig:zero-shot-temp}, representing pre-trained models using loss landscape metrics as features makes them piecewise linearly separable. In addition, the threshold (or boundaries) learned during training (top) can be generalized to the test set (bottom).\looseness-1

\paragraph{\ourmethod has high transferability from small-scale to large-scale models.}
As evidenced in Figures~\ref{fig:temp-fail-data} and \ref{fig:temp-fail-para}, \ourmethod demonstrates strong transferability from small-scale to large-scale models.
Recall that the test set $F^{\prime}$ comprises 1560 pre-trained models, including large-scale models trained with up to 50K data points and 44.66M parameters.
From Figure~\ref{fig:temp-fail-data}, we see that \ourmethod achieves 75.90\% accuracy in predicting across the entire test set $\mathcal{F}^{\prime}$ when trained on small-scale models with only 5K data points. In Figure~\ref{fig:temp-fail-para}, \ourmethod achieves 82.56\% accuracy on $\mathcal{F}^{\prime}$ when trained on models with fewer than 0.01M parameters. This high transferability is attributed to the universal multi-regime pattern in the space of model configurations (as illustrated in~\citet{yang2021taxonomizing} using extensive experiments), which emerges early even if we only consider small-scale models.
The multi-regime pattern is further illustrated in Figure~\ref{fig:vis-scale-data-transfer}, which shows that the decision boundaries established by \ourmethod for models trained with small data amounts closely resemble those of models trained with large data amounts.
This observation underscores the practicality of our method, as \ourmethod can be trained in few-shot experiments using small-scale models.
In turn, \ourmethod can be used to diagnose failures in large-scale models.

\paragraph{A notable transition of diagnosis performance is observed in scale transfer.} 
A significant transition in diagnostic performance occurs during scale transfer. 
In Figure~\ref{fig:temp-fail-para}, we note a sudden increase in \ourmethod's accuracy from 82.82\% to 91.28\% when the training set begins to include models with 0.10M parameters. 
This jump in accuracy can be attributed to including models that represent a more complete range of loss landscape regimes. 
Figure~\ref{fig:vis-scale-parameter-transfer} in the Appendix provides the details of this transition.
Models smaller than 0.10M parameters are often limited to a certain range of loss landscape metric values, such as consistently negative mode connectivity, making it challenging to establish accurate and generalizable diagnostic thresholds.
This highlights the challenges in extrapolating diagnostic insights from very small-scale models to larger ones, a phenomenon discussed by~\citet{zhou2023three}, where the optimal hyperparameters differ significantly between small and large models.
Interestingly, this also suggests that, for an effective diagnosis of extremely large models, extremely large models are not necessary; medium-sized models (those with adequate mode connectivity, as defined by~\citet{zhou2023three}), are sufficient.\looseness-1

\section{Which Hyperparameter Leads to Failure}
We now turn to Q2, which considers situations where, despite the possibility of altering two hyperparameters for improved performance, budget constraints require prioritizing the search for only one of them.
Specifically, we compare two factors: the optimizer hyperparameter ($t$) and the model size ($p$), defined in~\eqref{eqn:Q2}.
Similar to Q1, we also study a one-step configuration change task in Appendix~\ref{sec:one-step-q2}.
We will evaluate \ourmethod using the accuracy of the diagnosis.
In addition, we study a variation of Q2 that focuses on the amount of data versus the optimizer\ in Appendix~\ref{sec:appendix-data-temp}. All findings demonstrate the effectiveness of \ourmethod.
\looseness-1

\begin{figure*}[!th]
    \hfill\begin{subfigure}{0.32\linewidth}
        \includegraphics[width=\linewidth]{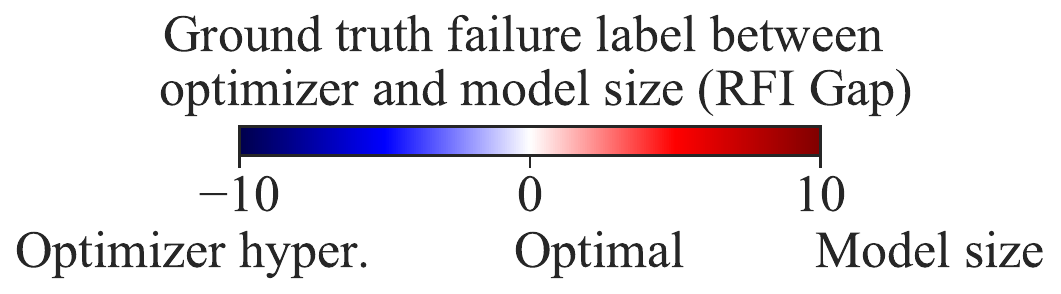}
    \end{subfigure} 
    \begin{subfigure}{0.32\linewidth}
        \includegraphics[width=\linewidth]{figs/md_tree/decision_bd.pdf}
    \end{subfigure} \\
    \begin{minipage}[b]{0.48\linewidth}
        \centering
        \includegraphics[width=\linewidth]{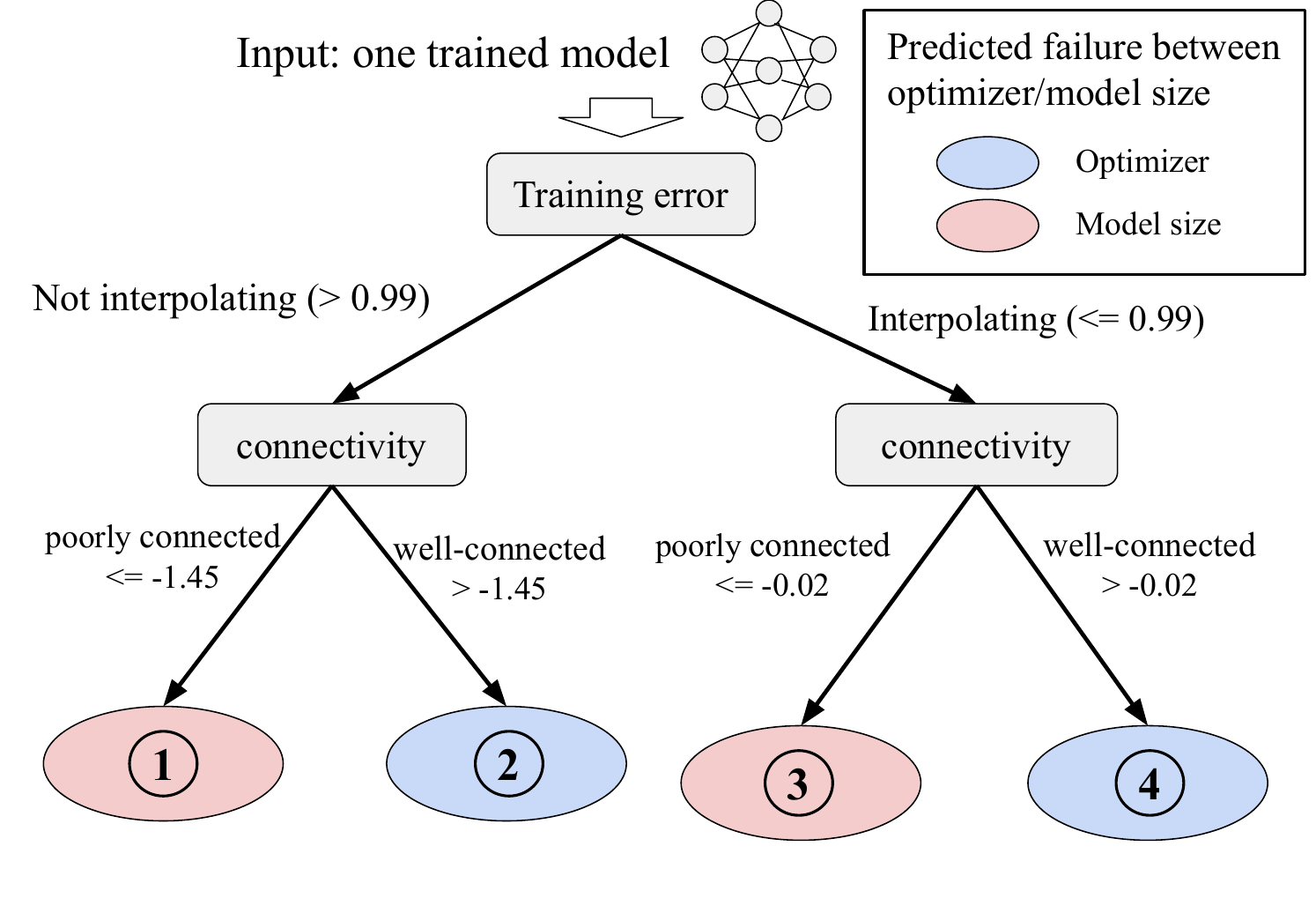}
        \vspace{5mm}
        \subcaption{MD tree for Q2}~\label{fig:zero-shot-width-temp-tree}
    \end{minipage}
    \begin{minipage}[b]{0.48\linewidth}
    \centering
    \begin{subfigure}{0.48\linewidth}
        \includegraphics[width=\linewidth]{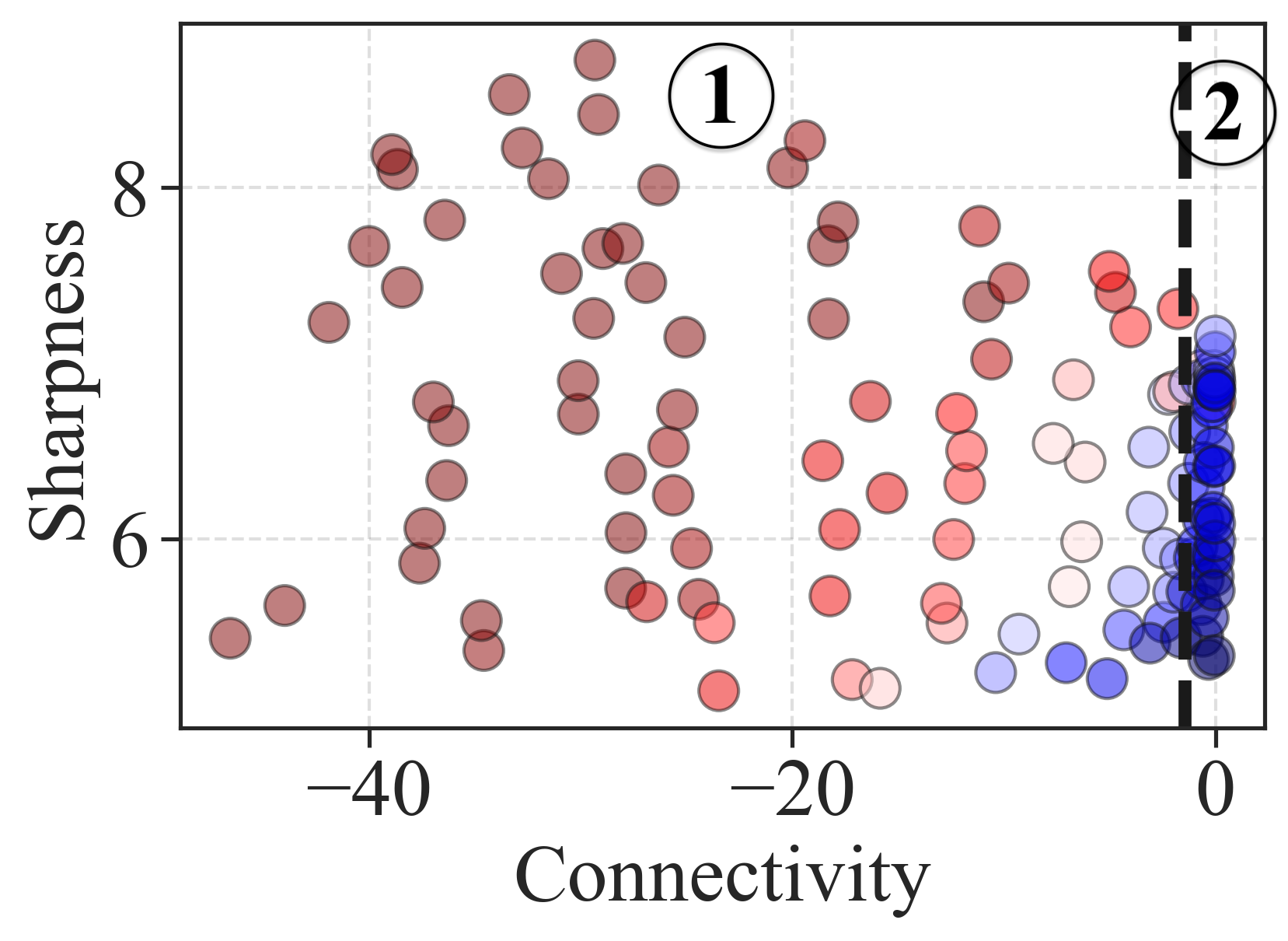}
        \caption{Training (w/o label noise)}~\label{fig:zero-shot-width-temp-b}
    \end{subfigure} 
    \begin{subfigure}{0.5\linewidth}
        \includegraphics[width=\linewidth]{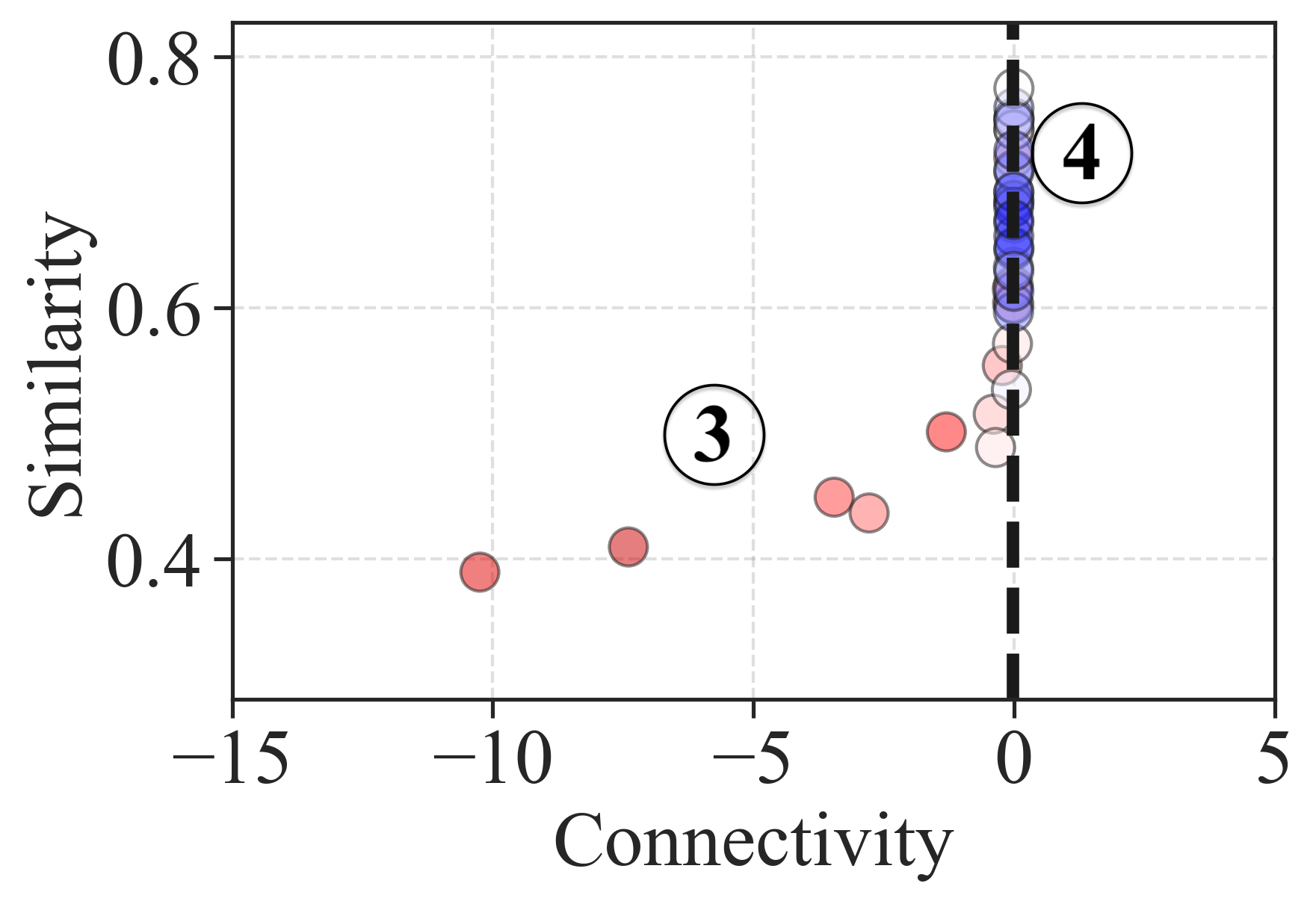}
        \caption{Training (w/o label noise)}~\label{fig:zero-shot-width-temp-c}
    \end{subfigure} \\
    \begin{subfigure}{0.48\linewidth}
        \includegraphics[width=\linewidth]{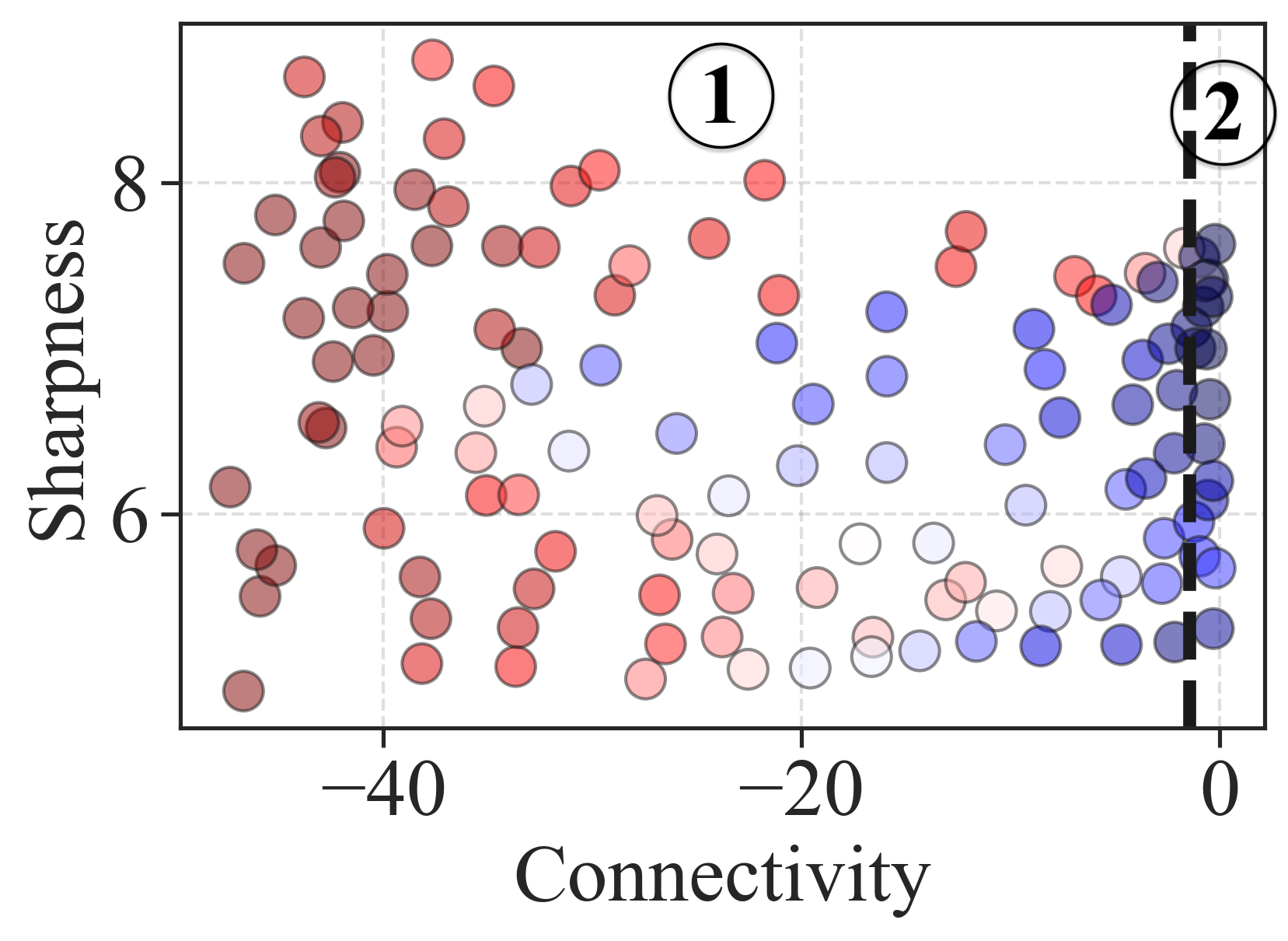}
        \caption{Test (w/ label noise)}~\label{fig:zero-shot-width-temp-d}
    \end{subfigure} 
    \begin{subfigure}{0.5\linewidth}
        \includegraphics[width=\linewidth]{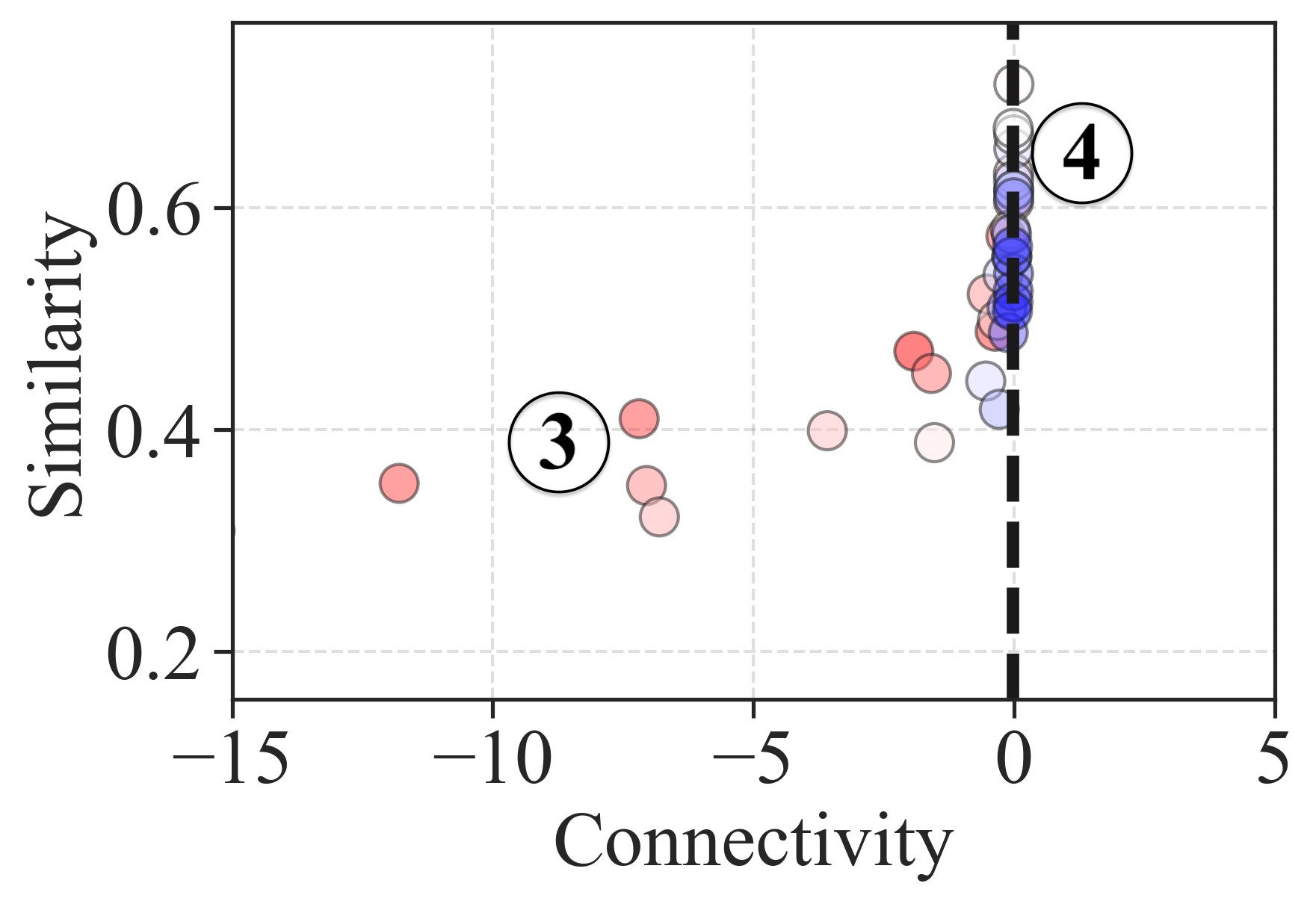}
        \caption{Test (w/ label noise)}~\label{fig:zero-shot-width-temp-e}
    \end{subfigure} 
    \end{minipage} \vspace{-3mm}
    \caption{
    \textbf{(Visualizing MD tree and its diagnosis results for Q2).}
    \emph{Left}: structure of the tree defined in Section~\ref{sec:diag-method}. The color of the leaf node indicates the predicted class by \ourmethod.
    The threshold values are learned from the training set.
    \emph{Right}: 
    The first row represents training samples, and the second row represents test samples.
    Each colored circle represents one sample (which is one pre-trained model configuration), and the color represents the ground-truth label: blue means the failure source is the optimizer, while red means model size.
    The black dashed line indicates the decision boundary of \ourmethod.
    Each numbered regime on the right corresponds to the leaf node with the same number on the tree.
    The samples in~\ref{fig:zero-shot-width-temp-b} and those in~\ref{fig:zero-shot-width-temp-c} are separated by training error. The same applies to~\ref{fig:zero-shot-width-temp-d} and~\ref{fig:zero-shot-width-temp-e}.}~\label{fig:zero-shot-width-temp} \vspace{-5mm}
\end{figure*}

\subsection{Evaluating Diagnosis Methods}\label{sec:q2-few-shot} 
\begin{figure}[!th]
    \centering
    \includegraphics[width=0.7\linewidth,keepaspectratio]{figs/md_tree/temp_legend.pdf} \\
    \begin{subfigure}{0.32\linewidth}
        \includegraphics[width=\linewidth]{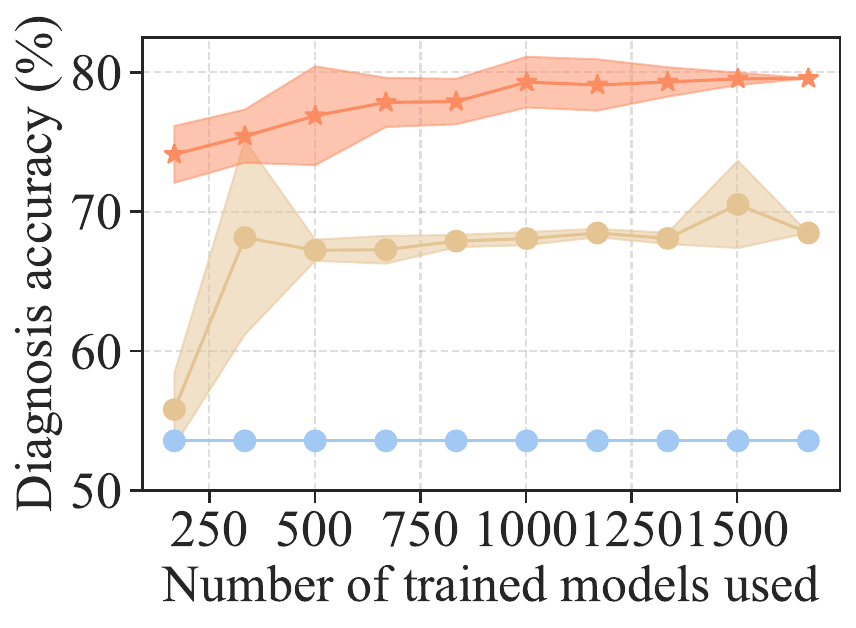}
        \caption{Dataset transfer}~\label{fig:width-vs-temp-a}
    \end{subfigure} 
    \begin{subfigure}{0.32\linewidth}
        \includegraphics[width=\linewidth]{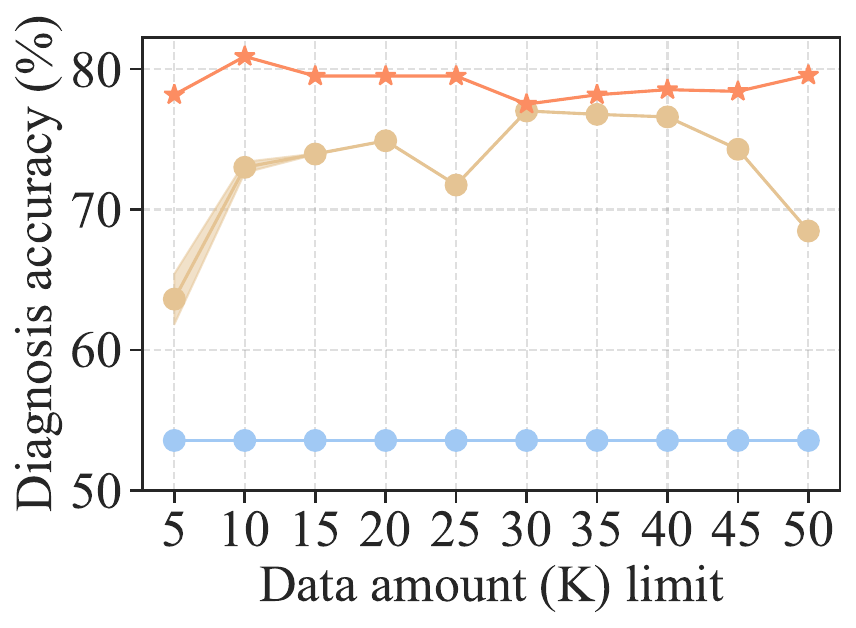}
        \caption{Scale transfer}~\label{fig:width-vs-temp-b}
    \end{subfigure} \vspace{-4mm}
    \caption{\textbf{(Comparing \ourmethod to baseline methods on Q2 in dataset and scale transfer).} 
    $y$-axis indicates the diagnosis accuracy.
    (a) $x$-axis indicates the number of pre-trained models in the training set.
    (b) $x$-axis indicates the maximum amount of training (image) data for the models in the training set.\looseness-1
    }\vspace{-4mm}~\label{fig:width-vs-temp} 
\end{figure}

Similar to Section~\ref{sec:q1-few-shot}, we compare \ourmethod with the baseline methods and provide an interpretable visualization in Figure~\ref{fig:zero-shot-width-temp}. The diagnosis accuracy is reported in Figure~\ref{fig:width-vs-temp}.

\paragraph{\ourmethod can provide interpretable visualization and transfer well from small-scale to large-scale models.}
In Figure~\ref{fig:width-vs-temp-a}, \ourmethod reaches 74.11\% with 160 trained models, significantly outperforming the validation-based method by 18.31\%.
The visualizations in Figure~\ref{fig:zero-shot-width-temp} highlight \ourmethod's ability to use loss landscape metrics to distinctly separate models. In contrast, the validation metrics shown in Figure~\ref{fig:zero-shot-width-temp-test} present complex and less interpretable boundaries.

Figure~\ref{fig:width-vs-temp-b} demonstrates \ourmethod's effective transferability, maintaining high performance (78.17\%) even when trained on small-scale models with 5K data points. This is supported by Figure~\ref{fig:vis-scale-data-transfer-q2}, which shows that the connectivity-based decision boundary established with limited data can be successfully applied to models trained with larger data amount.
However, \ourmethod exhibits some limitations, such as the misclassification of some test models in Figure~\ref{fig:zero-shot-width-temp-d}, numbered regime~\textcircled{\raisebox{-0.9pt}{1}}. This limitation results in \ourmethod achieving a maximum accuracy of 79.56\% in Q2, which is lower than the 89.85\% accuracy in Q1.
This could be attributed to shifts in the model's distribution caused by training with label noise.
Future research could explore whether fine-tuning in the target domain enhances \ourmethod's generalization.

\begin{wrapfigure}{r}{0.55\textwidth}
    \centering
    \includegraphics[width=0.95\linewidth]{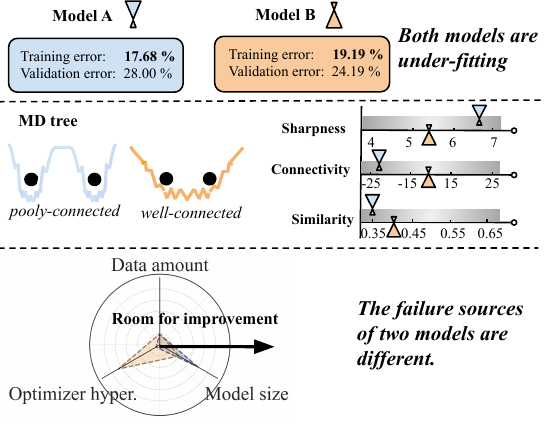}~\label{fig:model-vs-hyper}
    \caption{\textbf{(Case Study: \ourmethod vs. validation method for Q2).}
    \emph{Top}: Validation metrics provide limited diagnosis: both models have the same underfitting issues.
    \emph{Middle}: Loss landscape metrics distinguish the models: Model A has poor connectivity, while Model B has good connectivity.
    \emph{Bottom}: Model A suffers from a size issue, and Model B has a training hyperparameter issue.\looseness-1} \vspace{-6mm}
\label{fig:one-out-of-two-model} 
\end{wrapfigure}

\paragraph{Comparing \ourmethod with validation-based methods.}
We provide insights into the advantages of using loss landscape metrics in \ourmethod. We find that loss landscape metrics are more sensitive to hyperparameter changes than validation metrics. This is demonstrated by the case study in Figure~\ref{fig:one-out-of-two-model}, which addresses Q2 in a \emph{zero-shot scenario}. 
Two pre-trained models, A and B, need to be diagnosed. 
Validation metrics only show both models with high training errors, indicating underfitting, but fail to distinguish the different failure sources (optimizer vs. model size) that lead to the underfitting.
However, loss landscape metrics reveal that Model A, with poor connectivity, suffers from model size issues, whereas Model B, with good connectivity, is affected by optimizer settings.
This is further confirmed by the radar plot, where Model A shows a higher RFI in model size, and Model B has a higher RFI in optimizer hyperparameters. 
In Figure~\ref{fig:one-out-of-two-data-temp}, we show a similar zero-shot analysis on two overfitting models.

\paragraph{The importance of \ourmethod's tree hierarchy.}
Lastly, we evaluate the role of \ourmethod's fixed tree structure, which partitions data according to a specific sequence of metrics: training error, connectivity, and then sharpness or similarity.
This structure is inspired by~\citet{yang2021taxonomizing}, which highlights the usefulness of encoding known inductive biases in model diagnosis.
We evaluate a conventional DT employing the same loss landscape metrics 
\{$\mathcal{E}_\text{tr},\mathcal{C}, \mathcal{H}_{t}, \mathcal{S}$\} but without adhering to this precise order. As shown in Figure~\ref{fig:fix-abl-q1-q2}, \ourmethod consistently outperforms standard DT in Q1 and Q2, especially with a limited number of trained models. This shows that this inductive bias imposes a beneficial regularization on \ourmethod when data is limited. Details are provided in Appendix~\ref{sec:corr-result-tree-hier}.

\section{Conclusion}
\vspace{-1mm}
In conclusion, \ourmethod introduces an innovative framework for diagnosing the underperformance of trained NN models without retraining, meeting the urgent need for effective diagnostics with low computational cost.
Using loss landscape metrics, \ourmethod outperforms conventional validation-based methods by providing more accurate and generalizable diagnoses.
Through quantitative few-shot classifier predictions and qualitative visualization, \ourmethod proves its efficacy in identifying critical failure sources, such as inappropriate optimizer hyperparameters or inadequate model sizes.
This has been demonstrated in various scenarios, including dataset and scale transfers, without relying on detailed training configurations.

\section*{Impact Statement}
Our research focuses on creating a diagnostic method for pre-trained models. 
Although \ourmethod can be used in various applications, including those with potential adverse effects, the algorithm itself does not present immediate negative social impacts.
On the contrary, \ourmethod has significant social value, particularly in helping universities and researchers with limited computing resources to effectively use pre-trained models.

\section*{Acknowledgment}
We would like to thank Caleb Geniesse and Tiankai Xie for helping with the implementation of loss landscape metrics. We would like to thank Vignesh Kothapalli, Tianyu Pang, Lei Hsiung, and Michael Mahoney for their valuable feedback.

\bibliographystyle{plainnat}
\bibliography{example_paper}

\newpage
\appendix

\section*{Appendix}

\section{Applying Diagnosis to One-step Configuration Change}~\label{sec:one-step-change}
\vspace{-5mm}

\subsection{Experimental Setup}\label{sec:appendix-one-step-change}
Here we elaborate on the experimental setup for the one-step configuration change task.
The one-step change includes 1) determining the change direction, and 2) determining the size of the adjustment step.
The first one, for Q1, means determining whether to increase or decrease the optimizer hyperparameter. 
For Q2, it means first determining whether to increase the model size or adjust the optimizer hyperparameter, if the first decision chooses the latter one, then the second decision has to be made to decide whether to increase or decrease.
This change direction is determined by the diagnosis method such as \ourmethod, e.g. if \ourmethod finds that the failure source is the optimizer hyperparameter is too large ($m_t^{\uparrow}$), then we choose to decrease it.

In addition to comparing \ourmethod with the diagnosis methods \textbf{Validation} + DT and \textbf{Hyperparameter} + DT, we also compare it with Random diagnosis and Optimal diagnosis. Random diagnosis, which selects the change direction randomly between two options, represents the lower bound of performance for diagnosis-based hyperparameter changes.
Optimal diagnosis, using the ground-truth label to choose the best direction, represents the upper bound of performance, equivalent to a diagnosis method achieving 100\% test accuracy.
We examine three approaches for determining the change steps: ``fixed (one) step size'', ``random step size'', and ``optimal step size''.
The ``fixed'' refers to taking one tuning step, ``random'' refers to taking a random step size along the determined direction, and ``optimal'' refers to taking the step size that can reach the best performance along the chosen direction, representing the upper bound of performance improvement of a specific diagnosis method.\looseness-1

\subsection{One-step Change on Optimizer Hyperparameter (Q1)}\label{sec:one-step-q1}
We extend the practicality of \ourmethod, by applying it to a task involving a one-step change in the model configuration.
These methods predict the direction--either an increase or decrease--of an adjustment in an optimizer hyperparameter (e.g. batch size here). 
This adjustment is executed using one of three different types of step size: a fixed (one) step size, a random step size, or the optimal step size.
We then evaluate the impact of this adjustment by measuring the test accuracy improvement of the adjusted model on CIFAR-10. 
Diagnosis methods are built using the training set $\mathcal{F}$ and evaluated on the test set $\mathcal{F}^{\prime}$.
More information on the setup and evaluation can be found in Appendix~\ref{sec:appendix-one-step-change}.

Table~\ref{tb:one-step-temp} shows that \ourmethod enhances task performance, outperforming three baselines, including the {\bf Validation} approach, across three different step size settings.
The task performance gains correlate with the diagnostic accuracy shown in Figure~\ref{fig:temp-fail}, with \ourmethod leading, followed by the {\bf Validation} method, and then the {\bf Hyperparameter} approaches. 
An important takeaway is the impact of accurate diagnostics on improving task performance through a single hyperparameter adjustment.
Specifically, an accurate diagnosis like \ourmethod can boost performance by 5.52\% with the optimal step size, markedly better than the Random diagnosis (3.50\%).
This highlights the importance of precise diagnostic tools for effective tuning. 
Future work will explore predicting the optimal step size for hyperparameter adjustments using loss landscape metrics.

\begin{table}[!h]
\centering
\caption{\textbf{(Q1: task performance improvement (CIFAR-10 accuracy ($\uparrow$, \%)) of performing one-step change only on the optimizer hyperparameter).}
The diagnosis methods decide whether to increase or decrease the given trained model's hyperparameter and then the decision is combined with three types of step sizes. 
We trained using 96 configurations randomly sampled from $\mathcal{F}$ and evaluated the average improvement across 1560 sub-optimal model configurations from $\mathcal{F}^{\prime}$ for testing. Each experiment was repeated with five random seeds, and we reported the mean and standard deviation.\looseness-1}
\vspace{1mm}
\resizebox{0.7\linewidth}{!}{
\begin{tabular}{c|ccc}
\toprule
\textbf{Diagnosis method}  & w/ fixed step      & w/ random step   & w/ optimal step  \\
\midrule
Optimal diagnosis     & 1.48 $\pm$ 0.00 & 2.92 $\pm$ 0.03 & 5.83 $\pm$ 0.00 \\
\midrule
Random diagnosis     & 0.92 $\pm$ 0.03  & 1.78 $\pm$ 0.03 & 3.50 $\pm$ 0.10 \\
\textbf{Hyperparameter} + DT        & 0.92 $\pm$ 0.07  & 1.84 $\pm$ 0.26 & 3.52 $\pm$ 0.50 \\
\textbf{Validation} + DT & 1.24 $\pm$ 0.12 & 2.31 $\pm$ 0.17 & 4.63 $\pm$ 0.23 \\
\gb \ourmethod  & \textbf{1.34 $\pm$ 0.06} & \textbf{2.74 $\pm$ 0.13} & \textbf{5.52 $\pm$ 0.15}  \\
\bottomrule
\end{tabular} \label{tb:one-step-temp}
}
\end{table}

\subsection{One-step Change on Optimizer Hyperparameter or Model Size (Q2)}\label{sec:one-step-q2}
We applied the diagnosis methods for addressing Q2 to a one-step configuration change task, with the setup detailed in Appendix~\ref{sec:appendix-one-step-change}.
The diagnosis methods decide whether to adjust the optimizer hyperparameter or increase the model size. 
If the former option is selected, the diagnosis method for Q1 will further determine whether to increase or decrease the hyperparameter, and all decisions are combined with three types of adjustment step sizes.
Table~\ref{tab:task-improve-width-temp} demonstrates that \ourmethod surpasses three baseline methods in enhancing task performance, affirming its better diagnostic accuracy. 
This experiment highlights the importance of precise hyperparameter adjustment for task improvement, a goal that \ourmethod facilitates. \looseness-1

\begin{table}[!th]
\centering
\caption{
\textbf{(Q2: task performance improvement (CIFAR-10 accuracy ($\uparrow$, \%)) of performing one-step hyperparameter change on the optimizer hyperparameter and model size).}
The diagnosis method decides whether to adjust the optimizer hyperparameter or increase the model size. 
We used 156 configurations randomly sampled from $\mathcal{F}$ for training and evaluated the average improvement on 1560 sub-optimal configurations from $\mathcal{F}^{\prime}$. Each experiment was repeated with five random seeds, and we reported the mean and standard deviation.\looseness-1
}\label{tab:task-improve-width-temp}
\vspace{2mm}
\resizebox{0.7\linewidth}{!}{
\begin{tabular}{c|ccc}
\toprule
\textbf{Method}           & w/ fixed step      & w/ random step   & w/ optimal step  \\
\midrule
Optimal diagnosis     & 1.92 $\pm$ 0.00 & 5.47 $\pm$ 0.05 & 8.83 $\pm$ 0.00 \\
\midrule
Random diagnosis    & 1.19 $\pm$ 0.03 & 3.09 $\pm$ 0.11 & 5.03 $\pm$ 0.10 \\
\textbf{Hyperparameter} + DT          & 1.48 $\pm$ 0.00 & 4.36 $\pm$ 0.09 & 6.59 $\pm$ 0.00 \\
\textbf{Validation} + DT  & 1.42 $\pm$ 0.12 & 4.09 $\pm$ 0.59 & 6.33 $\pm$ 0.61 \\
\gb \ourmethod    & \textbf{1.73 $\pm$ 0.04} & \textbf{4.79 $\pm$ 0.16} & \textbf{8.10 $\pm$ 0.22} \\
\bottomrule
\end{tabular}
}
\end{table}

\FloatBarrier
\section{Loss Landscape Metrics}\label{app:loss-land}
In this section, we define loss landscape metrics, then we discuss the related work on sharpness. Lastly, we discuss the computational efficiency of these metrics.

\paragraph{Connectivity Metrics.}
For two models with two parameter configurations $\vtheta, \vtheta'$, we can learn a curve $\gamma(t), t \in [0,1]$ connecting $\vtheta$ and $\vtheta'$ such that $\gamma(0) = \vtheta$, $\gamma(1) = \vtheta'$, and the training error (loss) evaluated on $\gamma(t)$ for any $t\in[0,1]$ is minimal.
One approach to parameterizing $\gamma(t)$ is to use a Bezier curve~\cite{garipov2018loss} with $k+1$ bends, $\gamma_{\mathbf{\phi}}(t) = \sum_{j = 0}^k \binom{k}{j}(1-t)^{k - j}t^j\vtheta_j, t \in [0,1]$, where $\vtheta_0 = \vtheta$, $\vtheta_k = \vtheta^{\prime}$, and $\mathbf{\phi} = \{ \vtheta_1, ..., \vtheta_{k-1} \}$ are trainable parameters of additional models. In this study, we use Bezier curves with three bends ($k = 2$).
We use the peak of the curve to quantify whether there exists a barrier:\looseness-1

\begin{align}
    \centering
    \mathcal{C}(\vtheta,\vtheta') &=  - \mathcal{E}_\text{tr}(\gamma(t^*)),  \nonumber
    \\  
   \text{where, }  
   t^{*}&=\underset{t}{\arg \max }\left|\frac{1}{2}\left(\mathcal{E}_\text{tr}(\vtheta)+\mathcal{E}_\text{tr}\left(\vtheta^{\prime}\right)\right)-\mathcal{E}_\text{tr}\left(\gamma_\mathbf{\phi}(t)\right)\right|
\end{align}

\paragraph{Similarity Metrics.}
CKA similarity measures the difference in features between two parameter configurations $\vtheta, \vtheta'$. Let $\{\mathbf{x}_1,\mathbf{x}_2, \dots, \mathbf{x}_s \}$ be randomly sampled data points, and let $F_{\vtheta} = [f_{\vtheta}(\mathbf{x}_1) \dots f_{\vtheta}(\mathbf{x}_s)]^\top \in \mathbb{R}^{s \times d_{\text{out}}}$ be the output of the network.
Then, the CKA similarity between two parameter configurations $\vtheta, \vtheta'$ is given by:
\begin{equation}
     \mathcal{S}(\vtheta,\vtheta') = \frac{\operatorname{Cov}(F_{\vtheta},F_{\vtheta'})}{\sqrt{\operatorname{Cov}(F_{\vtheta},F_{\vtheta})\operatorname{Cov}(F_{\vtheta'},F_{\vtheta'})}}.
\end{equation}
where $\operatorname{Cov}(\mathbf{X}, \mathbf{Y})=(s-1)^{-2} \operatorname{tr}\left(\mathbf{X} \mathbf{X}^{\top} \mathbf{H}_s \mathbf{Y} \mathbf{Y}^{\top} \mathbf{H}_s\right)$, and $\mathbf{H}_s=\mathbf{I}_s-s^{-1} \mathbf{1 1}^{\top}$ is a centering~matrix. \looseness-1

Measuring the connectivity and similarity metrics necessitates two distinct sets of weights, $\vtheta$ and $\vtheta^{\prime}$, each corresponding to the same configuration trained with different random seeds. 
In the diagnostic dataset, each data sample is a training configuration instead of a single trained model. This is because each configuration contains five pre-trained models trained with different random seeds. For each training configuration, we evaluate a single similarity/connectivity score.

\paragraph{Sharpness Metrics.}
The Hessian matrix at a given point $\vtheta_0$ in the parameter space can be represented as $\nabla_\vtheta^2 \mathcal{L}\left(\vtheta_0\right)$.
We report the leading eigenvalue $\mathcal{H}_e = \lambda_{\max }\left(\nabla_\vtheta^2 \mathcal{L}\left(\vtheta_0\right)\right)$ and the trace $\mathcal{H}_t = \operatorname{tr}\left(\nabla_\vtheta^2 \mathcal{L}\left(\vtheta_0\right)\right)$ to summarize the local curvature properties in a single value.

Here, we discuss related work on sharpness.
We elaborate the connection between \citet{yang2021taxonomizing} and \citet{dinh2017sharp, keskar2017large, foret2021sharpnessaware, yao2018hessian}. 
The connection is that prior work finds that in some cases flat minima have better generalization~\cite{keskar2017large, foret2021sharpnessaware}, but there exist some cases such as adversarial training and reduction of $\ell_2$ regularization that sharp minima generalize better~\cite{dinh2017sharp, yao2018hessian}. \citet{yang2021taxonomizing} studies why this happens, and the explanation was that prior work neglects global loss landscape properties such as connectivity. Furthermore, \citet{yang2021taxonomizing} shows that combining the local and global metrics leads to a more accurate prediction of generalization performance. This motivates us to combine connectivity and sharpness in our MD tree to predict the failure sources. 

\paragraph{Computational Efficiency.} To demonstrate the low computational burden of loss landscape measurements, we provide details on calculating sharpness and connectivity, along with additional runtime results.
For measuring sharpness, we follow the implementation by \citet{yao2020pyhessian}, using the Hessian trace to represent sharpness. The Hessian trace is computed using Hutchinson's method~\cite{10.1145/1944345.1944349} from RandNLA.
Let $\mathbf{H}$ denote the Hessian matrix. This method approximates the Hessian trace using $\mathbb{E}[\mathbf{v}^T\mathbf{H}\mathbf{v}]$, where $\mathbf{v}$ is a random vector with components i.i.d. sampled from the Rademacher distribution. The computational cost of a Hessian matrix-vector multiplication is equivalent to a single gradient backpropagation. We estimate the expectation by drawing multiple random samples until convergence or until 100 iterations are reached. Thus, the cost of measuring sharpness for a trained model is at most 100 backpropagations.
For measuring connectivity, we adopt the implementations of \citet{yang2021taxonomizing} and \citet{garipov2018loss} to compute mode connectivity. This process involves 50 epochs of training to search for a three-bend Bezier curve and five inferences on the training set to evaluate the curve's barrier.

In Table~\ref{tb:compute-eff}, we compare the runtime of training a single model configuration with the time taken to evaluate two loss landscape metrics on the same configuration. The model configuration involves training ResNet18 on CIFAR-10 with a batch size of 128, using five random seeds for 150 epochs. Our measurements indicate that, on average over 10 runs, each training epoch takes 23.32 seconds. We also calculate the percentage increase in time due to measuring loss landscape metrics compared to model training. Our results show that measuring sharpness and connectivity increases the total training time by only 0.244\% and 7.04\%, respectively, indicating that the computational burden is reasonable. The testing platform used was a Quadro RTX 6000 GPU paired with an Intel Xeon Gold 6248 CPU.

\begin{table}[!th]
\centering
\caption{Comparison of runtime for model training versus sharpness and connectivity measurements used in \ourmethod.
}
\resizebox{0.9\linewidth}{!}{
\begin{tabular}{c|ccc}
\toprule
                     & Model training    & Sharpness  &  Connectivity       \\
\midrule
 Runtime (seconds)      & 17492.25                &     	42.73        &         1232.76       \\
 Increase in runtime compared to model training (\%)               &   -           &   0.244\%    &     7.04\%     \\
\bottomrule
\end{tabular}~\label{tb:compute-eff}
} 
\end{table}

\section{Model Diagnosis Framework}\label{app:exp-setup}

\subsection{Room for Improvement (RFI) of Failure Sources}\label{app:rfi-def}
Here, we provide the complete definitions of $\text{RFI}$ for the failure sources. We define four failure sources $\mathcal{M} = \{m^{\downarrow}_{t}, m^{\uparrow}_{t}, m_{p},  m_{n}\}$, and $\text{RFI}$ of $m_{p}$ is defined by \eqref{eqn:RFI} in Section~\ref{sec:failure_sources}.

The RFI of failure source $m^{\uparrow}_{t}$ (the optimizer hyperparameter is larger than the optimal choice) is defined as
\begin{align}\label{eqn:RFI-t-large}
\begin{split}
    \text{RFI}(m^{\uparrow}_{t}, f_0) &= \mathcal{E}_\text{val}(f_0) - \mathcal{E}_\text{val}(f^*),  \\ \text{where}   \quad
    f^{*} &= \mathcal{A}(p, n, q^*), \, \\  q^* &= \argmin_{q \in [t_{\min}, t]} \mathcal{E}_\text{val}(\mathcal{A}(p, n, q)).
\end{split}
\end{align}

The RFI of failure source $m^{\downarrow}_{t}$ (the optimizer hyperparameter is smaller than the optimal choice) is defined as
\begin{align}\label{eqn:RFI-t-small}
\begin{split}
    \text{RFI}(m^{\downarrow}_{t}, f_0) &= \mathcal{E}_\text{val}(f_0) - \mathcal{E}_\text{val}(f^*),  \\ \text{where}   \quad
    f^{*} &= \mathcal{A}(p, n, q^*), \, \\  q^* &= \argmin_{q \in [t, t_{\max}]} \mathcal{E}_\text{val}(\mathcal{A}(p, n, q)).
\end{split}
\end{align}

The RFI of failure source $m_{n}$ (the amount of data $n$ is too small) is defined as
\begin{align}\label{eqn:RFI-n}
\begin{split}
    \text{RFI}(m_{n}, f_0) &= \mathcal{E}_\text{val}(f_0) - \mathcal{E}_\text{val}(f^*),  \\ \text{where}   \quad
    f^{*} &= \mathcal{A}(p, q^*, t), \, \\  q^* &= \argmin_{q \in [n, n_{\max}]} \mathcal{E}_\text{val}(\mathcal{A}(p, q, t)).
\end{split}
\end{align}

\subsection{Rationale for Considering a Binary Classification Problem}\label{app:rationale-binary-class}
We consider a binary classification problem that predicts \{$G > 0$, $G < 0$\}, rather than including near-optimal cases ($G = 0$) to avoid a three-class classification.
First, the number of near-optimal configurations is relatively low compared to the other two classes, leading to a class imbalance in the training set.
For example, when predicting whether an optimizer hyperparameter is large or small, only 130 out of 1690 configurations are ``optimal'' (given our current sampling granularity), while the remaining 1560 configurations fall into the other two classes.
Second, this imbalance will not disappear even if we increase the sampling granularity because, technically speaking, the ``optimal'' configurations form the set of critical points in the hyperparameter space and thus should have a lower dimensionality than the hyperparameter space itself.
That is, even if we increase the granularity by 100$\times$, the ``optimal'' configurations will only occupy an even lower ratio.
Therefore, we exclude these optimal configurations from the training and test sets and do not consider the third class in the problem.

\subsection{Collections of Trained Models}\label{sec:coll-model}
In our few-shot studies, we use models sampled from $\mathcal{F}$ as the \emph{training set} to train the \ourmethod classifier; then we use another model collection $\mathcal{F}^{\prime}$ as the test set.
$\mathcal{F}^{\prime}$ comprises models trained with the same varying factors, but each model is trained with 10\% label noise. 
Both $\mathcal{F}$ and $\mathcal{F}^{\prime}$ include various ResNet models trained on the CIFAR-10 dataset from scratch, and each differs in three factors:
1) the number of parameters ($p$) is varied by changing the width of ResNet-18 among \{2, 3, 4, 6, 8, 11, 16, 23, 32, 45, 64, 91, 128\}, 
2) the optimizer hyperparameter ($t$) is varied by changing the batch size among \{16, 24, 32, 44, 64, 92, 128, 180, 256, 364, 512, 724, 1024\},
3) the amount of data samples ($n$) is varied via subsampling the subset from CIFAR-10 training set between 10\% to 100\%, with increments of 10\%.
The combination of these three factors provides 1690 configurations, each comprising 5 runs with different random seeds.

\subsection{Setup of Dataset/Scale Transfer}\label{sec:case-study}
In the few-shot transfer learning setup, we detail the procedure for labeling \text{RFI} and $G$ for each sample in the training set.
The training/test set refers to the collections of pre-trained models. 
To label a training sample, additional models are required to compute \text{RFI} and $G$. 
We build the training set by sampling $\mathcal{F}$ in a structured manner. 
For example, to diagnose whether the optimizer hyperparameter of model $f_0$ is too large or too small, we sample $\mathcal{F}$ by fixing the parameter and data amount $(p, n)$. 
We then obtain all the possible optimizer hyperparameters within the range $[t_{\min}, t]$ and $[t, t_{\max}]$, where $t$ is the optimizer hyperparameter used to train $f_0$, and $t_{\min}$ and $t_{\max}$ are the minimum and maximum allowable values.
Using this sampling, we can label $\text{RFI}(m^{\uparrow}_t)$ using \eqref{eqn:RFI-t-large}, $\text{RFI}(m^{\downarrow}_t)$ using \eqref{eqn:RFI-t-small}, and compute $G = \text{RFI}(m^{\uparrow}_t) - \text{RFI}(m^{\downarrow}_t)$.
In all experiments, when we report the number of pre-trained models, that includes those used for labeling $G$ and $\text{RFI}$.\looseness-1

\section{Diagnosis Methods}\label{sec:app-diag} 

\subsection{Tree Construction}\label{sec:app-diag-tree}
In addition to Section~\ref{sec:diag-method}, we provide further details on constructing \ourmethod.
We prioritize connectivity over sharpness at the shallower levels of the tree due to its effectiveness in distinguishing dichotomous phenomena in the NN hyperparameter space.
Prior work \cite{yang2021taxonomizing, zhou2023three} has demonstrated that the trends of sharpness, similarity, and model performance, influenced by temperature-like parameters, can vary significantly depending on the connectivity value, even exhibiting opposite behaviors.
We treat sharpness and similarity as alternatives at the same level, using sharpness in the main experiments. An ablation study in Appendix~\ref{app:abl-cka} demonstrates that using similarity instead of sharpness provides comparable results.\looseness-1

For Q1 and Q2, we use different fixed subtree structures derived from the complete tree shown in Figure~\ref{fig:md-tree}. We ensure the tree structures for each task remain consistent across all evaluations. While the tree structures are fixed, the metric thresholds at each internal node are learned from the training set.
Future work could explore developing an algorithm that dynamically terminates the branching.

\subsection{Hyperparameters}\label{sec:app-diag-hyper}
\paragraph{Hyperparameters for Baseline Methods.}
Our baseline methods include \textbf{Hyperparameter} + DT, \textbf{Validation} + DT, and \textbf{Loss landscape} + DT.
We implement the standard DT method with the Gini impurity criterion and the best splitting strategy. 
We set the maximum depth of the tree to be 4 and the minimum number of samples required to split an internal node to be 2. 
We run all the methods for five runs with random seeds \{42, 90, 38, 18, 72\}.

\paragraph{Hyperparameters for \ourmethod.}
We provide the hyperparameters of our method in Table~\ref{tb:md-tree-hyper}. We run \ourmethod for five runs with random seeds \{42, 90, 38, 18, 72\}.
\begin{table}[!th]
\centering
\caption{Initial values and search ranges of loss landscape metric threshold used in \ourmethod.
}
\vspace{2mm}
\resizebox{0.65\linewidth}{!}{
\begin{tabular}{c|cccc}
\toprule
                     & Training error     & Connectivity  &  Sharpness   &  Similarity    \\
\midrule
 Initial value       & 0.5                &     -10        &         \{5, 7\}    &    0.5\\
 Bound               & [0, 1]             &   [-30, 0]     &     [4, 9]    &  [0.2, 0.8] \\
\bottomrule
\end{tabular}~\label{tb:md-tree-hyper}
} 
\end{table}

\vspace{-4mm}

\section{Broader Transfer Scenarios of \ourmethod}\label{sec:broader-transfer-sce}
In this section, we demonstrate the robustness of the proposed \ourmethod in diagnosing failure sources across three common scenarios. 
The first scenario involves out-of-distribution (OOD) generalization~\cite{hendrycks2018benchmarking}, where pre-trained models handle distribution shifts during testing. 
The second scenario considers class-imbalanced training, in which pre-trained models are initially trained on long-tailed datasets. 
The third scenario involves diagnosing unseen model architectures, specifically when the target models are Vision Transformer~\cite{dosovitskiy2021an}, while \ourmethod is trained on ResNet architecture models.\looseness-1

\subsection{Experimental Setup of Additional Transfer Scenarios}\label{sec:broader-transfer-sce-setup}
In the three additional studies, we focus on the first diagnostic task (Q1), which assesses whether the optimizer hyperparameter is large or small. 
The training set for \ourmethod in these studies remains consistent with the original setup described in Section~\ref{sec:eval-setup}, using pre-trained ResNet-18 models sampled from $\mathcal{F}$, trained on a class-balanced, in-distribution CIFAR-10 dataset.
The deviation from the original setup is the creation of new test sets to evaluate the diagnostic methods under three scenarios:\looseness-1

\begin{itemize}[noitemsep,topsep=0pt,leftmargin=*,after=,before=]
\item OOD generalization: test models are generated as in the original setup, sampled from $\mathcal{F}^{\prime}$, but are evaluated on CIFAR-10C~\cite{hendrycks2018benchmarking} to obtain corruption accuracy. Failure source labels are then annotated based on the corruption accuracy.
\item Class-imbalanced training: the test set consists of models trained on CIFAR-10-LT with an imbalance ratio of 10, evaluated using the standard CIFAR-10 test set. To get enough test models to evaluate \ourmethod, we varied model widths in the range \{4, 6, 8, 16, 32, 45, 64\} and training batch sizes within the range \{8, 16, 32, 64, 128, 256, 512, 1024\} in the test set. Each configuration was trained for 50 epochs using three random seeds.
\item Transformer architecture: the test set comprises ViT-tiny models trained on CIFAR-10, evaluated using the standard CIFAR-10 test set. We vary the model configuration by changing the embedding size over \{6, 12, 24, 48, 60, 96, 192\}, and changing batch size over \{8, 16, 32, 64, 128, 256, 364, 512\}. We train each configuration for 150 epochs using three random seeds.
\end{itemize}

\begin{figure*}[!th]
    \centering
    \includegraphics[width=0.7\linewidth,keepaspectratio]{figs/md_tree/temp_legend.pdf} \\
    \centering
    \begin{subfigure}{0.32\linewidth}
        \includegraphics[width=\linewidth]{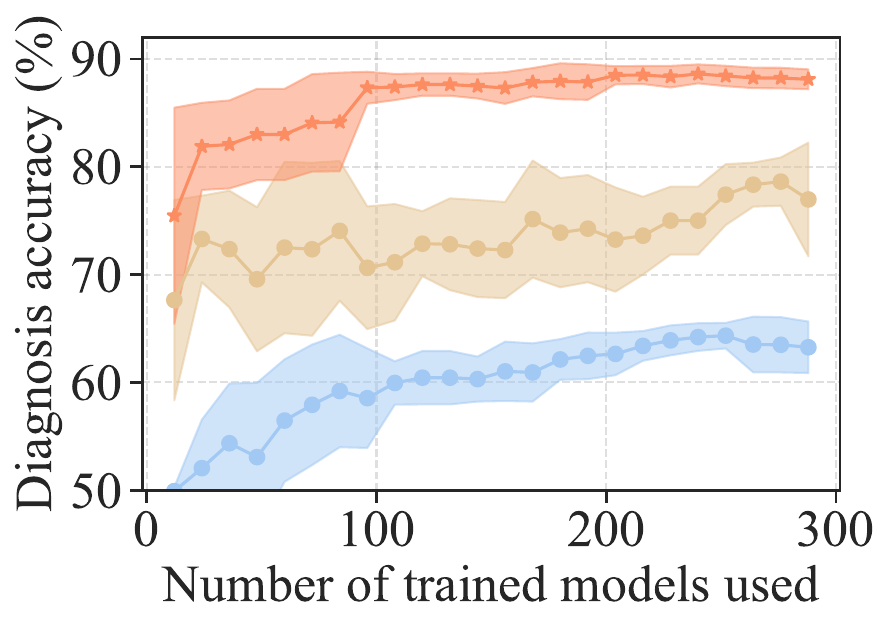}
        \caption{OOD generalization}
    \end{subfigure} 
    \centering
    \begin{subfigure}{0.32\linewidth}
        \includegraphics[width=\linewidth]{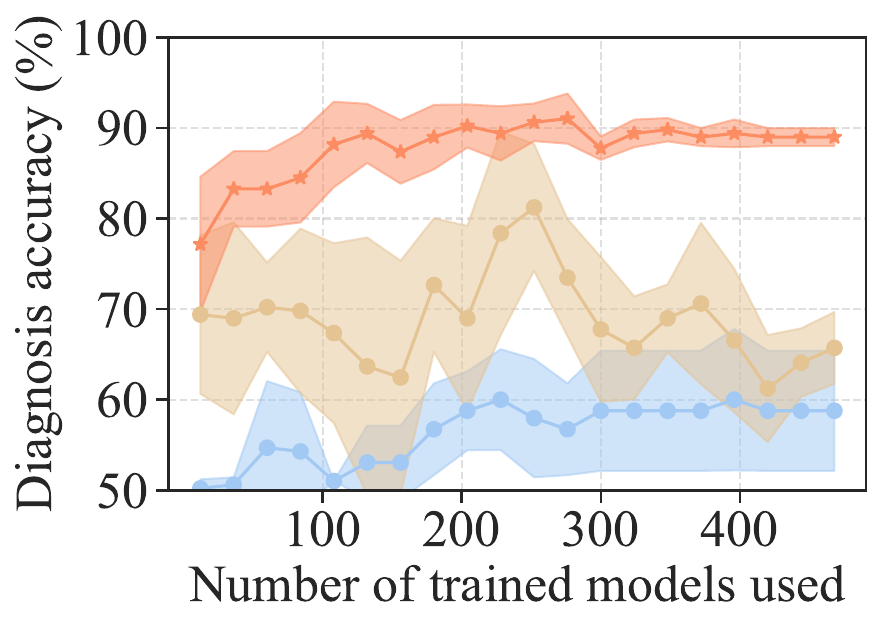}
        \caption{Class-imbalanced training}
    \end{subfigure} 
    \begin{subfigure}{0.32\linewidth}
        \includegraphics[width=\linewidth]{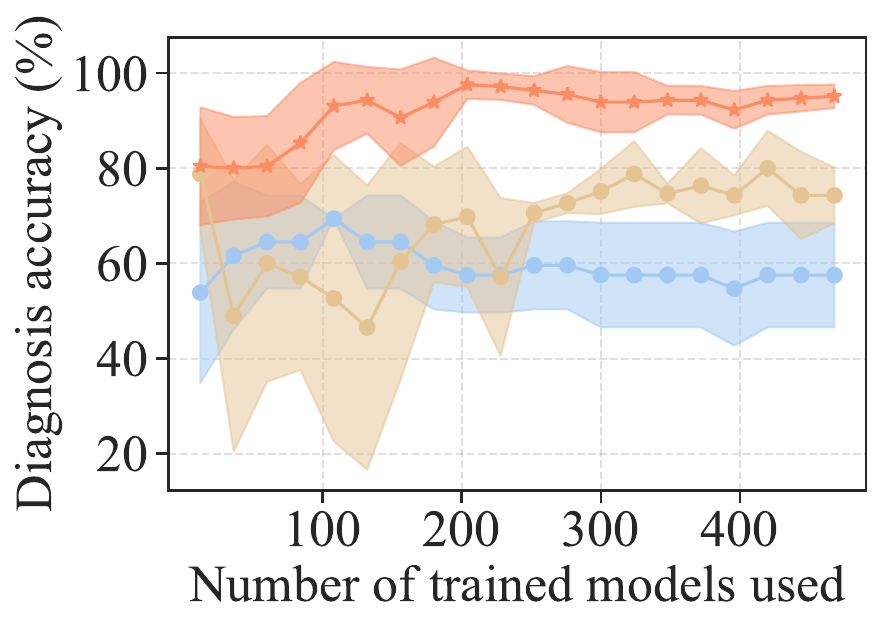}
        \caption{Transformer architecture}
    \end{subfigure} 
    \caption{\textbf{(Comparing \ourmethod to baseline methods in diagnosing trained models under three additional transfer scenarios).} The $y$-axis represents diagnosis accuracy, while the $x$-axis shows the number of pre-trained models used for the training set. (a) Models in the test set are evaluated on OOD CIFAR-10C test sets instead of the original ID test sets. (b) Models in the test set are trained on class-imbalanced CIFAR-10-LT datasets. (c) Models in the test set use transformer architectures.}\label{fig:transfer-case-acc}
\end{figure*}

\subsection{Empirical Evaluation}\label{sec:broader-transfer-sce-results}
\begin{figure}[!ht]
    \centering
    \includegraphics[width=0.7\linewidth,keepaspectratio]{figs/md_tree/temp_legend.pdf} \\
    \begin{subfigure}{0.32\linewidth}
    \includegraphics[width=\linewidth,keepaspectratio]{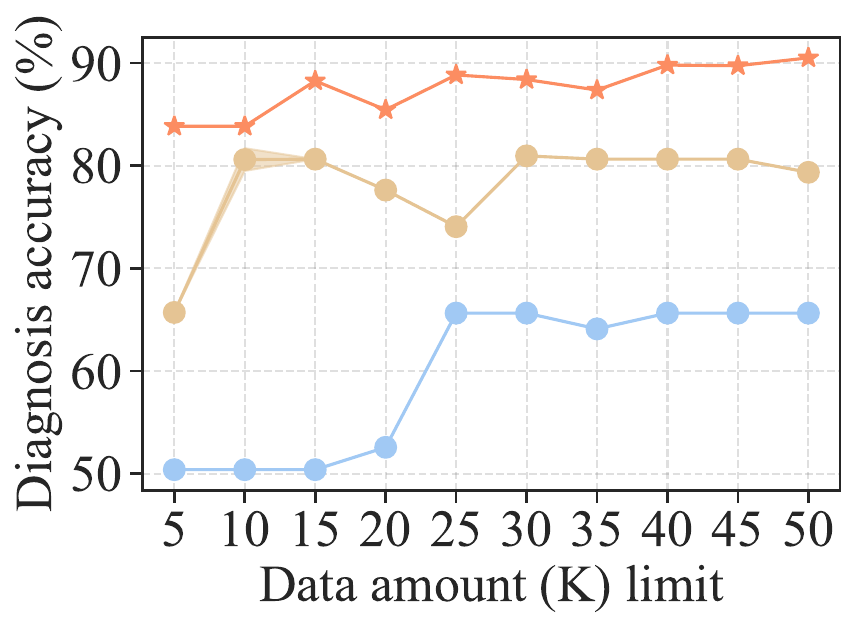} 
    \caption{Scale transfer}
    \end{subfigure}  
    \begin{subfigure}{0.32\linewidth}
    \includegraphics[width=\linewidth,keepaspectratio]{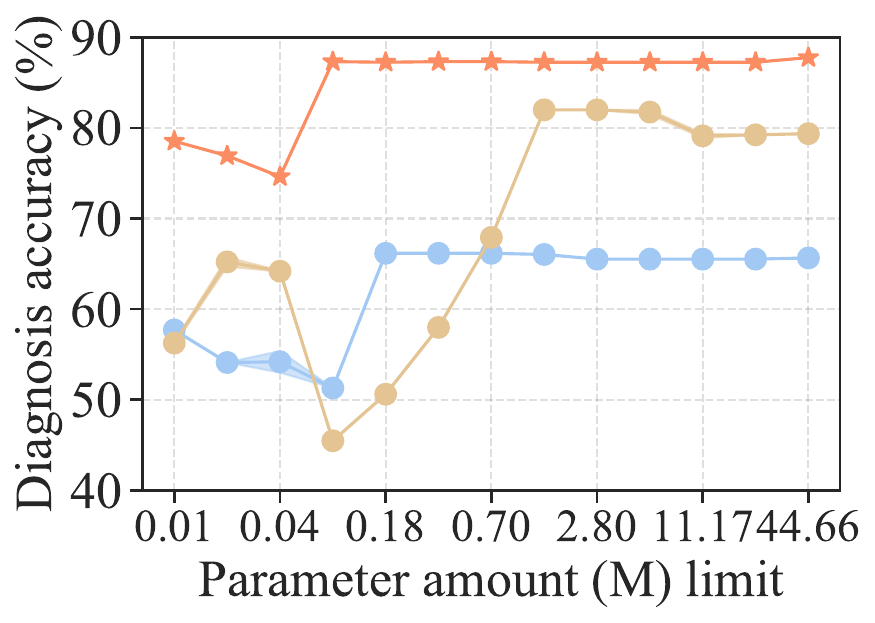} 
    \caption{Scale transfer}
    \end{subfigure}  
    \caption{\textbf{(Diagnosing trained models tested on out-of-distribution data (CIFAR-10-C)).}
    The $y$-axis indicates the diagnosis accuracy.
    (a) The $x$-axis indicates the maximum amount of training (image) data for training models in the training set.
    (b) The $x$-axis indicates the maximum number of parameters of the models in the training set.
  }~\label{fig:ood-scale-transfer}  \vspace{-8mm}
\end{figure}

We compared \ourmethod to two baseline methods across three transfer scenarios. Figure~\ref{fig:transfer-case-acc} and Figure~\ref{fig:ood-scale-transfer} show that \ourmethod consistently outperforms the baselines, demonstrating its robust generalizability. 
Visualizations interpreting these methods are available in Appendix~\ref{sec:broader-transfer-sce-vis}. 
Our observations indicate that in challenging transfer scenarios, validation-based methods perform worse due to increased distribution shifts between the training set's pre-trained models and the new test sets. 
For example, in the class-imbalanced training scenario, the validation errors of CIFAR-10-LT models differ markedly from those of CIFAR-10 models, causing decision trees built on this feature to fail in generalizing.
However, the loss landscape regimes remain transferable, providing interpretable diagnostics.

\subsection{Visualization}\label{sec:broader-transfer-sce-vis}
We present additional results of the studies on diagnosis under the three transfer scenarios, with the experimental setup shown in Section~\ref{sec:broader-transfer-sce-setup}.\vspace{-3mm}

\begin{figure}[!h]
    \centering
    \vspace{-2mm}
    \begin{subfigure}{0.25\linewidth}
        \includegraphics[width=\linewidth]{figs/md_tree/temp/zero_shot_legend.pdf}
    \end{subfigure} \\
    \begin{subfigure}{0.25\linewidth}
        \includegraphics[width=\linewidth]{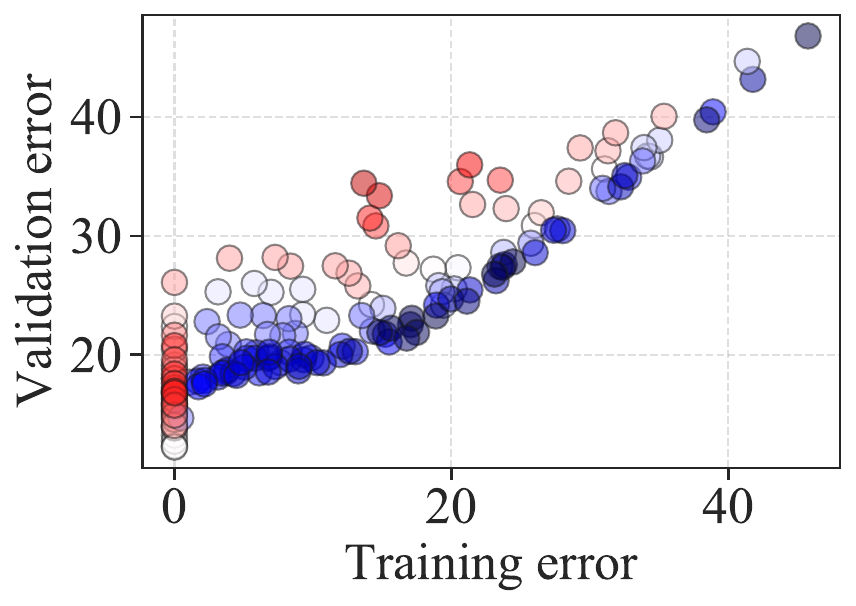}
        \caption{Training (CIFAR-10)}
    \end{subfigure}  
    \begin{subfigure}{0.25\linewidth}
        \includegraphics[width=\linewidth]{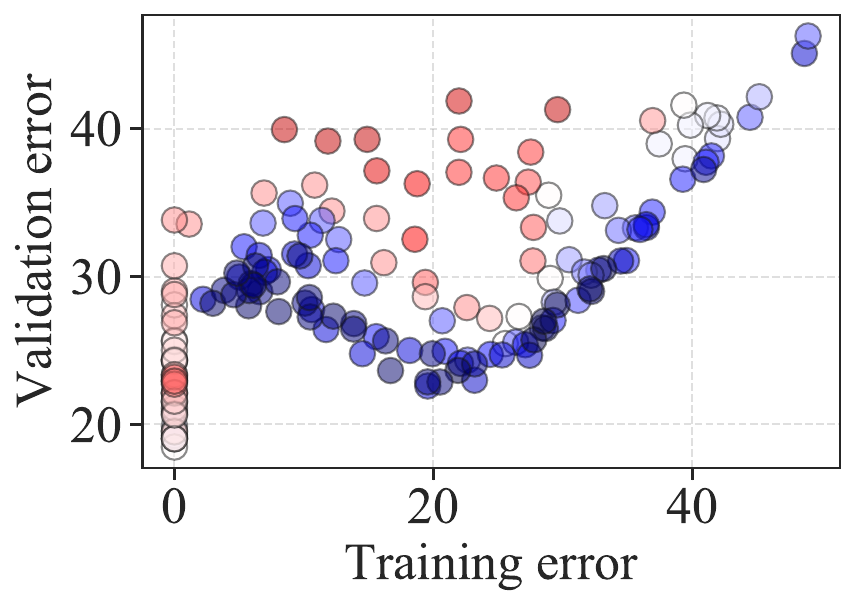}
        \caption{Test (CIFAR-10-C)}
    \end{subfigure}  \vspace{-3mm}
    \caption{\textbf{(Applying validation metrics to diagnose the models tested on out-of-distribution data (CIFAR-10-C)).} (a) models in the training set are evaluated on ID CIFAR-10 test data, (b) models in the test set are evaluated on OOD CIFAR-10-C test data.} \label{fig:ood-temp-vis-baseline}  
\end{figure}

\begin{figure*}[!th]
    \centering
    \begin{minipage}[b]{0.48\linewidth}
        \includegraphics[width=\linewidth]{figs/md_tree/structure/temp_increase_decision_tree_without_regime.pdf}
        \vspace{5mm}
        \subcaption{MD tree}\label{fig:ood-vis-zero-shot-temp-tree}
    \end{minipage}
    \begin{minipage}[b]{0.48\linewidth}
    \centering
    \begin{subfigure}{0.52\linewidth}
        \includegraphics[width=\linewidth]{figs/md_tree/temp/zero_shot_legend.pdf}
    \end{subfigure} 
    \begin{subfigure}{0.46\linewidth}
        \includegraphics[width=\linewidth]{figs/md_tree/decision_bd.pdf}
    \end{subfigure}
    \\
    \begin{subfigure}{0.48\linewidth}
        \includegraphics[width=\linewidth]{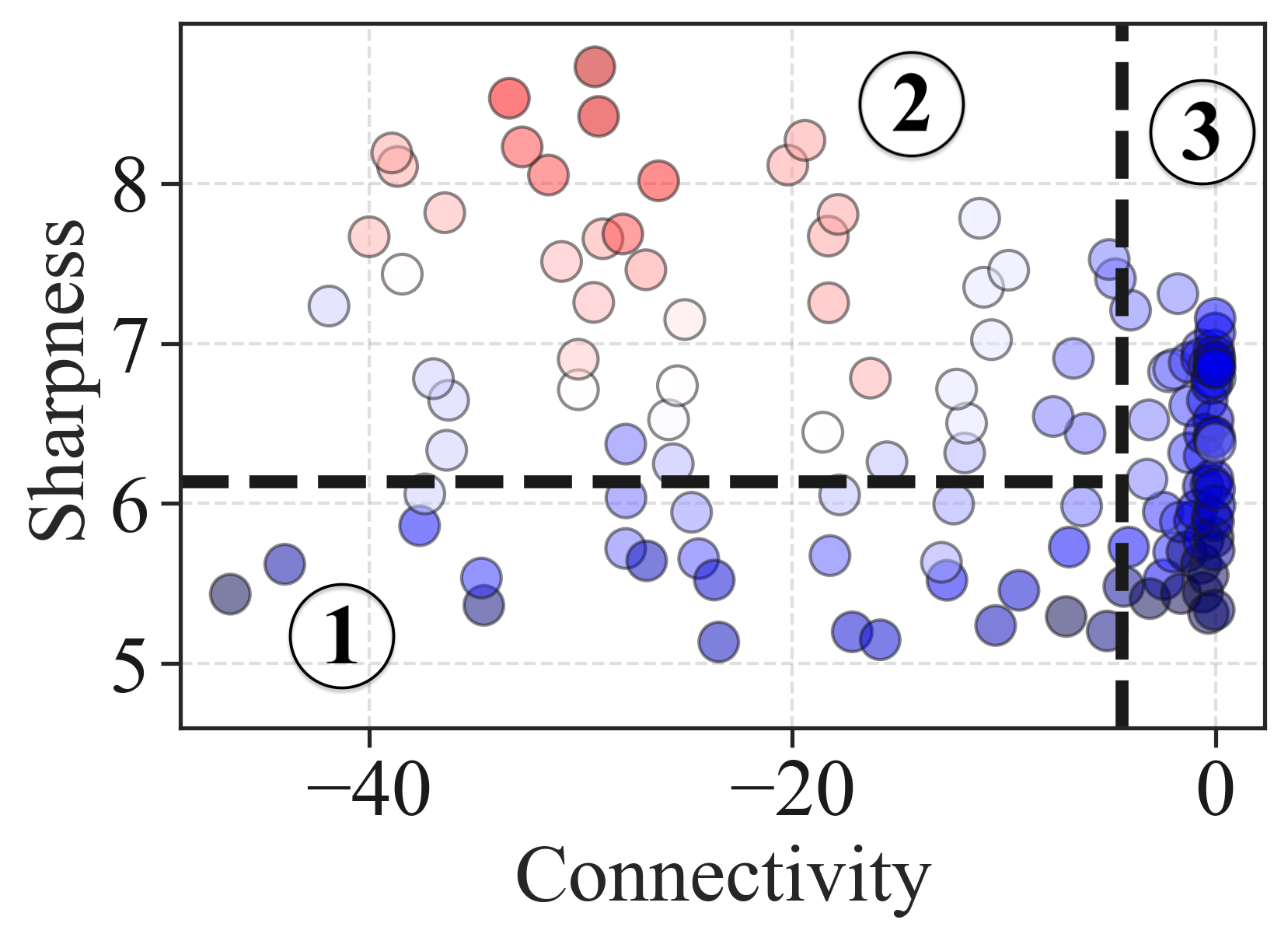}
        \caption{Training (CIFAR-10)} \label{fig:ood-vis-b} 
    \end{subfigure} 
    \begin{subfigure}{0.48\linewidth}
        \includegraphics[width=\linewidth]{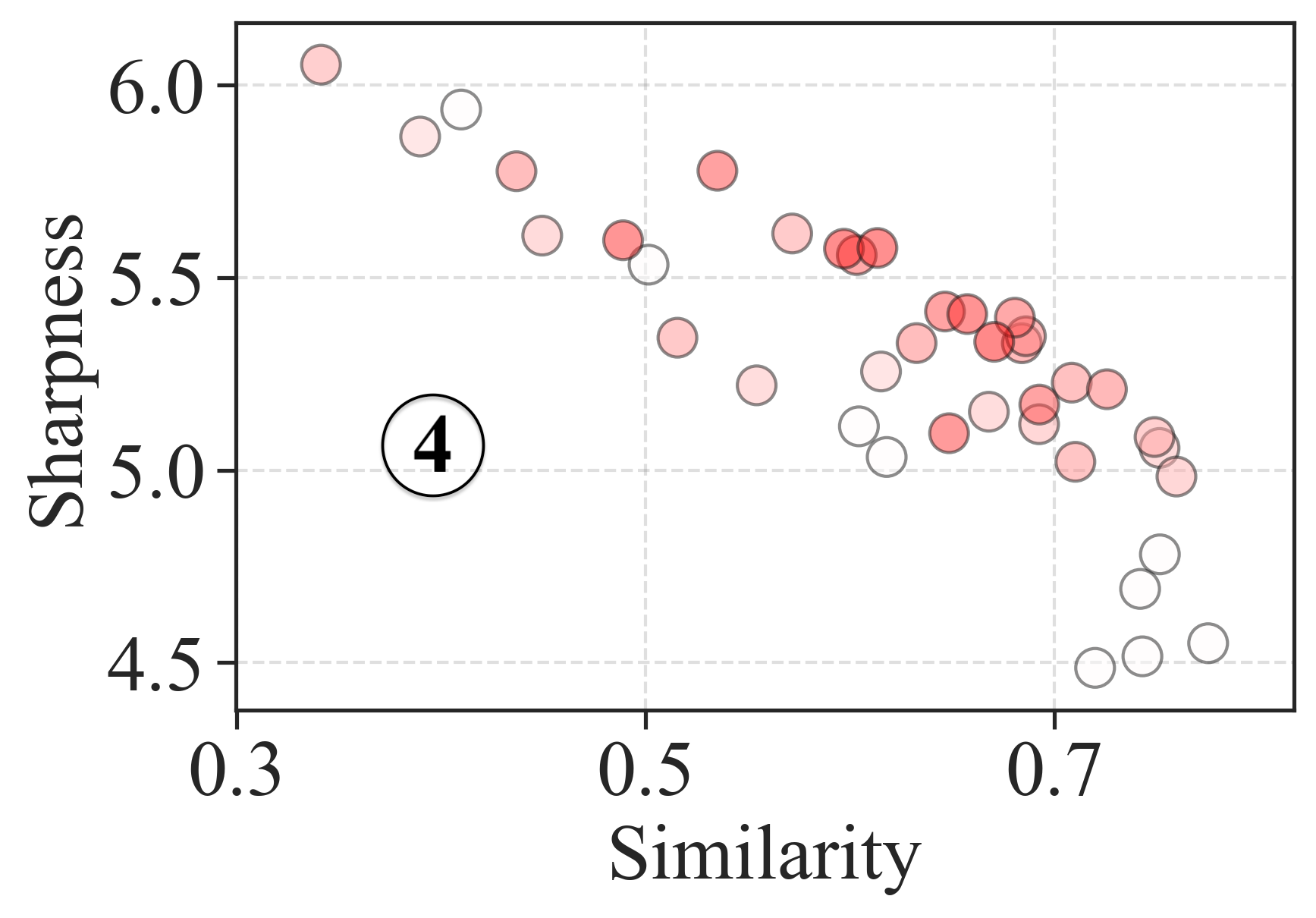}
        \caption{Training (CIFAR-10)}  \label{fig:ood-vis-c} 
    \end{subfigure} \\
    \begin{subfigure}{0.48\linewidth}
        \includegraphics[width=\linewidth]{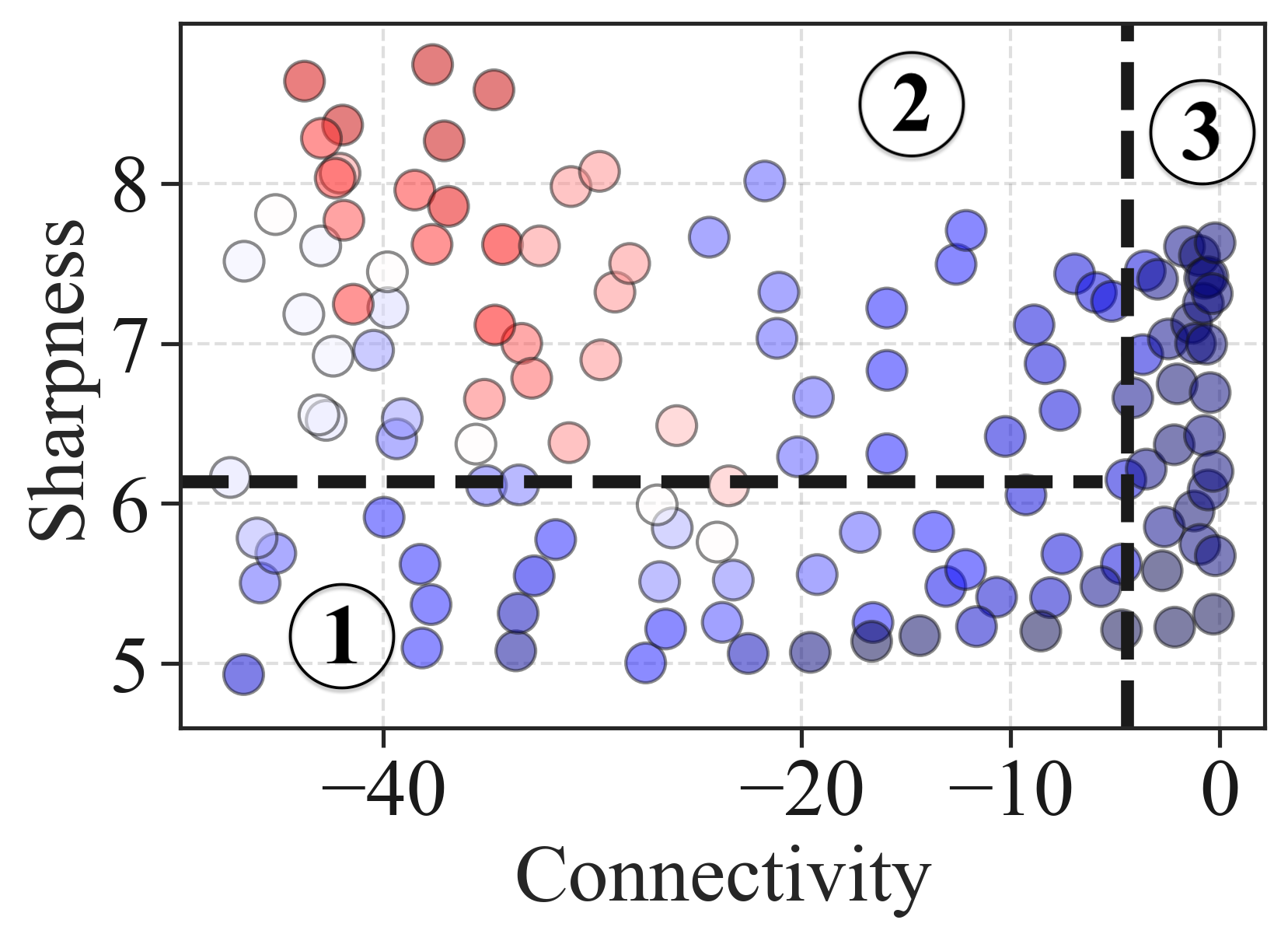}
        \caption{Test (CIFAR-10-C)  } \label{fig:ood-vis-d} 
    \end{subfigure} 
    \begin{subfigure}{0.48\linewidth}
        \includegraphics[width=\linewidth]{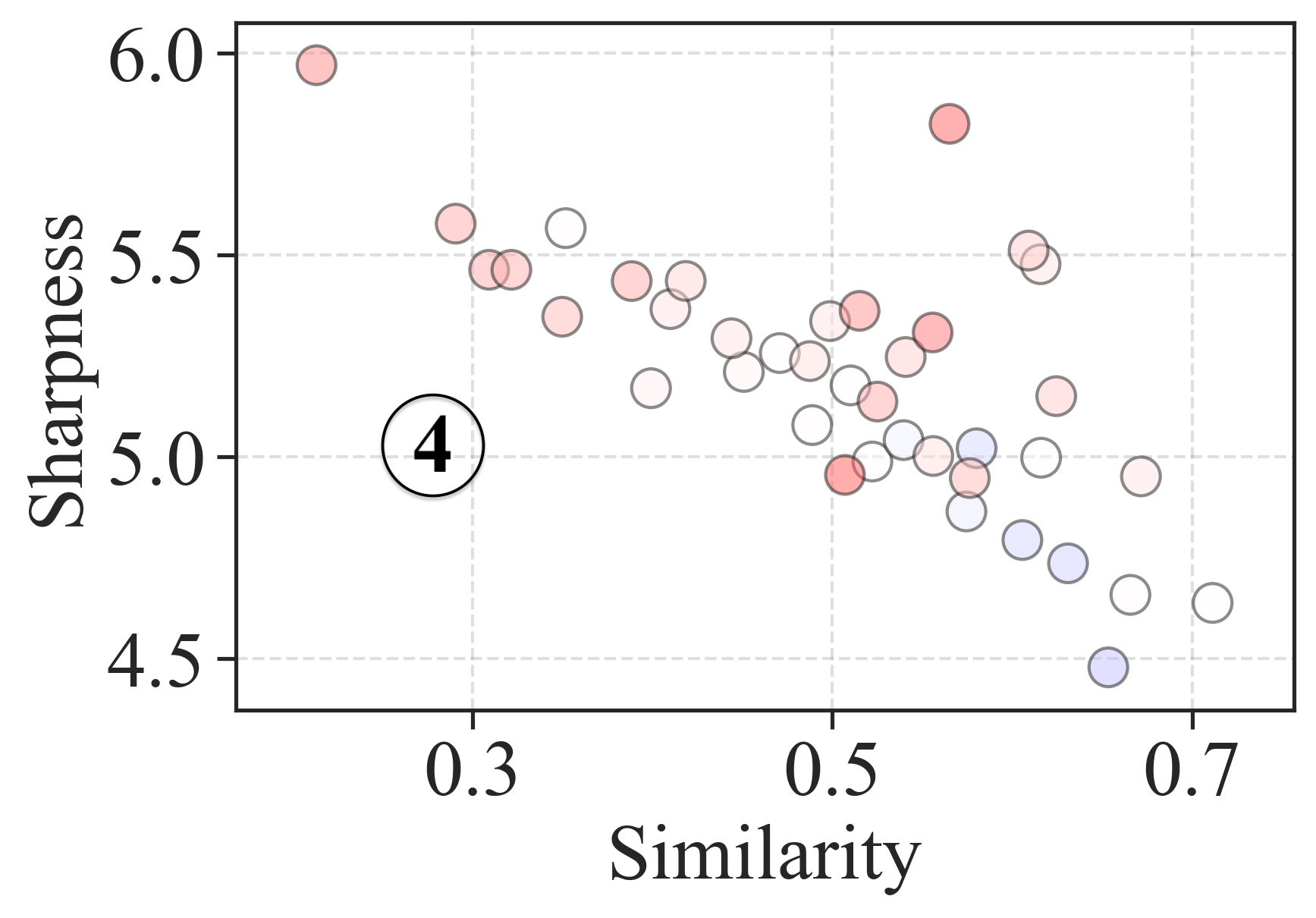}
        \caption{Test (CIFAR-10-C)} \label{fig:ood-vis-e} 
    \end{subfigure} 
    \end{minipage} \vspace{-3mm}
    \caption{\textbf{(Visualizing MD tree and its diagnosis results on models tested on out-of-distribution data (CIFAR-10-C)).}
    \emph{Left side}: structure of the MD tree. The color of the leaf node indicates the predicted class by \ourmethod.
    The threshold values are learned from the training set.
    \emph{Right side}: 
    The first row represents training samples, and the second row represents test samples.
    Each colored circle represents one sample (which is one pre-trained model configuration), and the color represents the ground-truth label: blue means the hyperparameter is too large, while red means small.
    The black dashed line indicates the decision boundary of \ourmethod.
    Each numbered regime on the right corresponds to the leaf node with the same number on the tree.
    The samples in~\ref{fig:ood-vis-b} and those in~\ref{fig:ood-vis-c} are separated by training error. The same applies to~\ref{fig:ood-vis-d} and~\ref{fig:ood-vis-e}.
    }\label{fig:ood-temp-vis-md}   \vspace{-2mm} 
\end{figure*}

\paragraph{OOD Generalization.} Figure~\ref{fig:ood-scale-transfer} illustrates the results of combining scale transfer and OOD generalization, confirming the effectiveness of the proposed methods. 
The visualizations of the validation-based method and our method are shown in Figures~\ref{fig:ood-temp-vis-baseline} and~\ref{fig:ood-temp-vis-md}, respectively.
Figure~\ref{fig:ood-temp-vis-baseline} highlights the issues associated with using validation metrics, showing a significant distribution shift between the training and test sets. 
Figure~\ref{fig:ood-vis-zero-shot-temp-tree} presents the \ourmethod structure and threshold.
Comparing Figures~\ref{fig:ood-vis-b} and~\ref{fig:ood-vis-d}, we observe that pre-trained models generally exhibit similar ID and OOD failure sources, which are effectively classified by our method's decision boundary. 
This observation aligns with the findings of \citet{miller2021accuracy}, which indicate a positive correlation between ID and OOD performance, suggesting that models often share the same ID and OOD failure sources.

\begin{figure}[!h]
    \centering
    \begin{subfigure}{0.25\linewidth}
        \includegraphics[width=\linewidth]{figs/md_tree/temp/zero_shot_legend.pdf}
    \end{subfigure} \\
    \begin{subfigure}{0.25\linewidth}
        \includegraphics[width=\linewidth]{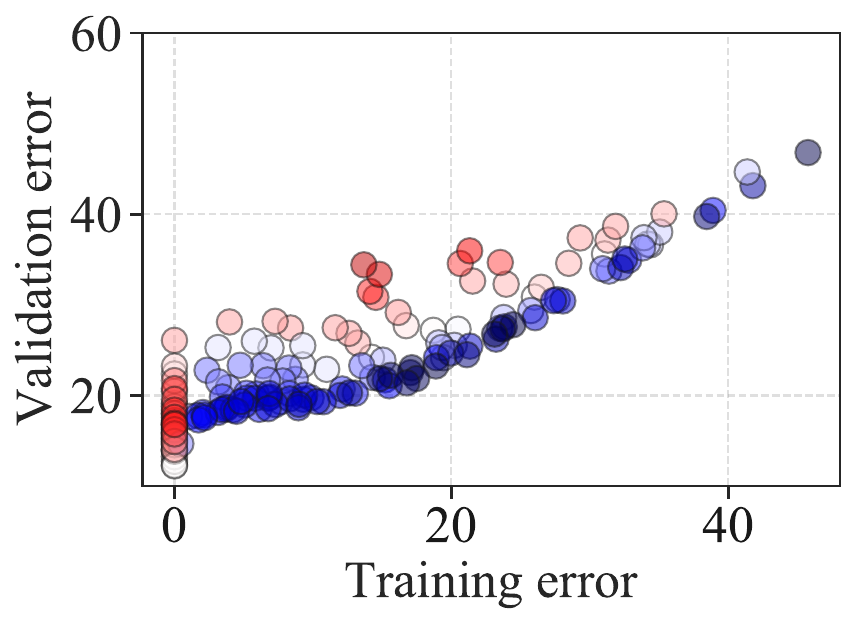}
        \caption{Training (CIFAR-10)}
    \end{subfigure}  
    \begin{subfigure}{0.25\linewidth}
        \includegraphics[width=\linewidth]{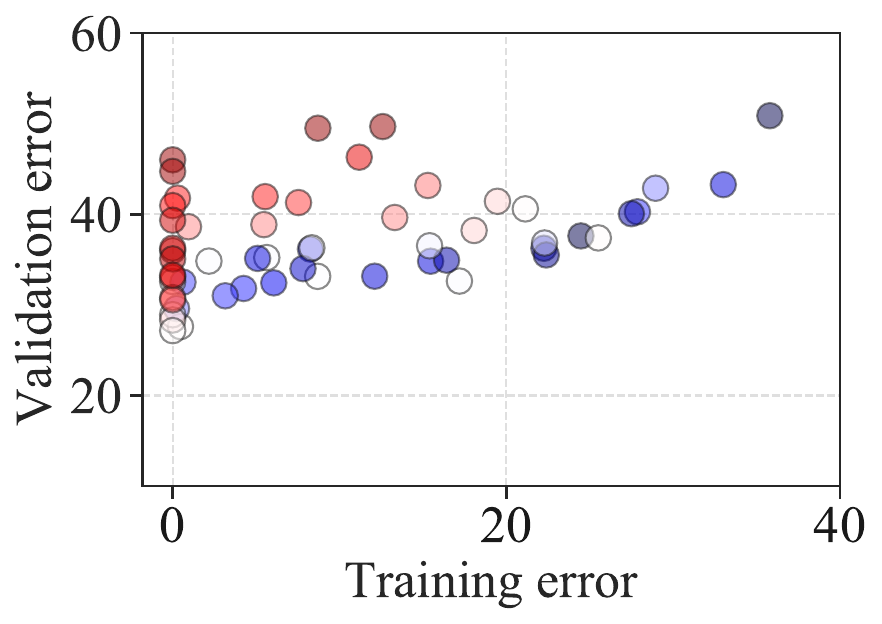}
        \caption{Test (CIFAR-10-LT)}
    \end{subfigure} \vspace{-3mm}
    \caption{\textbf{(Applying validation metrics to diagnose the models trained with imbalanced data (CIFAR-10-LT)).} (a) training set comprises ResNet18 models trained with CIFAR-10, (b) test set comprises ResNet18 models trained with CIFAR-10-LT.}~\label{fig:rebut-lt-val-vis}  
\end{figure}

\begin{figure*}[!th]
    \centering
    \begin{minipage}[b]{0.48\linewidth}
        \includegraphics[width=\linewidth]{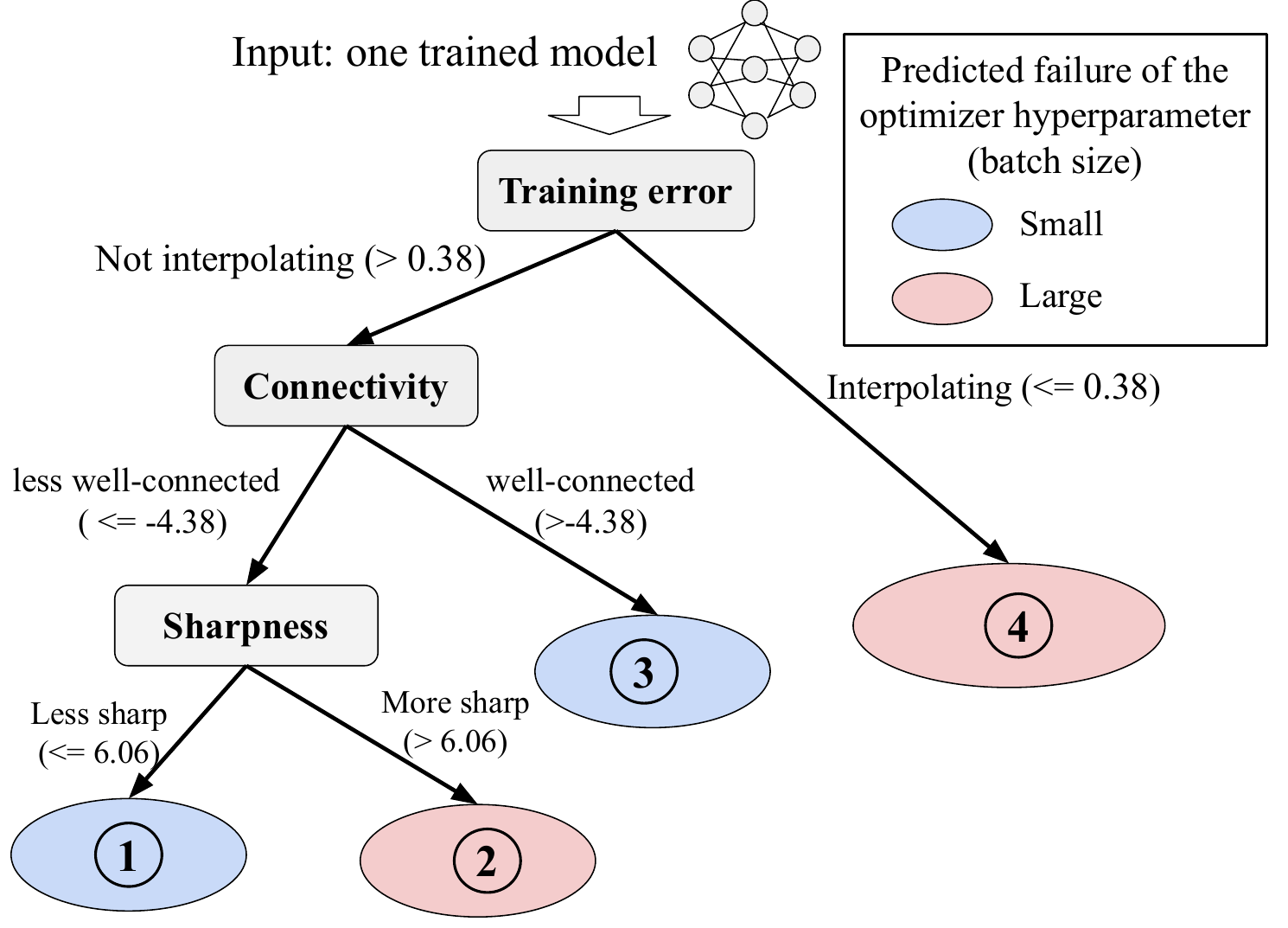}
        \vspace{5mm}
        \subcaption{MD tree}
    \end{minipage}
    \begin{minipage}[b]{0.48\linewidth}
    \centering
    \begin{subfigure}{0.52\linewidth}
        \includegraphics[width=\linewidth]{figs/md_tree/temp/zero_shot_legend.pdf}
    \end{subfigure} 
    \begin{subfigure}{0.46\linewidth}
        \includegraphics[width=\linewidth]{figs/md_tree/decision_bd.pdf}
    \end{subfigure}
    \\
    \begin{subfigure}{0.48\linewidth}
        \includegraphics[width=\linewidth]{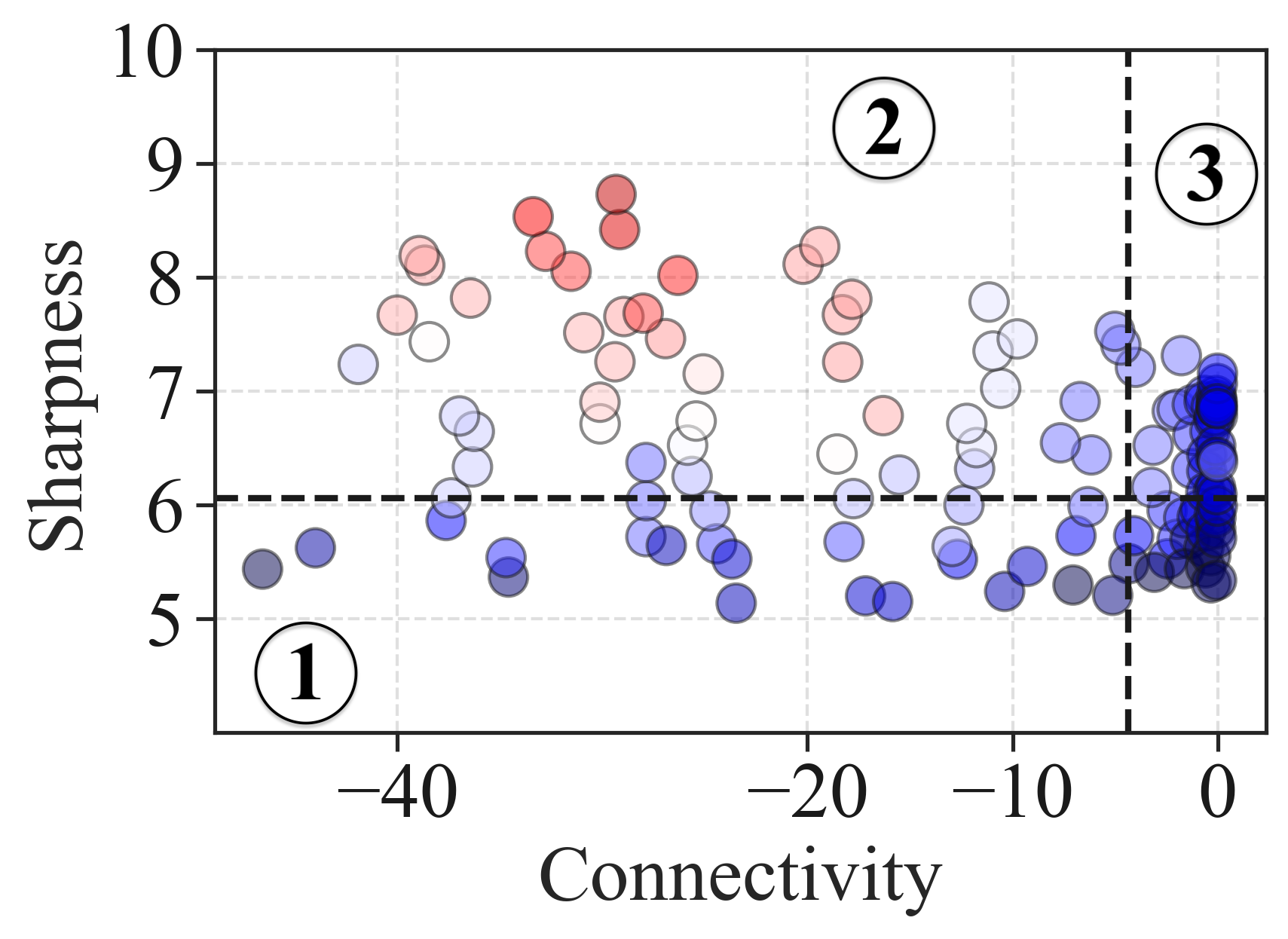}
        \caption{Training (CIFAR-10)} \label{fig:lt-vis-b} 
    \end{subfigure} 
    \begin{subfigure}{0.48\linewidth}
        \includegraphics[width=\linewidth]{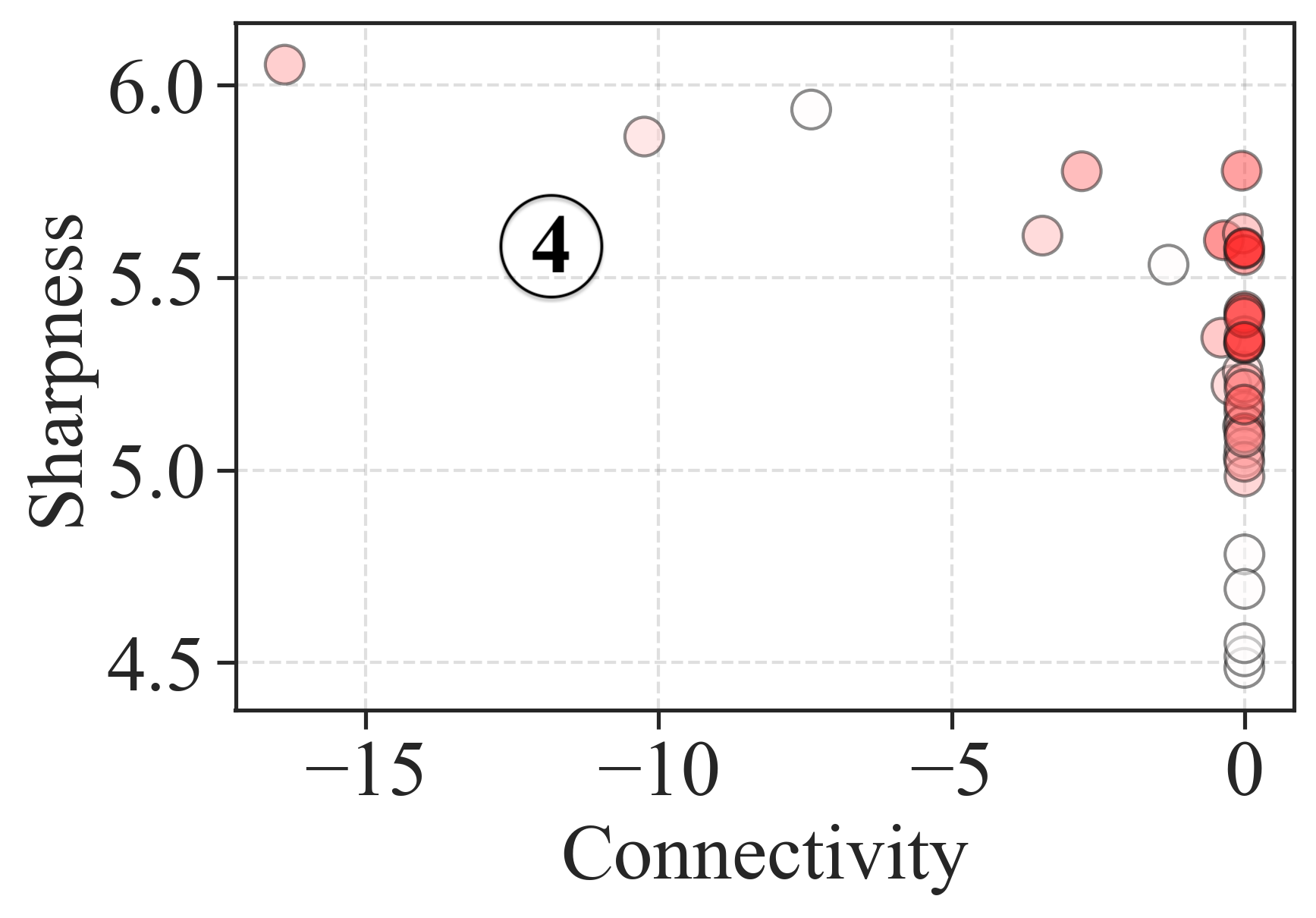}
        \caption{Training (CIFAR-10)}  \label{fig:lt-vis-c} 
    \end{subfigure} \\
    \begin{subfigure}{0.48\linewidth}
        \includegraphics[width=\linewidth]{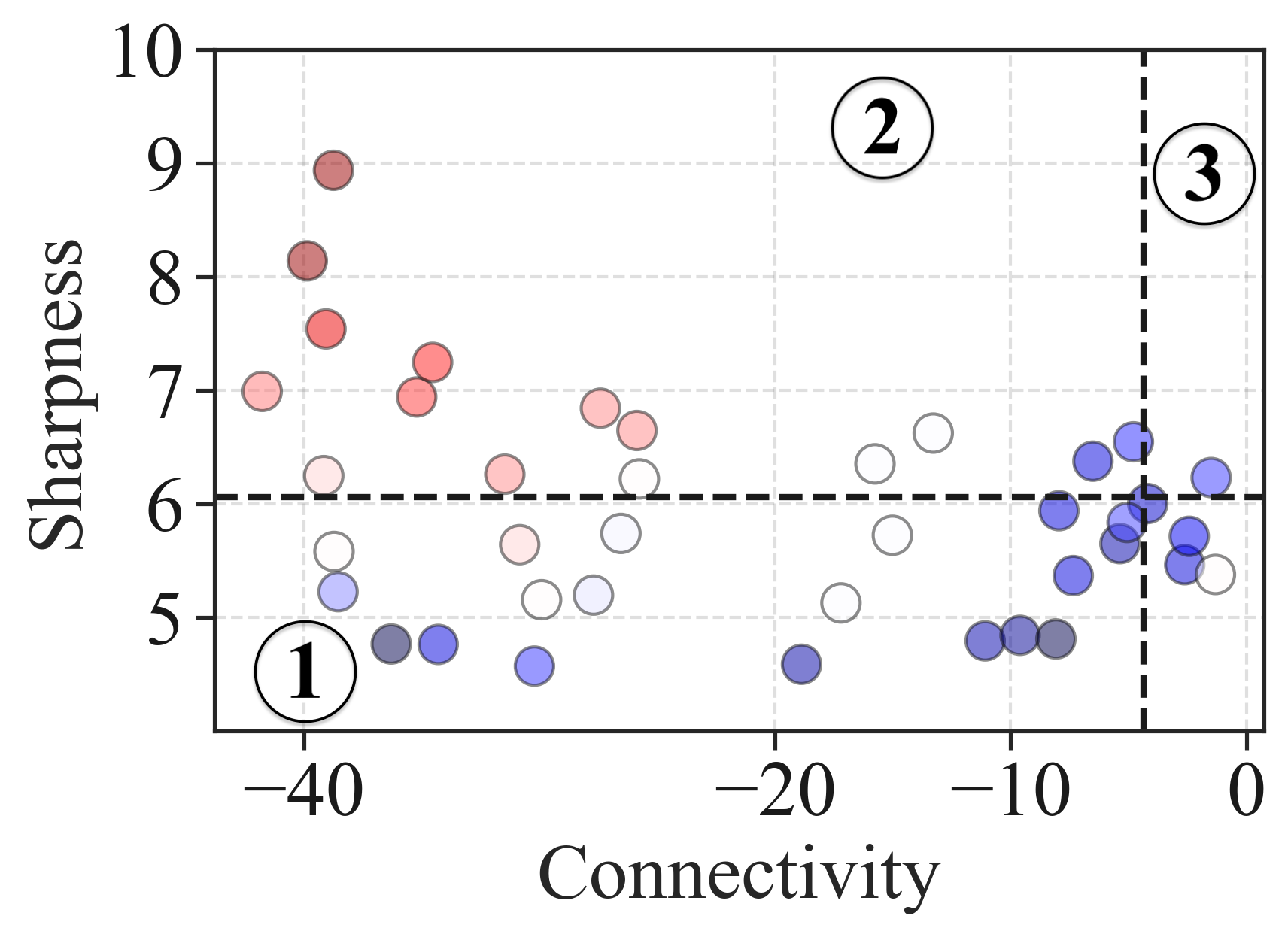}
        \caption{Test (CIFAR-10-LT)  } \label{fig:lt-vis-d} 
    \end{subfigure} 
    \begin{subfigure}{0.48\linewidth}
        \includegraphics[width=\linewidth]{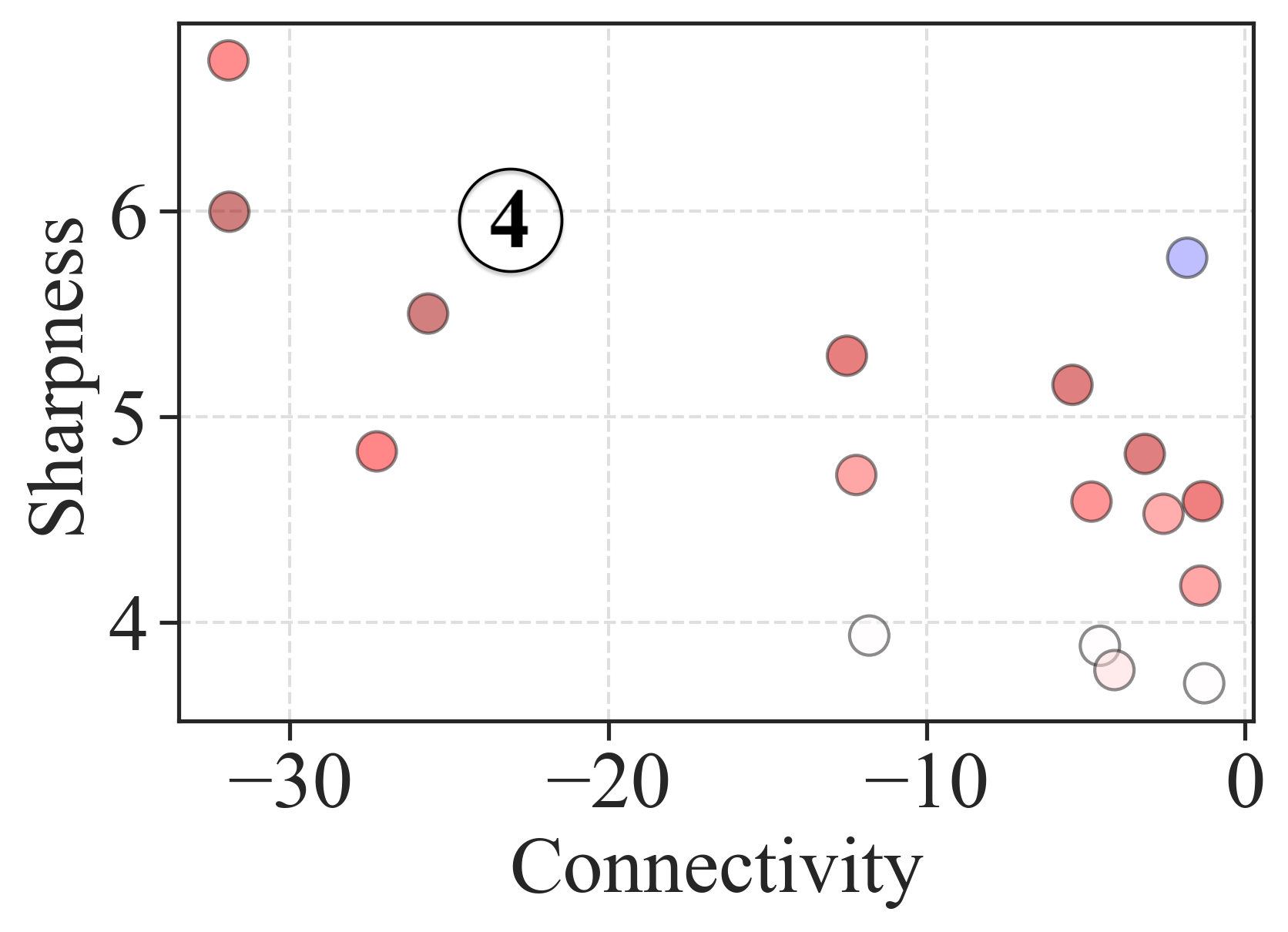}
        \caption{Test (CIFAR-10-LT)} \label{fig:lt-vis-e} 
    \end{subfigure} 
    \end{minipage} \vspace{-2mm}
    \caption{\textbf{(Visualizing MD tree and its diagnosis results on models trained on imbalanced data (CIFAR-10-LT)).}
    \emph{Left side}: structure of the MD tree. The color of the leaf node indicates the predicted class by \ourmethod.
    The threshold values are learned from the training set.
    \emph{Right side}: 
    The first row represents training samples, and the second row represents test samples.
    Each colored circle represents one sample (which is one pre-trained model configuration), and the color represents the ground-truth label: blue means the hyperparameter is too large, while red means small.
    The black dashed line indicates the decision boundary of \ourmethod.
    Each numbered regime on the right corresponds to the leaf node with the same number on the tree.
    The samples in~\ref{fig:lt-vis-b} and those in~\ref{fig:lt-vis-c} are separated by training error. The same applies to~\ref{fig:lt-vis-d} and~\ref{fig:lt-vis-e}.
    }~\label{fig:rebut-lt-md-vis}   \vspace{-4mm}
\end{figure*}

\FloatBarrier
\paragraph{Class-imbalanced Training.}
This subsection presents the visualization results for the class-imbalanced training scenario. 
Figure~\ref{fig:rebut-lt-val-vis} illustrates the substantial distribution shift in validation metrics between CIFAR-10 and CIFAR-10-LT models. 
Notably, the validation error for CIFAR-10-LT models is predominantly above 25\%, whereas most CIFAR-10 models exhibit validation errors below this threshold.
Figure~\ref{fig:rebut-lt-md-vis} demonstrates that the decision boundaries learned by the MD tree from CIFAR-10 models effectively transfer to CIFAR-10-LT models, highlighting the robustness of the MD tree's loss landscape metrics in handling class imbalance.

\vspace{-5mm}

\FloatBarrier

\begin{figure}[!h]
    \centering
    \begin{subfigure}{0.25\linewidth}
        \includegraphics[width=\linewidth]{figs/md_tree/temp/zero_shot_legend.pdf}
    \end{subfigure} \\
    \begin{subfigure}{0.25\linewidth}
        \includegraphics[width=\linewidth]{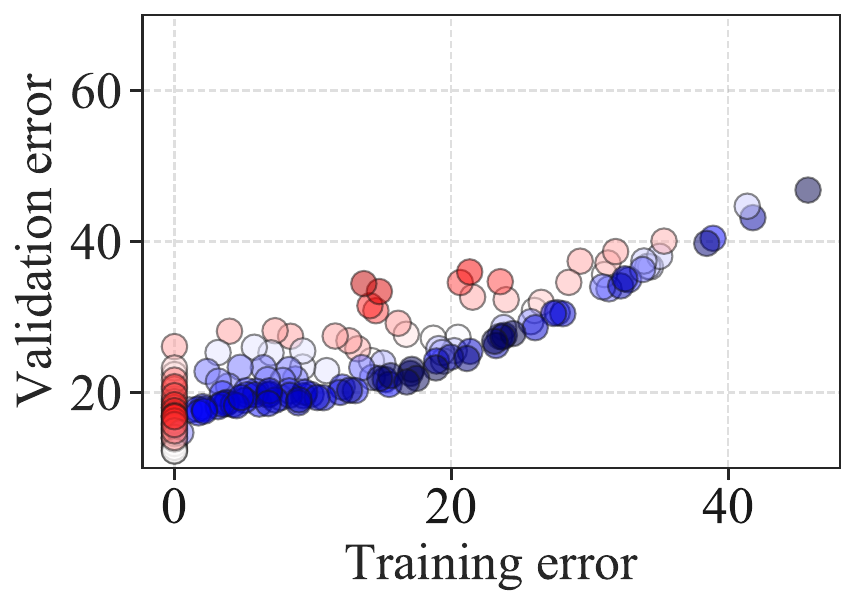}
        \caption{Training (ResNet18)}
    \end{subfigure}  
    \begin{subfigure}{0.25\linewidth}
        \includegraphics[width=\linewidth]{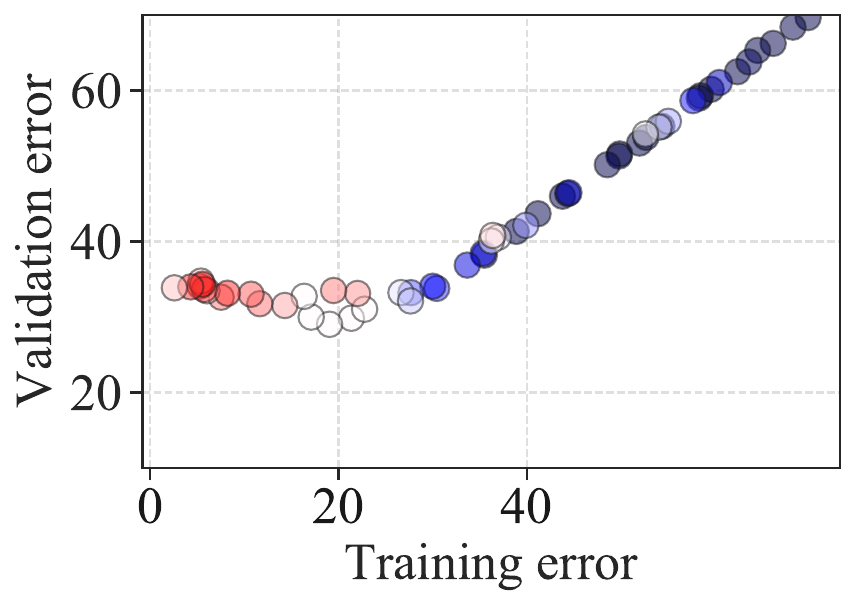}
        \caption{Test (ViT)}
    \end{subfigure} \vspace{-2mm}
    \caption{\textbf{(Applying validation metrics to diagnose the ViT-tiny models trained with CIFAR-10).} (a) training set comprises ResNet18 models trained on CIFAR-10, (b) test set comprises ViT-tiny models trained on CIFAR-10.}~\label{fig:rebut-vit-val-vis}
\end{figure}

\paragraph{Transformer Architecture.} Figure~\ref{fig:rebut-vit-val-vis} shows a significant shift in the training and validation error ranges between ResNet18 models in the training set and Vision Transformer (ViT) models in the test set.
This shift leads to poor performance when transferring diagnostic results from ResNet18 models to ViT models based on validation metrics.
This discrepancy aligns with findings by \citet{raghu2021vision} and \citet{d2021convit}, which highlight the challenges of training ViT models from scratch on small-scale datasets, resulting in high training and validation errors. 
In contrast, convolutional architectures, such as ResNet18, train efficiently on small datasets and exhibit low errors.

\begin{figure*}[!th]
    \centering
    \begin{minipage}[b]{0.48\linewidth}
        \includegraphics[width=\linewidth]{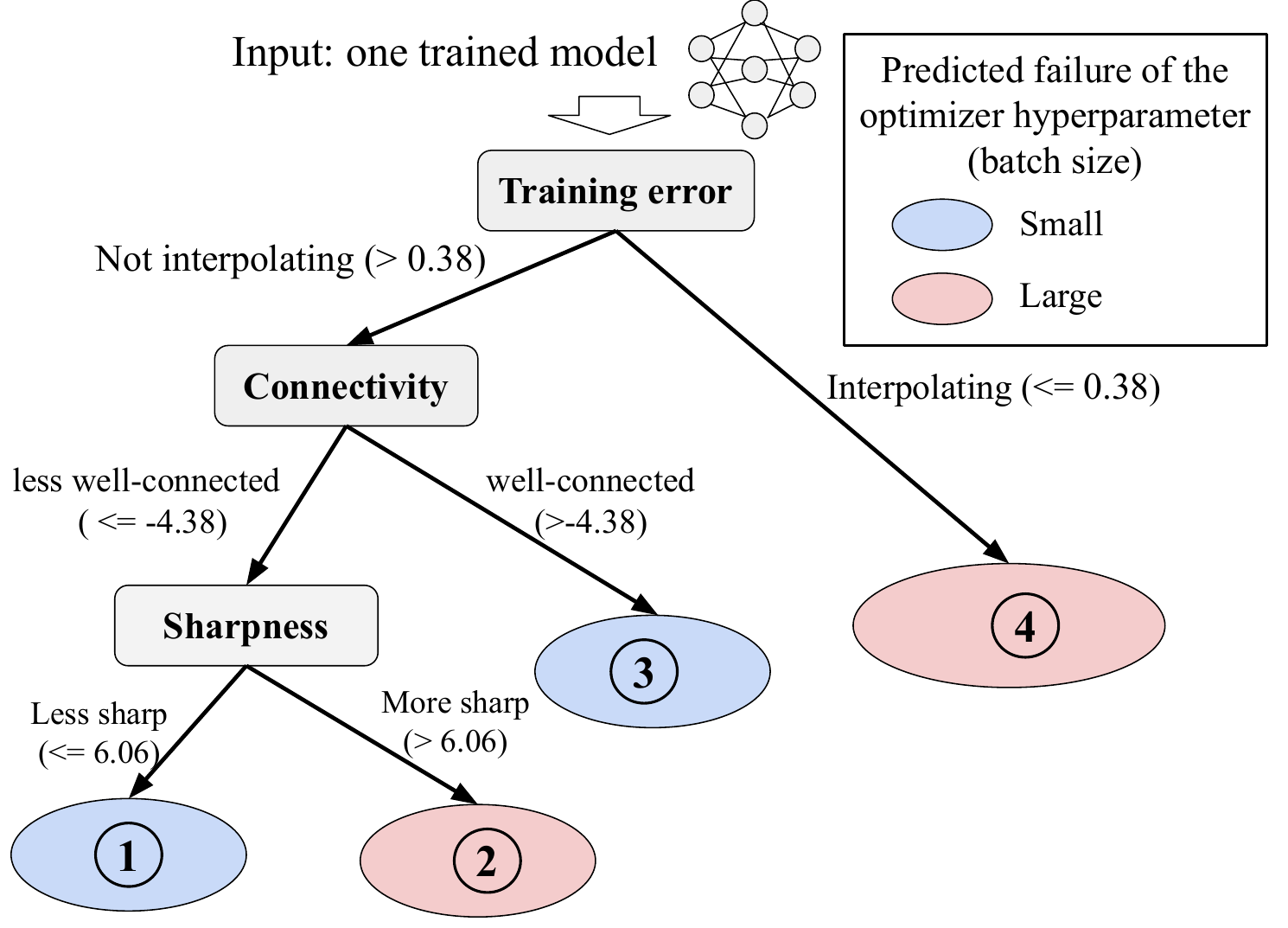}
        \vspace{5mm}
        \subcaption{MD tree}
    \end{minipage}
    \begin{minipage}[b]{0.48\linewidth}
    \centering
    \begin{subfigure}{0.52\linewidth}
        \includegraphics[width=\linewidth]{figs/md_tree/temp/zero_shot_legend.pdf}
    \end{subfigure} 
    \begin{subfigure}{0.46\linewidth}
        \includegraphics[width=\linewidth]{figs/md_tree/decision_bd.pdf}
    \end{subfigure}
    \\
    \begin{subfigure}{0.48\linewidth}
        \includegraphics[width=\linewidth]{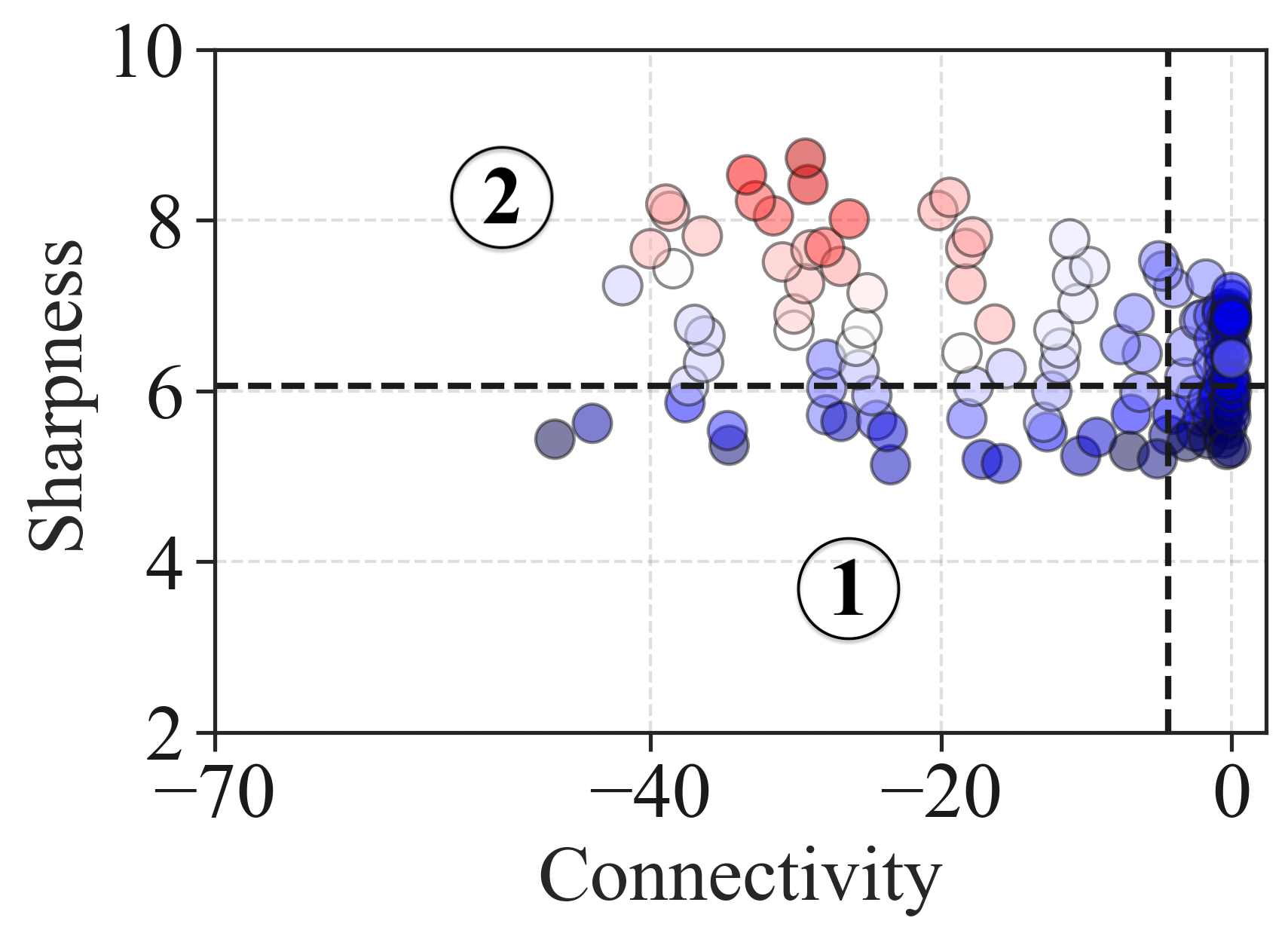}
        \caption{Training (ResNet18)} \label{fig:vit-vis-b} 
    \end{subfigure} \\
    \begin{subfigure}{0.48\linewidth}
        \includegraphics[width=\linewidth]{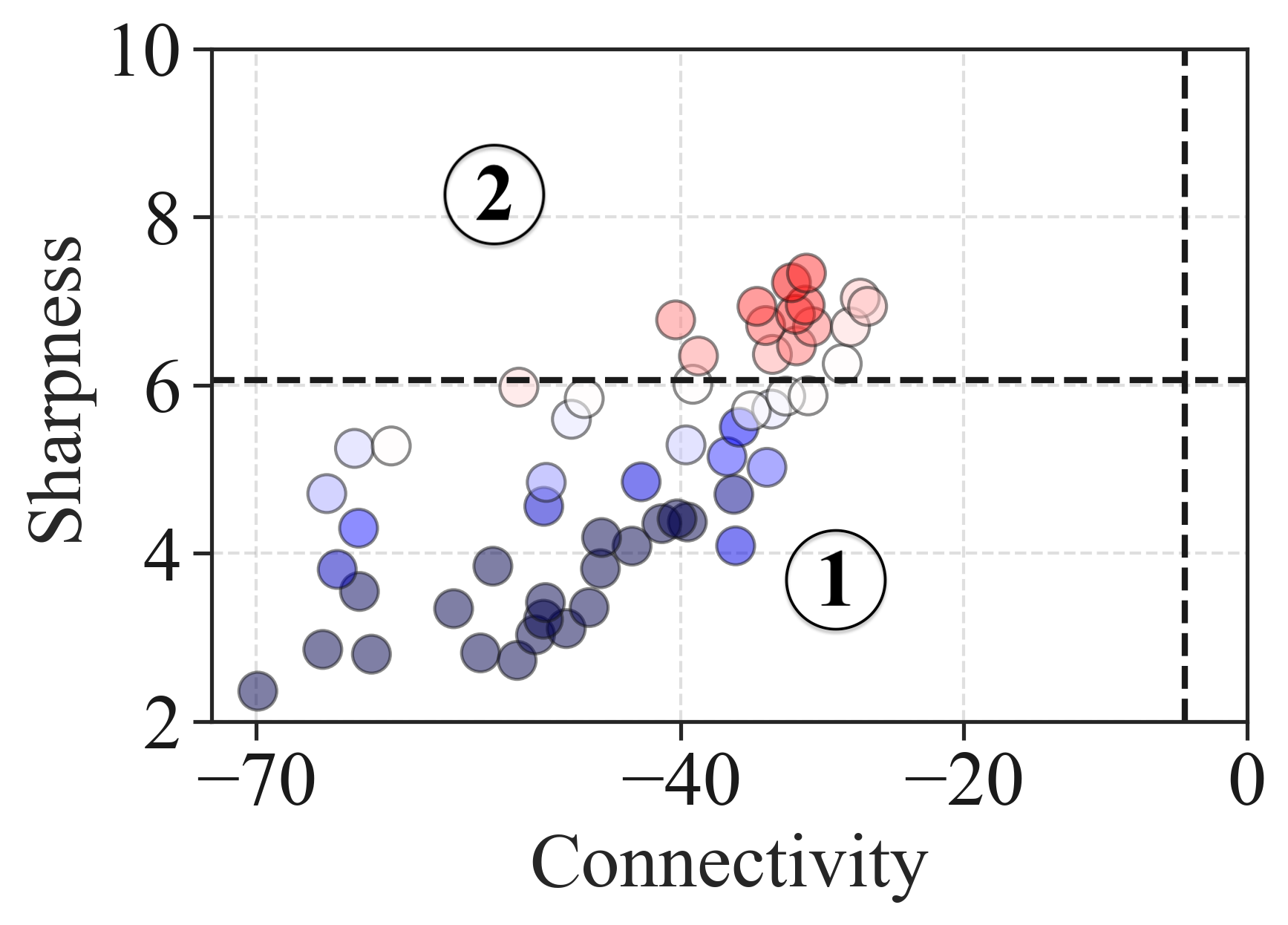}
        \caption{Test (ViT)  } \label{fig:vit-vis-d} 
    \end{subfigure} 
    \end{minipage}  \vspace{-2mm}
    \caption{\textbf{(Visualizing MD tree and its diagnosis results on ViT-tiny models trained on CIFAR-10).}
    \emph{Left side}: structure of the MD tree. The color of the leaf node indicates the predicted class by \ourmethod.
    The threshold values are learned from the training set.
    \emph{Right side}: 
    The first row represents training samples, and the second row represents test samples.
    Each colored circle represents one sample (which is one pre-trained model), and the color represents the ground-truth label: blue means the hyperparameter is too large, while red means small.
    The black dashed line indicates the decision boundary of \ourmethod.
    Each numbered regime on the right corresponds to the leaf node with the same number on the tree.
    }\label{fig:rebut-vit-md-vis}
\end{figure*}

Figure~\ref{fig:rebut-vit-md-vis} reveals that ViT models fall into Regimes~\textcircled{\raisebox{-0.9pt}{1}} and~\textcircled{\raisebox{-0.9pt}{2}} within our MD tree constructed from ResNet18 models, with Regimes~\textcircled{\raisebox{-0.9pt}{3}} and~\textcircled{\raisebox{-0.9pt}{4}} absent for these ViT models. 
This finding shows the differing loss landscape properties between convolutional neural networks and transformer architectures. 
Despite this difference, the MD tree maintains strong diagnostic performance for ViT models. 
This result is attributed to the effective transfer of the sharpness metric's decision boundary, which separates Regimes \textcircled{\raisebox{-0.9pt}{1}} and \textcircled{\raisebox{-0.9pt}{2}}, from ResNet18 to ViT models.

\section{Corroborating Results}\label{sec:corr-result}
We provide additional results to corroborate the findings in the main paper.
Section~\ref{sec:corr-result-baseline} presents the comparison with more baseline methods.
Section~\ref{sec:corr-result-tree-hier} demonstrates that \ourmethod's tree hierarchy helps improve its generalization ability.
Section~\ref{sec:corr-result-q1-q2} visualizes the validation metrics of the pre-trained models in Q1 and Q2 tasks, which is used to explain why validation metrics are not effective in failure source diagnosis.
Section~\ref{sec:corr-transfer-q1-q2} visualizes the loss landscape metrics of the pre-trained models under scale transfer scenario in Q1 and Q2 tasks, which is used to explain why \ourmethod can transfer from small-scale models to large-scale models.\looseness-1

\subsection{More Baseline Comparison} \label{sec:corr-result-baseline}
\begin{figure}[!h]
    \centering
    \includegraphics[width=\linewidth,keepaspectratio]{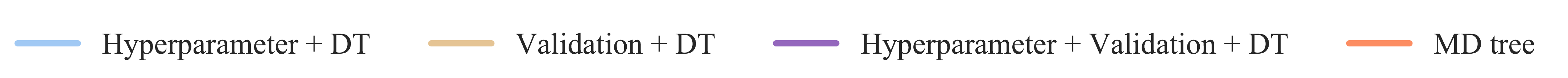} \\
    \begin{subfigure}{0.3\linewidth}
    \includegraphics[width=\linewidth,keepaspectratio]{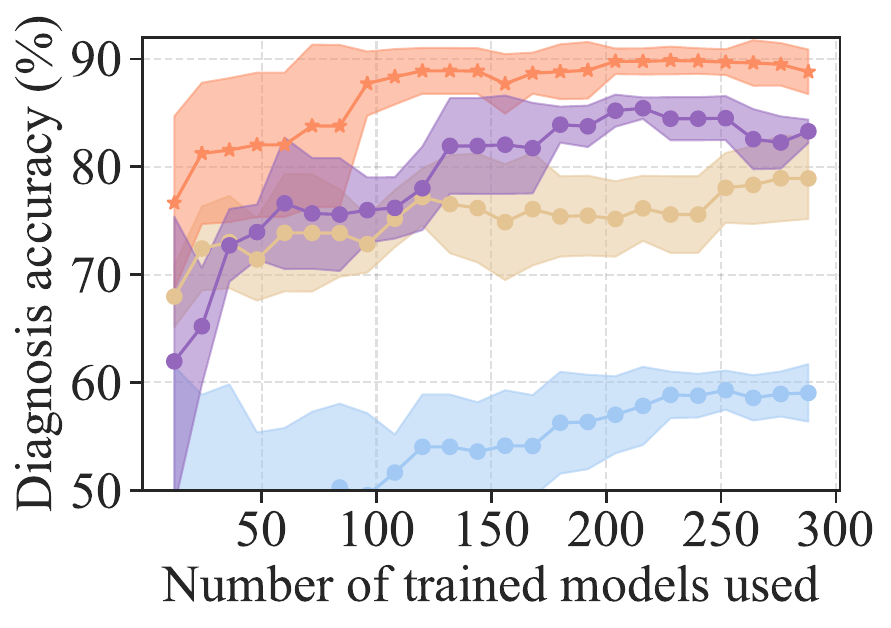}
    \caption{Dataset transfer}~\label{fig:combined-baseline-fail-random}
    \end{subfigure} 
    \begin{subfigure}{0.3\linewidth}
        \includegraphics[width=\linewidth]{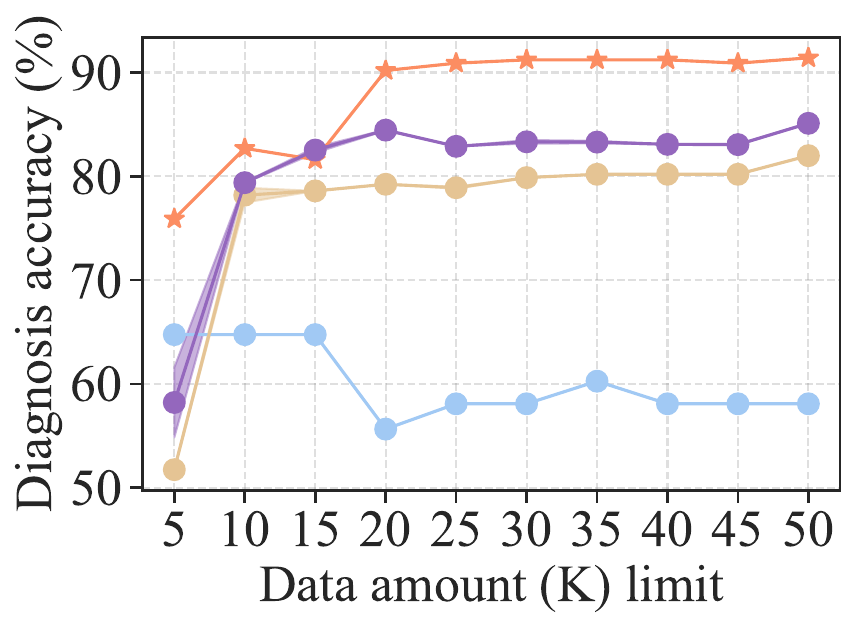}
        \caption{Scale transfer}~\label{fig:combined-baseline-fail-data} 
    \end{subfigure} 
    \hspace{4mm}
    \begin{subfigure}{0.3\linewidth}
        \includegraphics[width=\linewidth]{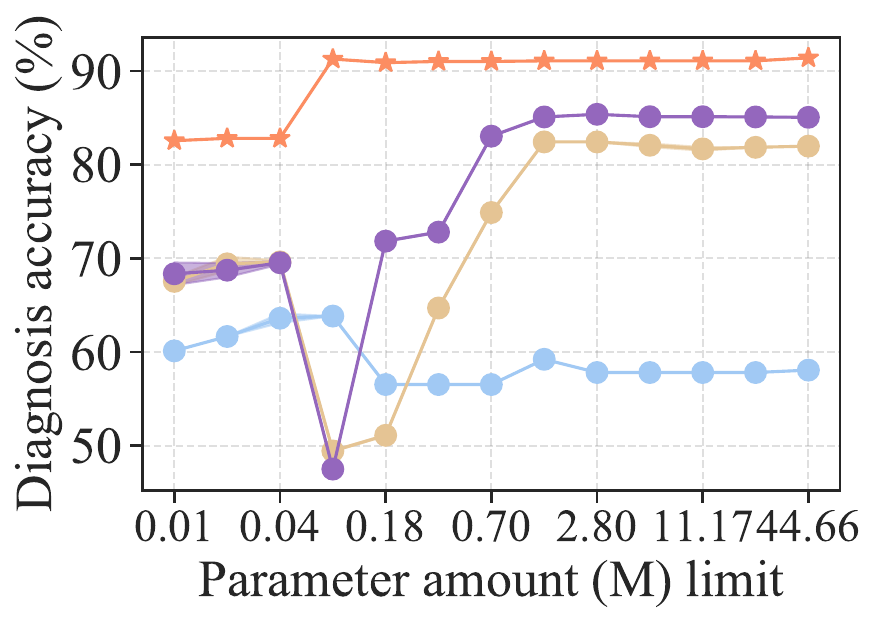}
        \caption{Scale transfer}~\label{fig:combined-baseline-fail-para} 
    \end{subfigure} 
    \vspace{-5mm}
    \caption{\textbf{(Comparing \ourmethod to additional baseline method (Hyperparameter + Validation + DT) on Q1 tasks of predicting optimizer hyperparameter is large or small).}
    The $y$-axis indicates the diagnosis accuracy.
    (a) The $x$-axis indicates the number of pre-trained models used for building the training set. 
    (b) The $x$-axis indicates the maximum amount of training (image) data for training models in the training set.
    (c) The $x$-axis indicates the maximum number of parameters of the models in the training set. The additional baseline outperforms the other two baseline methods, but our method \ourmethod still outperforms all three in dataset transfer and scale transfer scenarios.\looseness-1
  }~\label{fig:combined-baseline}  \vspace{-6mm}
\end{figure}

In Figure~\ref{fig:combined-baseline}, we present additional results by comparing our method to a stronger baseline that utilizes the validation metric and hyperparameter as the features of the decision tree. 
This method is titled ``Hyperparameter + Validation + DT'' and is represented as the purple curve. We can see that the additional baseline outperforms the previous two baseline methods. 
However, \ourmethod (orange curve) still outperforms the three baselines in both the dataset and the scale transfer scenarios. 
The experiments are repeated for five runs and the mean and standard deviation are reported.

\subsection{Investigating Importance of \ourmethod's Tree Hierarchy} \label{sec:corr-result-tree-hier}
\begin{figure}[!ht]
    \centering
    \begin{subfigure}{0.3\linewidth}
    \includegraphics[width=\linewidth,keepaspectratio]{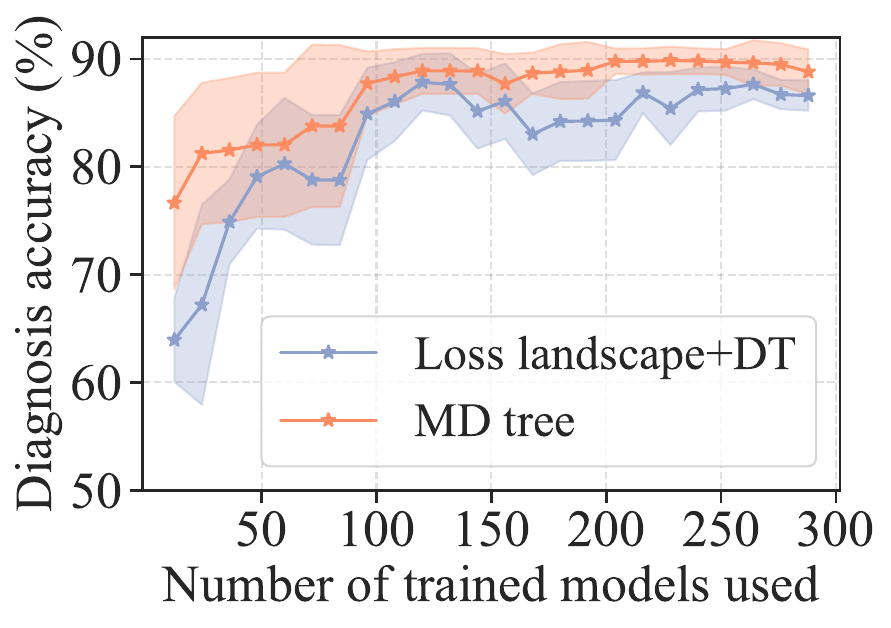} 
    \caption{Q1: optimizer hyper.}
    \end{subfigure}
    \begin{subfigure}{0.3\linewidth}
    \includegraphics[width=\linewidth,keepaspectratio]{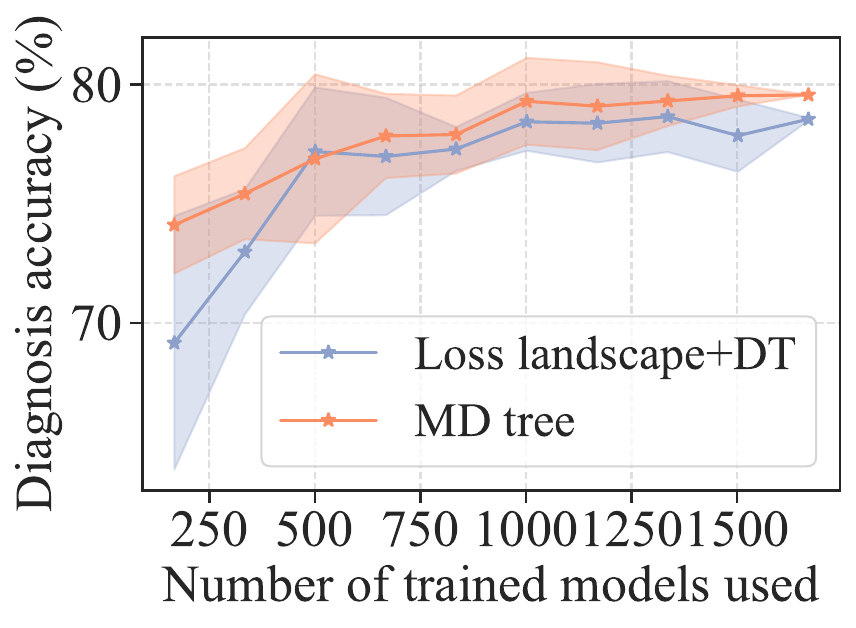} 
    \caption{Q2: optimizer vs. model}
    \end{subfigure} \vspace{-3mm}
    \caption{\textbf{(Comparing \ourmethod to standard DT on Q1 and Q2 task, with both utilizing the loss landscape metrics as features).} The $y$-axis indicates the diagnosis accuracy. The $x$-axis indicates the number of pre-trained models used for building the training set. \ourmethod outperforms normal DT with unfixed hierarchy trained with loss landscape metrics especially when using fewer models as the training set. \looseness-1
  }~\label{fig:fix-abl-q1-q2} \vspace{-5mm}
\end{figure}

\begin{figure}[!ht]
    \centering
    \includegraphics[width=0.45\linewidth,keepaspectratio]{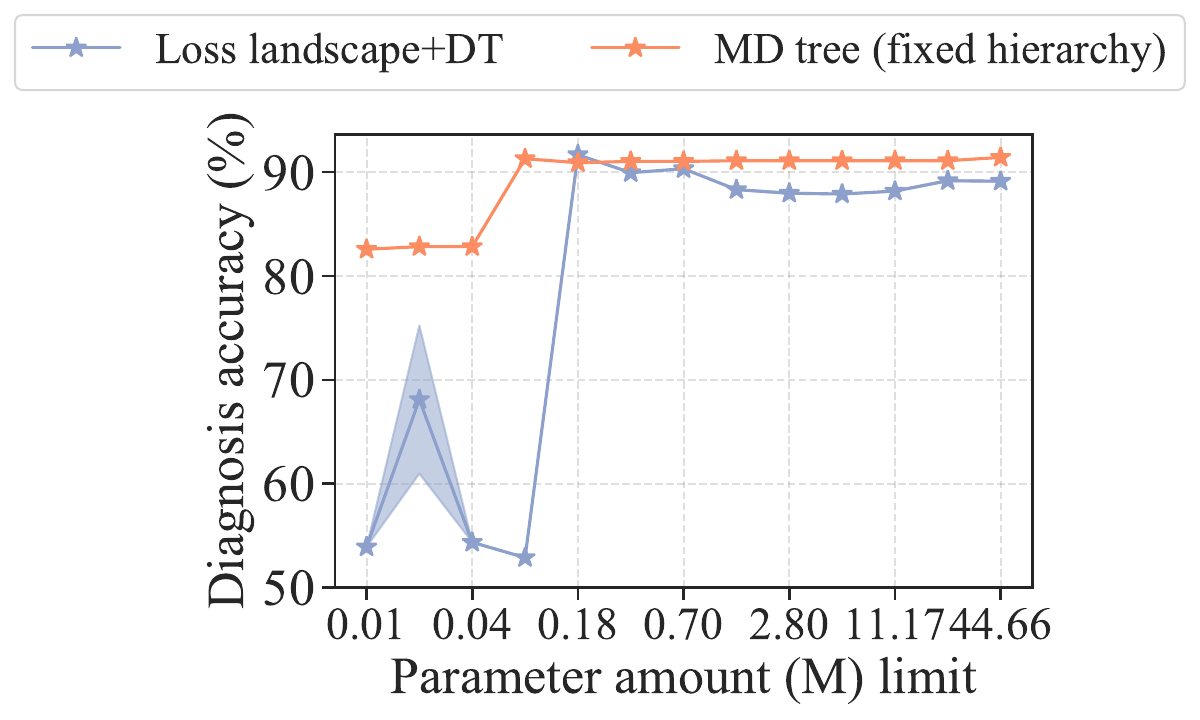} \vspace{-2mm}
    \caption{\textbf{(The fixed hierarchy of MD tree helps \ourmethod to generalize to unseen models with larger scales).} The $y$-axis indicates diagnosis accuracy in Q1 while the $x$-axis represents the maximum number of parameters of the models in the training set. \ourmethod outperforms normal DT with unfixed hierarchy trained with loss landscape metrics especially when using small-scale models as the training set.\looseness-1}~\label{fig:fix-abl-scale}
\end{figure}

We summarize the two main reasons why \ourmethod can generalize to diagnosing unseen models: 1) the multi-phase pattern in the NN hyperparameter space, and 2) the fixed tree hierarchy of our method. The first reason has been discussed in Section~\ref{sec:q1-few-shot}, here we elaborate on the second reason.
To achieve better generalization, we restricted the capacity of the decision tree by using a decision tree with only a few loss landscape features and fixing the tree hierarchy, prioritizing those features shown by \citet{yang2021taxonomizing} to have sharp phase transitions. The goal is to evaluate whether the loss landscape metrics are useful and avoid overfitting.
In Figure~\ref{fig:fix-abl-q1-q2}, we compare two methods on Q1 and Q2: 1) \ourmethod is our method, which uses a fixed hierarchy. 2) ``Loss landscape + DT'' is the baseline method, which uses the same set of loss landscape metrics as \ourmethod, but uses a decision tree with unfixed and learnable architecture. The results show that \ourmethod outperforms the baseline method in both tasks, particularly when using fewer models as the training set. The same advantage of \ourmethod holds for scale transfer in Figure~\ref{fig:fix-abl-scale}. These findings demonstrate the importance of a fixed tree hierarchy in enhancing the generalizability of the diagnostic method.\looseness-1

\subsection{Visualization for Validation Metrics in Q1 and Q2}\label{sec:corr-result-q1-q2}

\begin{figure}[!h]
    \centering
    \begin{subfigure}{0.25\linewidth}
        \includegraphics[width=\linewidth]{figs/md_tree/temp/zero_shot_legend.pdf}
    \end{subfigure} \\
    \begin{subfigure}{0.25\linewidth}
        \includegraphics[width=\linewidth]{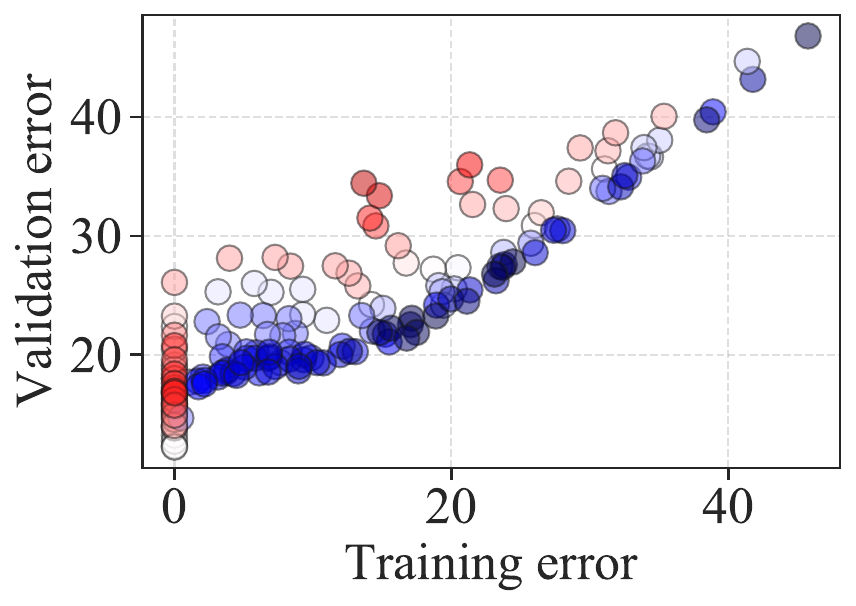}
        \caption{Training, w/o label noise}~\label{fig:zero-shot-temp-test-a}
    \end{subfigure} 
    \begin{subfigure}{0.25\linewidth}
        \includegraphics[width=\linewidth]{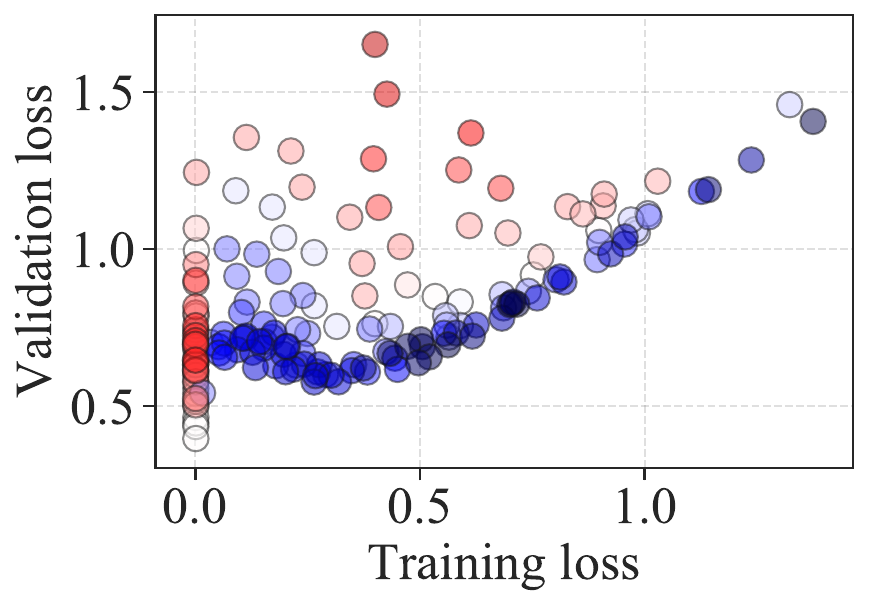}
        \caption{Training, w/o label noise}~\label{fig:zero-shot-temp-test-b}
    \end{subfigure} \\
    \vspace{-2mm}
    \begin{subfigure}{0.25\linewidth}
        \includegraphics[width=\linewidth]{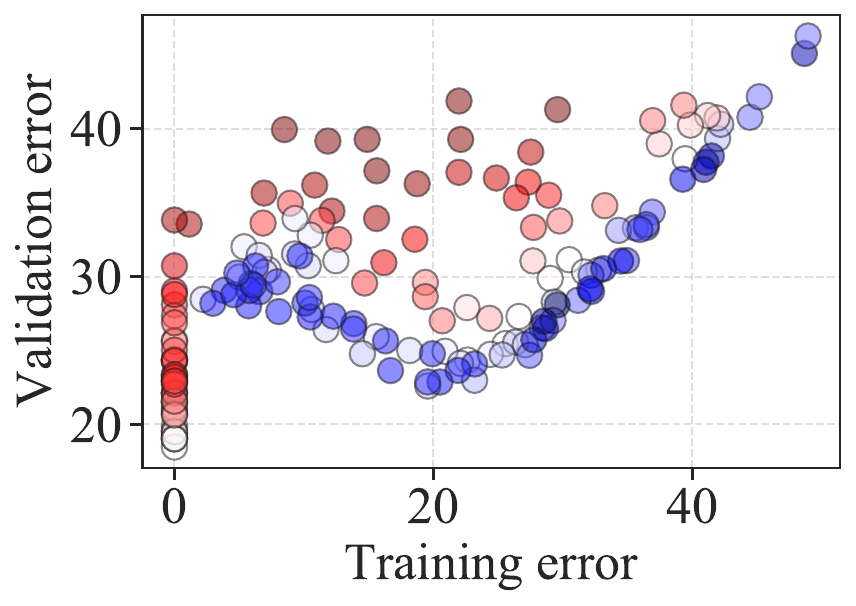}
        \caption{Test, w/ label noise}~\label{fig:zero-shot-temp-test-c}
    \end{subfigure} 
    \begin{subfigure}{0.25\linewidth}
        \includegraphics[width=\linewidth]{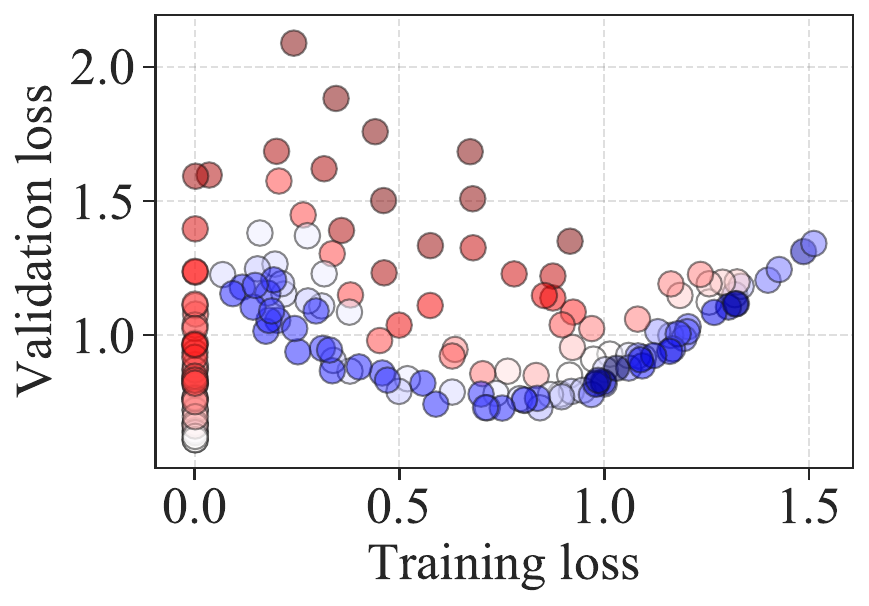}
        \caption{Test, w/ label noise}~\label{fig:zero-shot-temp-test-d}
    \end{subfigure} 
    \vspace{-3mm}
    \caption{\textbf{(Applying validation metrics such as training/validation error and loss to diagnose the failure source of models).} (a)(b) Training set comprises models trained without label noise, (c)(d) test set comprises models trained with label noise.}~\label{fig:zero-shot-temp-test} \vspace{-4mm}
\end{figure}

Figure~\ref{fig:zero-shot-temp-test} visualizes the validation metrics and the ground-truth labels Q1 diagnosis task (optimizer hyperparameter is large or small) of pre-trained models from the training or test set. 
We can tell that the validation metrics have a nonlinear and complicated correlation with the failure source labels. 
Also, the red-versus-blue boundary in the training set (Figure~\ref{fig:zero-shot-temp-test-a} and \ref{fig:zero-shot-temp-test-b}) are unable to make a good transfer to the test set  (Figure~\ref{fig:zero-shot-temp-test-c} and \ref{fig:zero-shot-temp-test-d}). This indicates the low transferability of validation metrics between different sets of models. 

\begin{figure}[!h]
    \centering
    \begin{subfigure}{0.35\linewidth}
        \includegraphics[width=\linewidth]{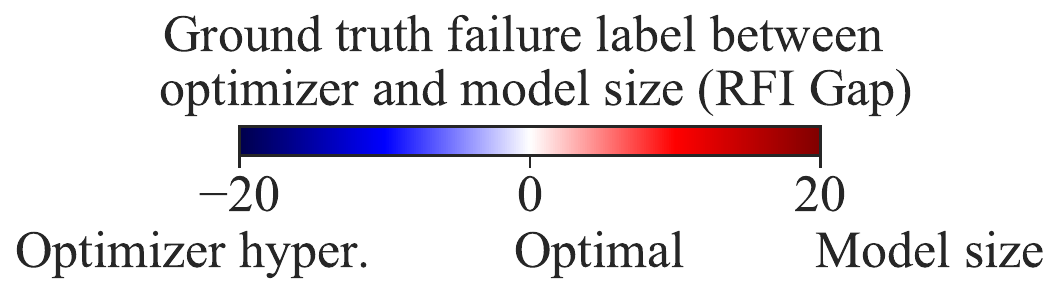}
    \end{subfigure} \\
    \begin{subfigure}{0.25\linewidth}
        \includegraphics[width=\linewidth]{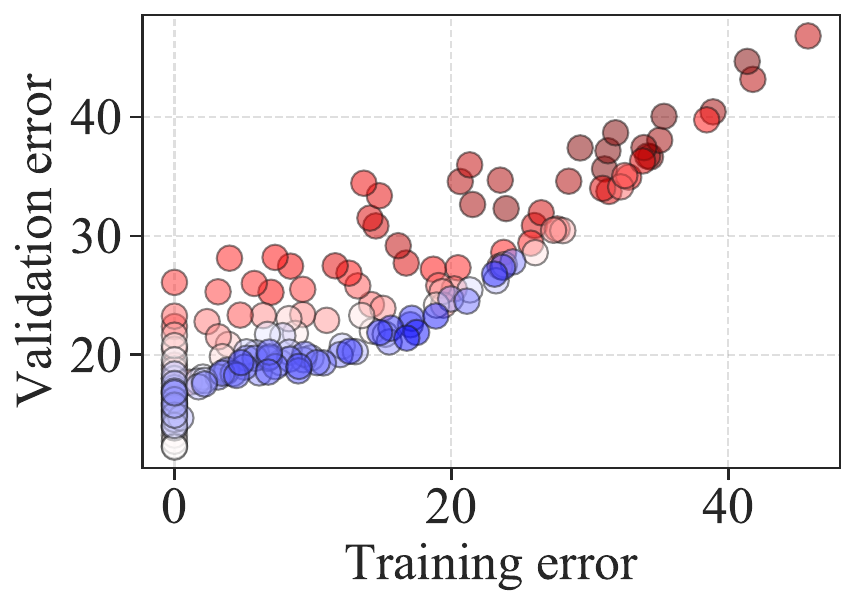}
        \caption{Training, w/o label noise}~\label{fig:zero-shot-width-temp-test-a}
    \end{subfigure} 
    \begin{subfigure}{0.25\linewidth}
        \includegraphics[width=\linewidth]{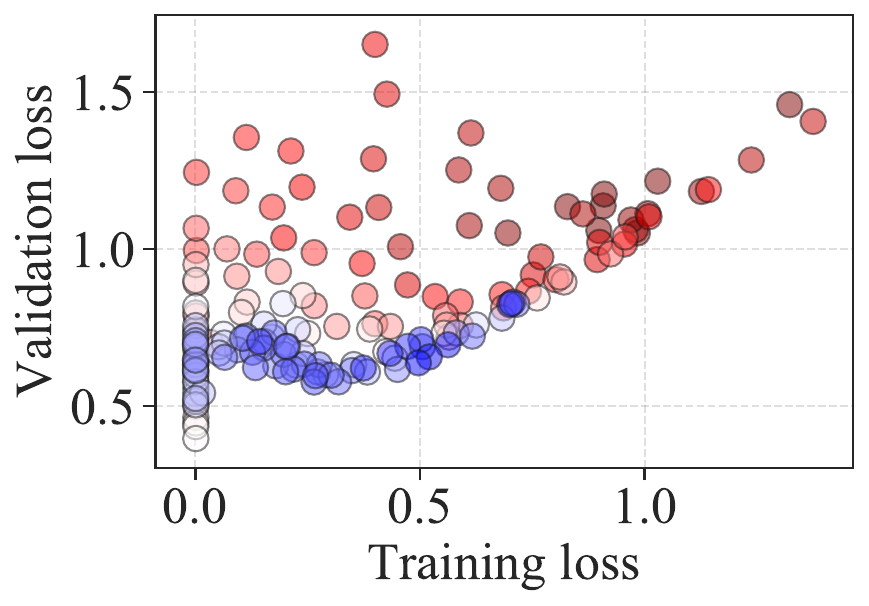}
        \caption{Training, w/o label noise}~\label{fig:zero-shot-width-temp-test-b}
    \end{subfigure}  \\
    \begin{subfigure}{0.25\linewidth}
        \includegraphics[width=\linewidth]{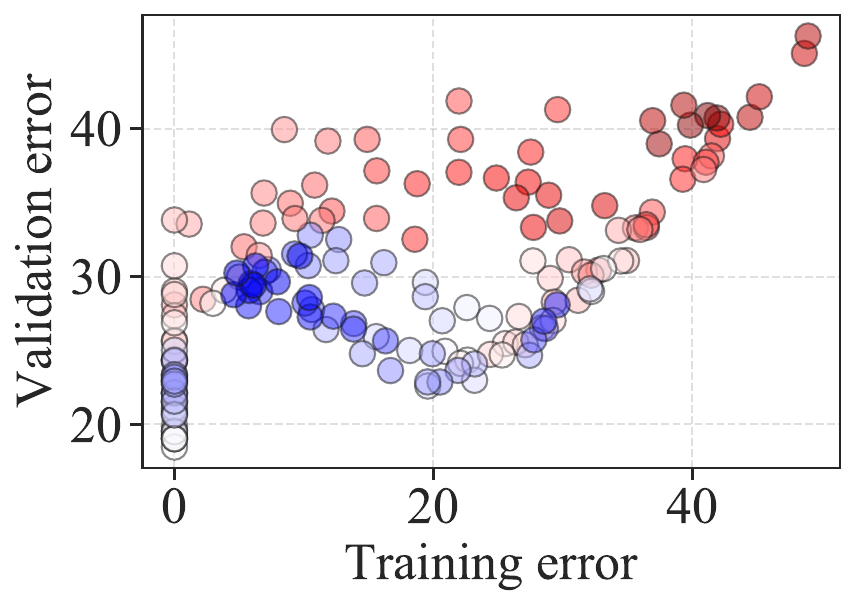}
        \caption{Test, w/ label noise}~\label{fig:zero-shot-width-temp-test-c}
    \end{subfigure} 
    \begin{subfigure}{0.25\linewidth}
        \includegraphics[width=\linewidth]{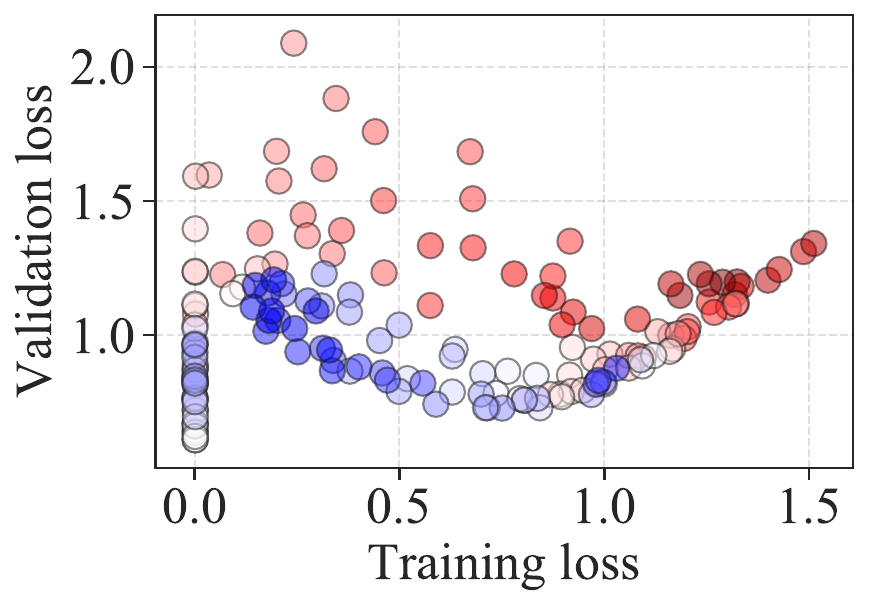}
        \caption{Test, w/ label noise}~\label{fig:zero-shot-width-temp-test-d}
    \end{subfigure} 
    \vspace{-3mm}
    \caption{\textbf{(Applying validation metrics such as training/validation error and loss to diagnose the failure source of models.)} (a)(b) training set comprises models trained without label noise, (c)(d) test set comprises models trained with label noise.}~\label{fig:zero-shot-width-temp-test} \vspace{-7mm}
\end{figure}

Figure~\ref{fig:zero-shot-width-temp-test} presents the visualization of validation metrics of pre-trained models in the Q2 diagnosis task (optimizer versus model size). Compared to those in Q1, we observe a more complicated correlation between failure source labels and validation metrics and worse transferability of decision boundary between training and test.

\vspace{-0.5mm}
\begin{figure}[!h]
    \centering
    \begin{subfigure}{0.28\linewidth}
        \includegraphics[width=\linewidth]{figs/md_tree/temp/zero_shot_legend.pdf}
    \end{subfigure} 
    \begin{subfigure}{0.3\linewidth}
        \includegraphics[width=\linewidth]{figs/md_tree/decision_bd.pdf}
    \end{subfigure} 
    \\
    \centering
    \begin{subfigure}{\linewidth}
        \includegraphics[width=\linewidth]{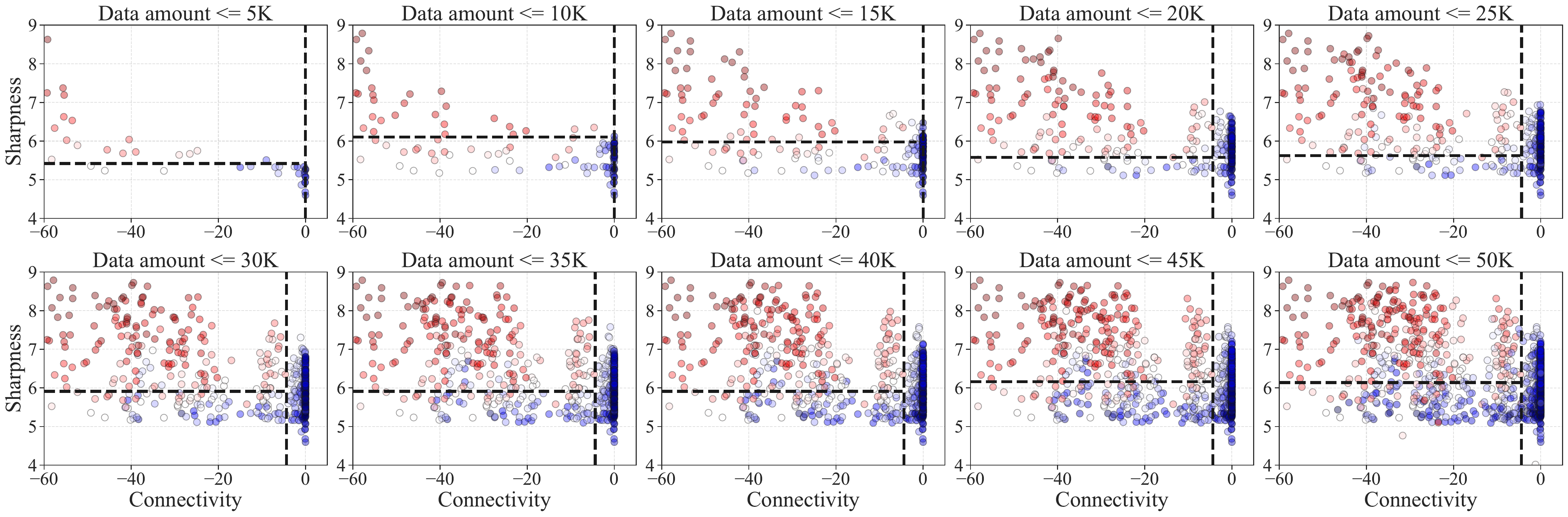}
    \end{subfigure} \vspace{-4mm}
    \caption{\textbf{(Visualization of models in training set in scale (data amount) transfer and decision boundary of \ourmethod in Q1).} Each subfigure presents the models under a specific data amount limit. The $x$-axis presents the connectivity and the $y$-axis presents the sharpness.\looseness-1}~\label{fig:vis-scale-data-transfer}\vspace{-5mm}
\end{figure}

\begin{figure}[!h]
    \centering
    \begin{subfigure}{0.4\linewidth}
        \includegraphics[width=\linewidth]{figs/md_tree/width_temp/zero_shot_legend.pdf}
    \end{subfigure}  
    \begin{subfigure}{0.3\linewidth}
        \includegraphics[width=\linewidth]{figs/md_tree/decision_bd.pdf}
    \end{subfigure} 
    \\
    \begin{subfigure}{\linewidth}
        \includegraphics[width=\linewidth]{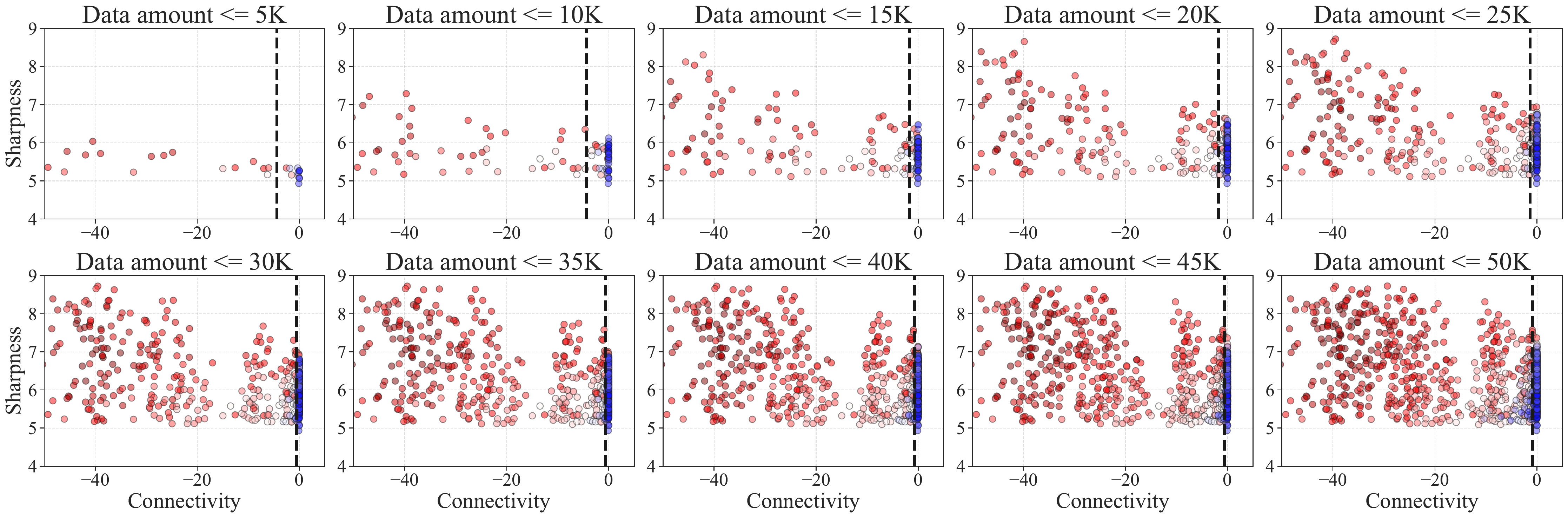}
    \end{subfigure}  \vspace{-3mm}
    \caption{\textbf{(Visualization of models in training set in scale (data amount) transfer and decision boundary of \ourmethod in Q2 task).} Each subfigure presents the models under a specific data amount limit. The models are partitioned into non-interpolating (larger training error) regimes by \ourmethod.
    The $x$-axis presents the connectivity and the $y$-axis presents the sharpness. }~\label{fig:vis-scale-data-transfer-q2}
\end{figure}

\subsection{Visualization for Scale Transfer in Q1 and Q2}\label{sec:corr-transfer-q1-q2}
\begin{figure}[!h]
    \centering
    \begin{subfigure}{0.25\linewidth}
        \includegraphics[width=\linewidth]{figs/md_tree/temp/zero_shot_legend.pdf}
    \end{subfigure} 
    \begin{subfigure}{0.3\linewidth}
        \includegraphics[width=\linewidth]{figs/md_tree/decision_bd.pdf}
    \end{subfigure}  \\
    \begin{subfigure}{\linewidth}
        \includegraphics[width=\linewidth]{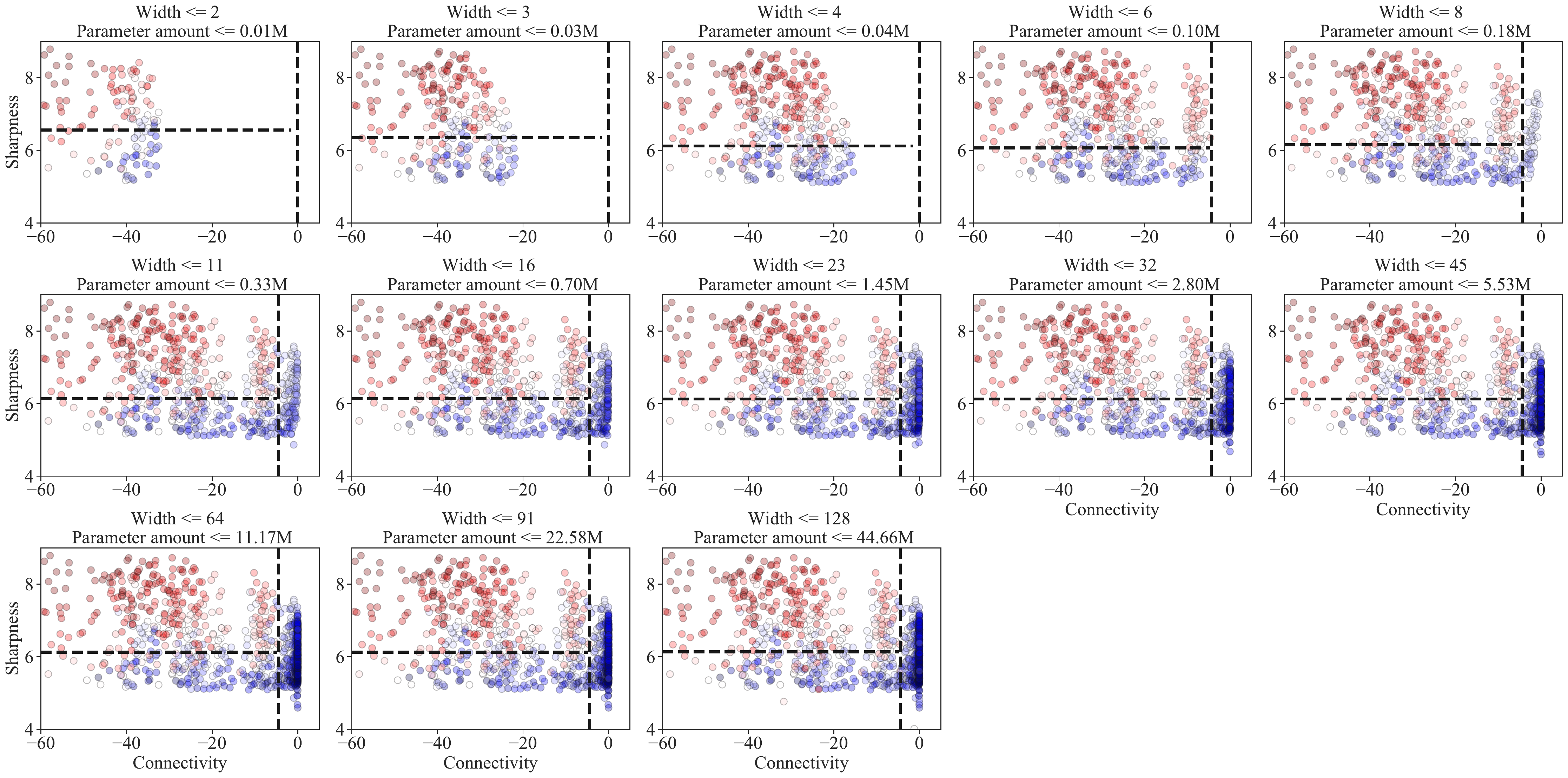}
    \end{subfigure}  \vspace{-3mm}
    \caption{\textbf{(Visualization of models in training set in scale (number of model parameters) transfer and decision boundary of \ourmethod in Q1 task).} Each subfigure presents the models under a specific parameter amount limit. The $x$-axis presents the connectivity and the $y$-axis presents the sharpness.}\vspace{-6mm}~\label{fig:vis-scale-parameter-transfer}
\end{figure}

Figure~\ref{fig:vis-scale-data-transfer} visualizes the loss landscape metrics of pre-trained models in Q1 and data amount transfer scenarios.
We can find that the decision boundaries established by \ourmethod for small data sets (20K data points) closely resemble those for larger scales (up to 50K data points), demonstrating the transferability of \ourmethod.
Figure~\ref{fig:vis-scale-data-transfer-q2} visualizes the loss landscape metrics of pre-trained models in Q2 and data amount transfer scenarios. 
For this task, we observe that the transferability of \ourmethod can be better even in the smallest data amount (less than 5K).
Figure~\ref{fig:vis-scale-parameter-transfer} visualizes the loss landscape metrics of pre-trained models in Q1 and parameter amount transfer scenarios.
Again, it shows that boundaries determined for small-size models (less than 0.1M parameters) resemble those for larger scales (up to 44.66M parameters). 
From all figures, we observe notable transitions of decision boundaries varying from small-scale to large-scale. 
These transitions occur because very small-scale models often exhibit limited ranges in loss landscape metrics, such as consistently high sharpness or consistently negative connectivity. This demonstrates the limitations of transferring from very small-scale to large-scale models.\looseness-1

\FloatBarrier

\section{A Variant of Q2: Which Hyperparameter (Data amount or Optimizer) Leads to failure}\label{sec:appendix-data-temp}
\begin{figure}[!h]
    \centering
    \includegraphics[width=0.7\linewidth,keepaspectratio]{figs/md_tree/temp_legend.pdf} \vspace{-2mm} \\
    \begin{subfigure}{0.32\linewidth}
        \includegraphics[width=\linewidth]{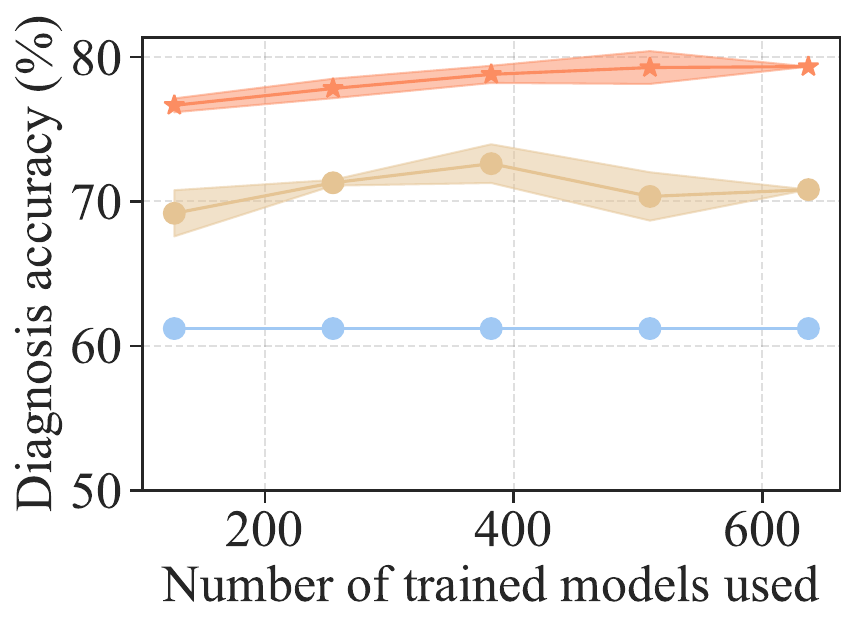} 
        \caption{Dataset transfer}~\label{fig:data-vs-temp-a}
    \end{subfigure} 
    \begin{subfigure}{0.32\linewidth}
        \includegraphics[width=\linewidth]{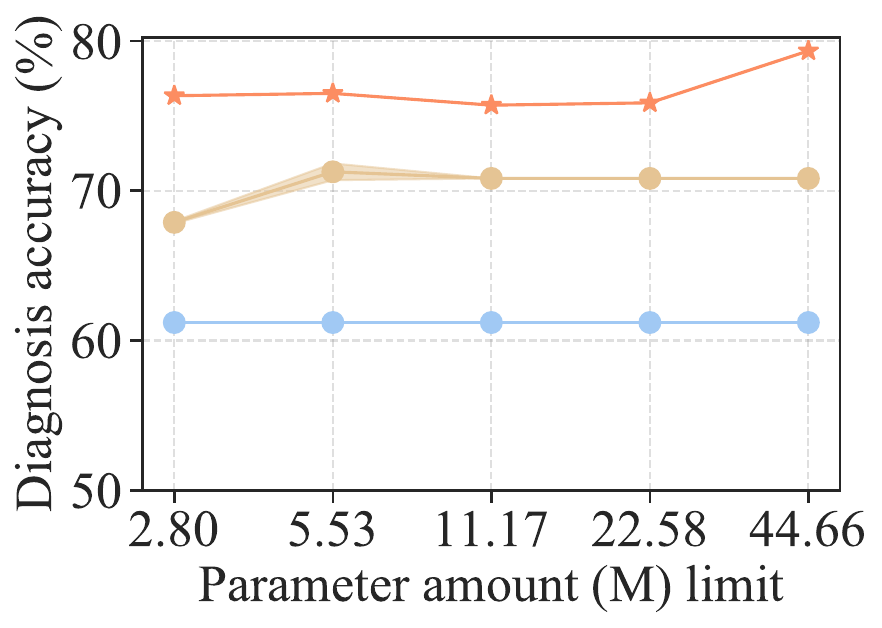}  
        \caption{Scale transfer}~\label{fig:data-vs-temp-b}
    \end{subfigure} \vspace{-6mm}
    \caption{\textbf{(Comparing \ourmethod to baseline methods in Q2 task in dataset and scale transfer).} 
    $y$-axis indicates the diagnosis accuracy.
    (a) $x$-axis indicates the number of pre-trained models used for building the training set.
    (b) $x$-axis indicates the maximum number of parameters of trained models in the training set for fitting the classifier.}~\label{fig:data-vs-temp} 
\end{figure}

\begin{figure}[!thb]
    \centering
    \includegraphics[width=0.5\linewidth]{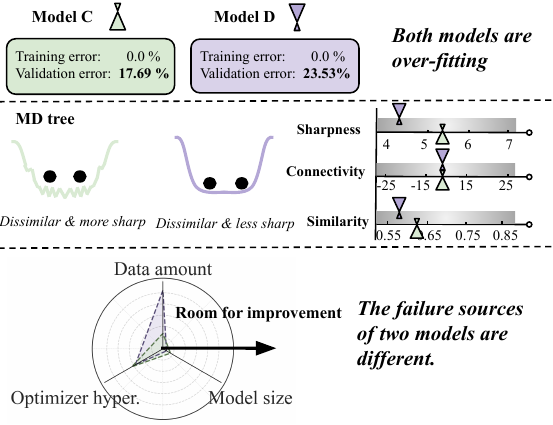}
    \caption{
    \textbf{(Case Study: \ourmethod vs. validation-based method for Q2).}
    \emph{Top}: Validation metrics provide limited diagnosis: both models are overfitting.
    \emph{Middle}: Loss landscape metrics further distinguish the two models: Model C has sharper local minima, while Model D has lower similarity. \emph{Bottom}: Model C has an optimizer issue, while Model D's problem is due to insufficient data.
    }
\label{fig:one-out-of-two-data-temp} 
\end{figure}

\begin{figure*}[!ht]
    \begin{minipage}[b]{0.48\linewidth}
        \includegraphics[width=\linewidth]{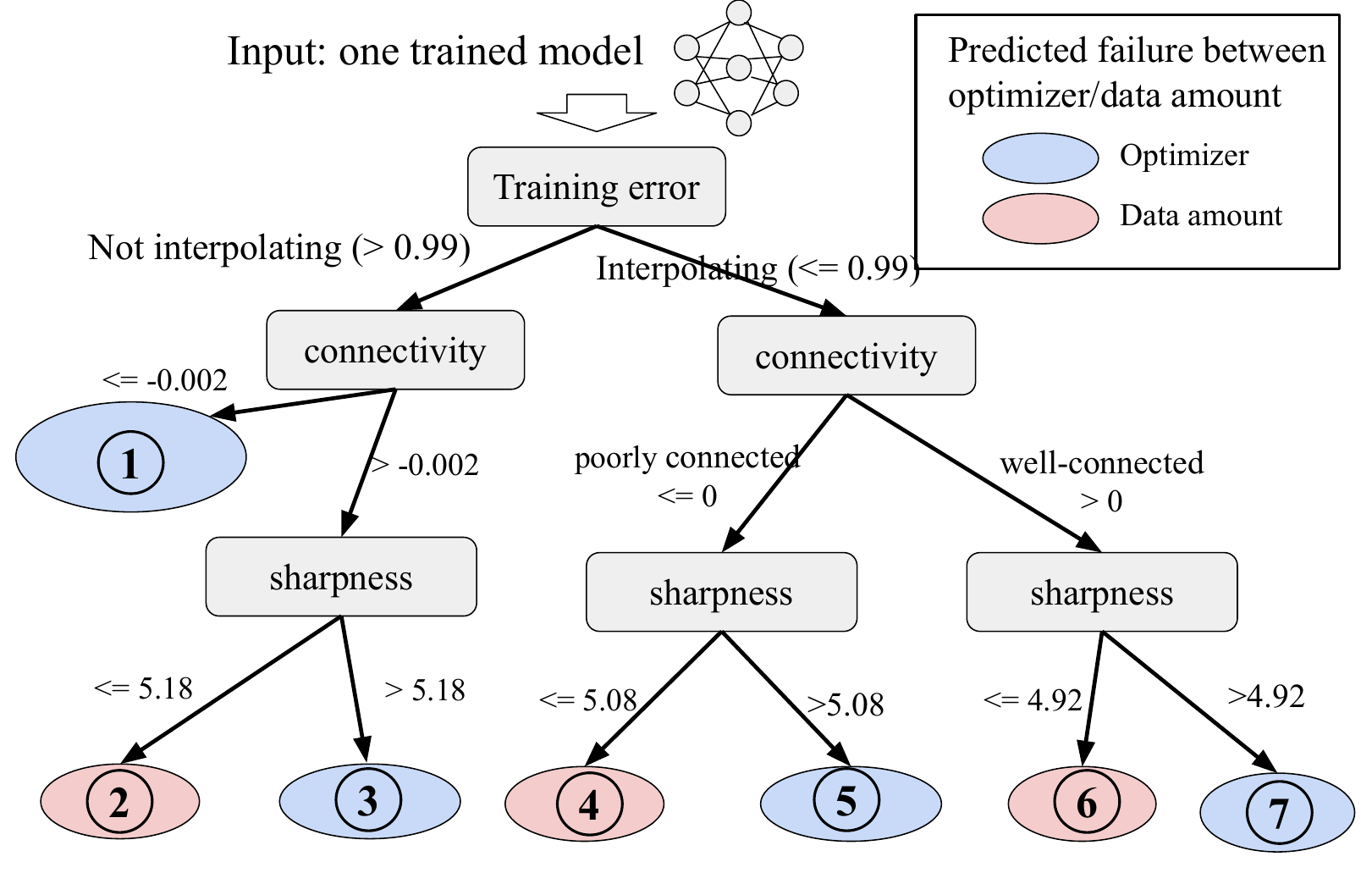}
        \vspace{5mm}
        \subcaption{MD tree for Q2}\label{fig:zero-shot-data-tree}
    \end{minipage}
    \centering
    \begin{minipage}[b]{0.48\linewidth}
    \centering
    \begin{subfigure}{0.55\linewidth}
        \includegraphics[width=\linewidth]{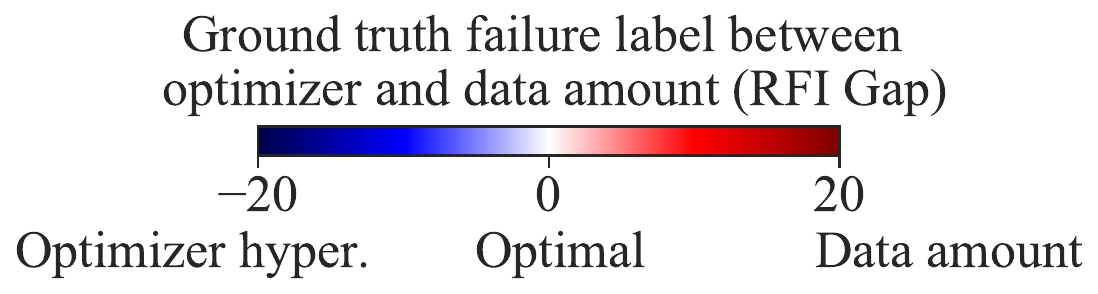}
    \end{subfigure} 
    \begin{subfigure}{0.43\linewidth}
        \includegraphics[width=\linewidth]{figs/md_tree/decision_bd.pdf}
    \end{subfigure} 
    \\
    \begin{subfigure}{0.48\linewidth}
        \includegraphics[width=\linewidth]{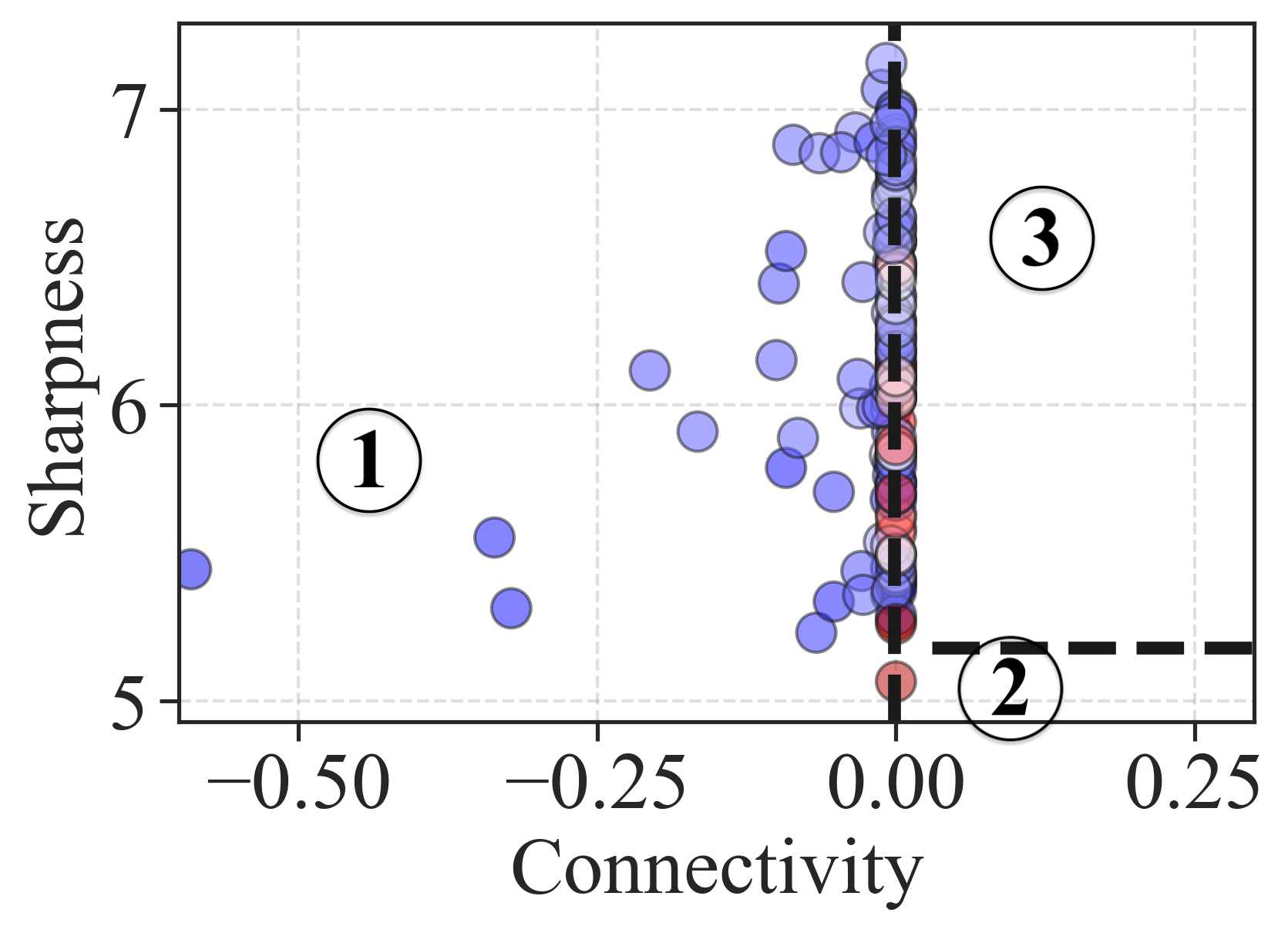}
        \caption{Training (w/o label noise)}~\label{fig:zero-shot-data-temp-b}
    \end{subfigure} 
    \begin{subfigure}{0.48\linewidth}
        \includegraphics[width=\linewidth]{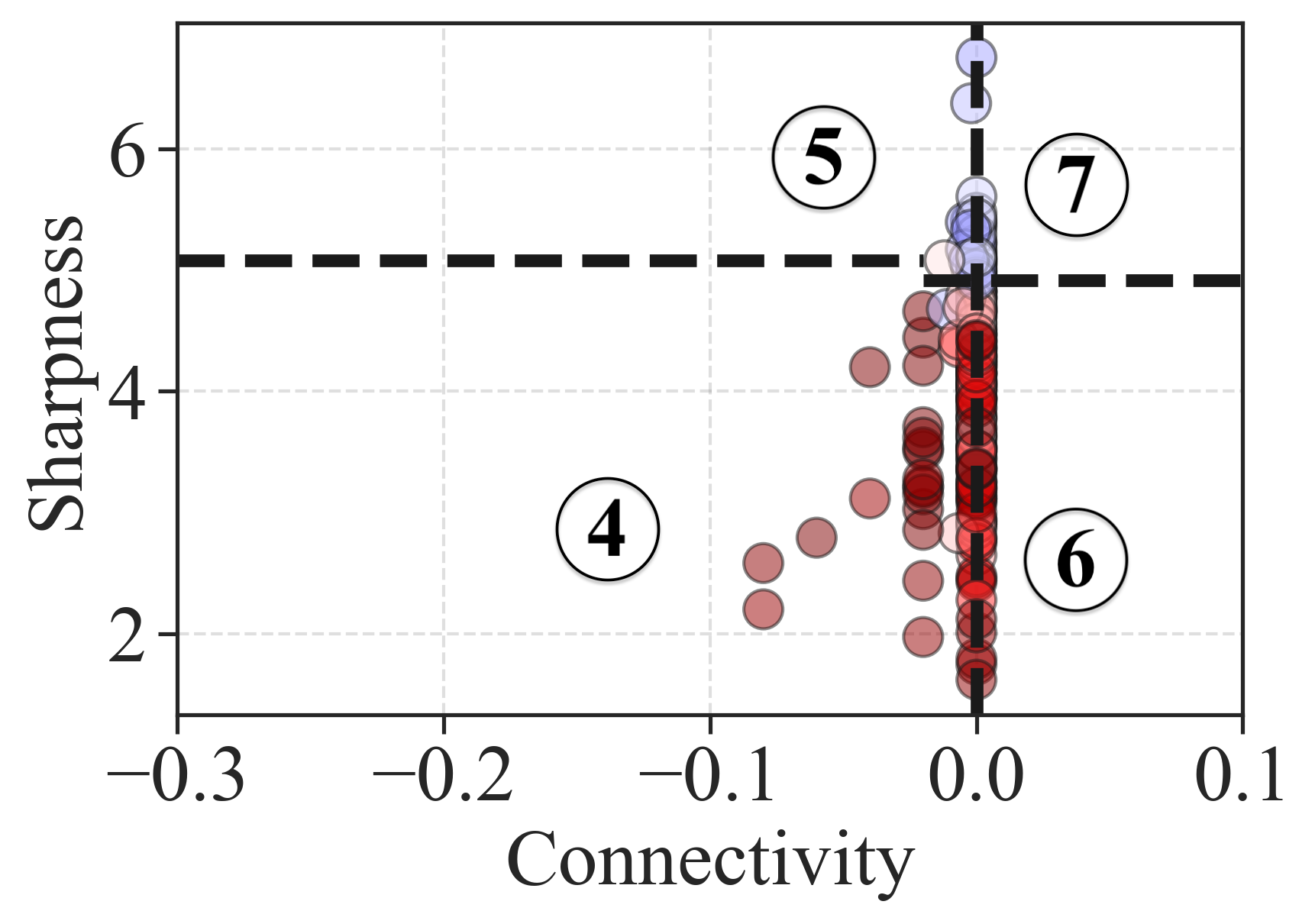}
        \caption{Training (w/o label noise)}~\label{fig:zero-shot-data-temp-c}
    \end{subfigure} \\
    \begin{subfigure}{0.48\linewidth}
        \includegraphics[width=\linewidth]{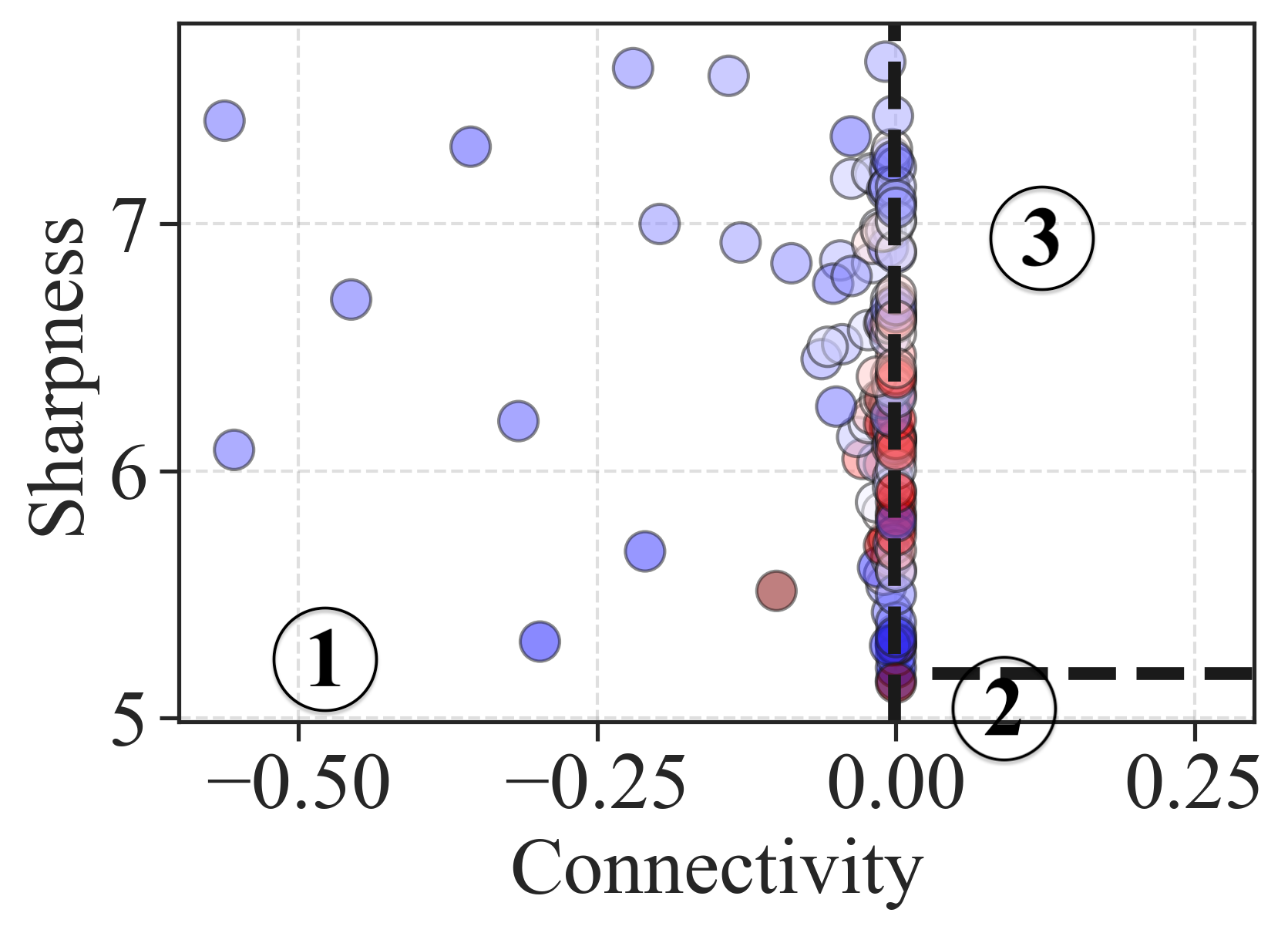}
        \caption{Test (w/ label noise)}~\label{fig:zero-shot-data-temp-d}
    \end{subfigure} 
    \begin{subfigure}{0.48\linewidth}
        \includegraphics[width=\linewidth]{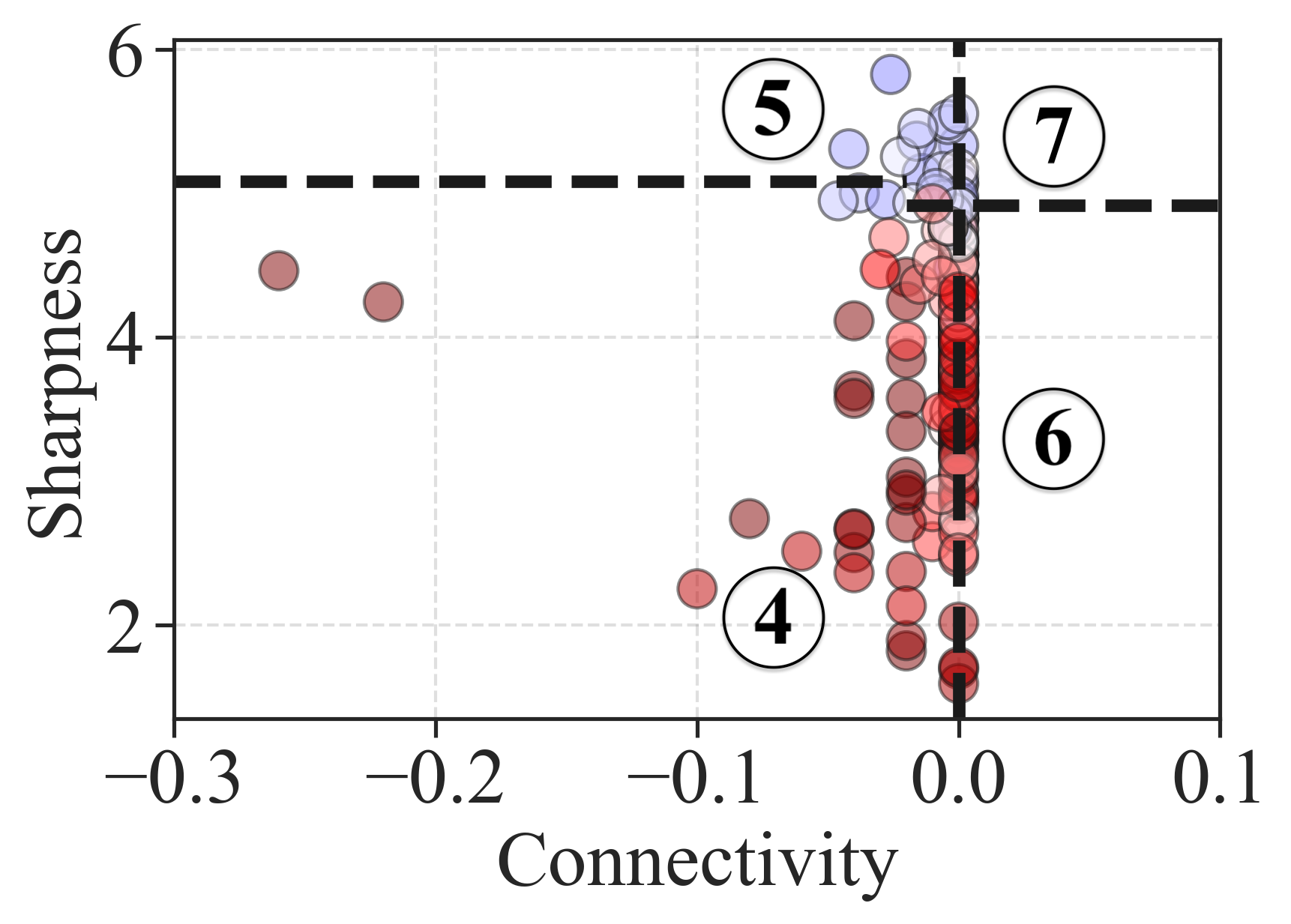}
        \caption{Test (w/ label noise)}~\label{fig:zero-shot-data-temp-e}
    \end{subfigure} 
    \end{minipage} \vspace{-3mm}
    \caption{\textbf{(Visualizing MD tree and its diagnosis results for a variant of Q2 (determining models' failure source on optimizer hyperparameter or data amount)).}
    \emph{Left}: structure of the tree defined in Section~\ref{sec:diag-method}. The color of the leaf node indicates the predicted class.
    The threshold values are learned from the training set.
    \emph{Right}: each circle represents one model configuration, the color represents the ground truth label RFI gap $G$, the blue color means the failure source is the optimizer, while red means the data amount.
    The black dashed line indicates the decision boundary of \ourmethod.
    The samples in~\ref{fig:zero-shot-data-temp-b} and those in~\ref{fig:zero-shot-data-temp-c} are separated by training error. The same applies to~\ref{fig:zero-shot-data-temp-d} and~\ref{fig:zero-shot-data-temp-e}.
    } \vspace{-3mm} ~\label{fig:zero-shot-data-temp}
\end{figure*}

We investigate a variation of Q2, aiming to identify the failure source between data amount and optimizer settings. 
The specifics of the dataset and the transfer setup are outlined in Section~\ref{app:exp-setup}. 
Diagnosis accuracy, presented in Figure~\ref{fig:data-vs-temp}, indicates that \ourmethod exceeds validation-based methods by a 10\% accuracy margin.
A case study in Figure~\ref{fig:one-out-of-two-data-temp} contrasts \ourmethod with validation-based approaches, demonstrating \ourmethod's ability to differentiate between two distinct failure sources that both result in overfitting.
Moreover, \ourmethod offers clear visual insights in Figure~\ref{fig:zero-shot-data-temp}, showing that models with lower sharpness typically suffer from insufficient data, whereas those with higher sharpness face optimizer issues.

\FloatBarrier

\vspace{-5mm}

\begin{figure*}[!th]
    \begin{minipage}[b]{0.48\linewidth}
        \includegraphics[width=\linewidth]{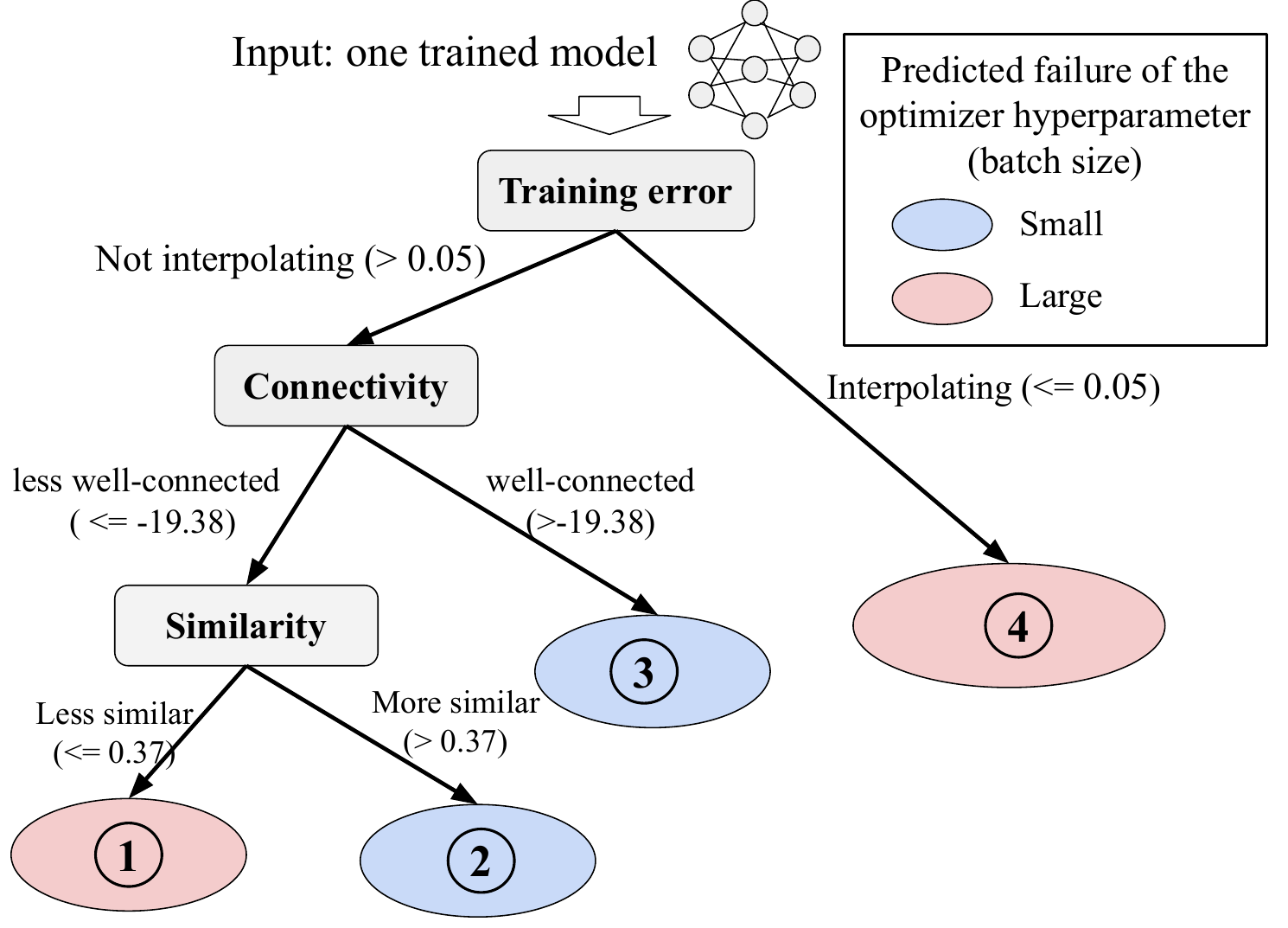}
        \vspace{5mm}
        \subcaption{Ablation study of MD tree for Q1}\label{fig:zero-shot-data-tree-cka}
    \end{minipage}
    \centering
    \begin{minipage}[b]{0.48\linewidth}
    \centering
    \begin{subfigure}{0.48\linewidth}
        \includegraphics[width=\linewidth]{figs/md_tree/temp/zero_shot_legend.pdf}
    \end{subfigure} 
    \begin{subfigure}{0.48\linewidth}
        \includegraphics[width=\linewidth]{figs/md_tree/decision_bd.pdf}
    \end{subfigure} 
    \\
    \begin{subfigure}{0.48\linewidth}
        \includegraphics[width=\linewidth]{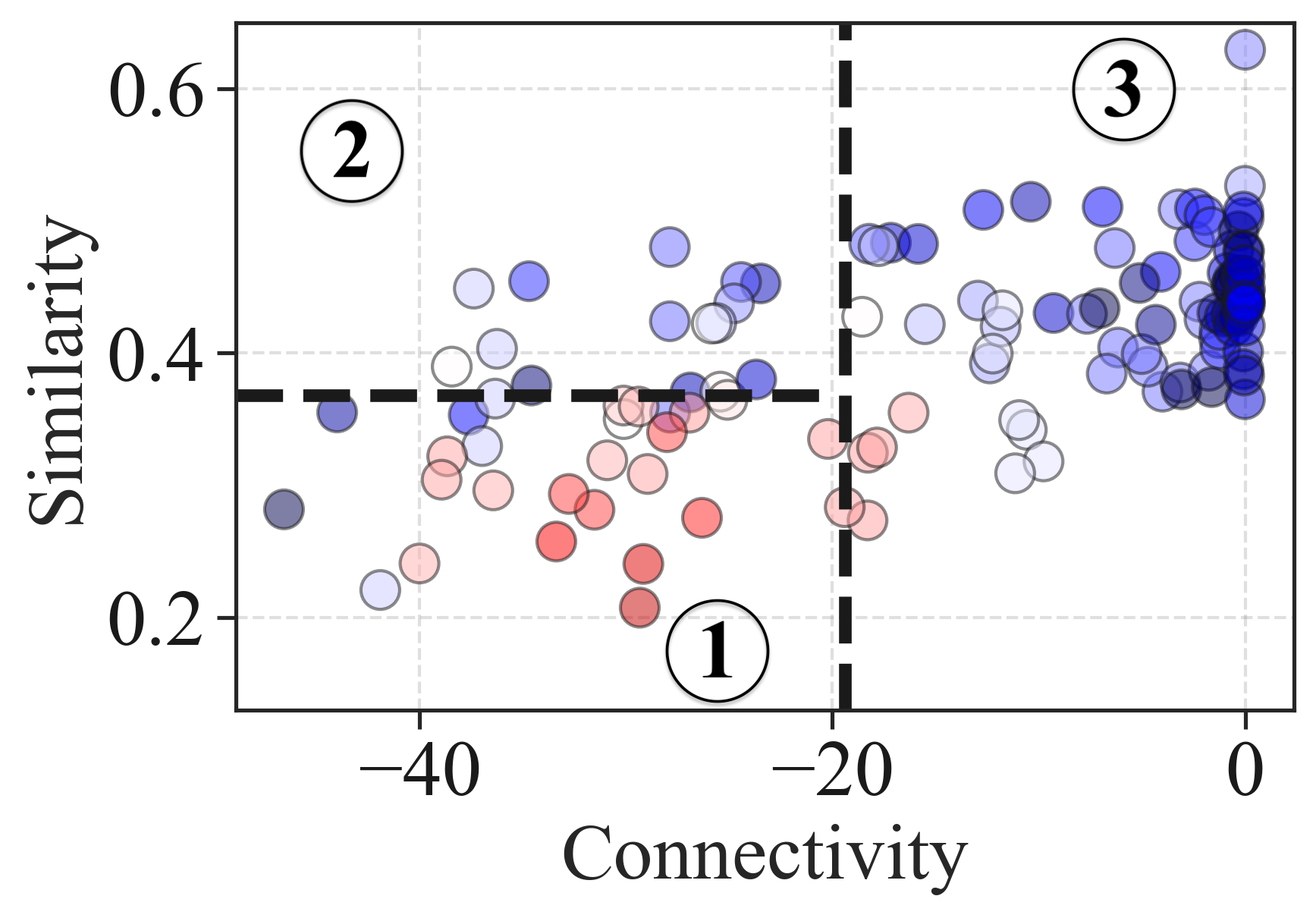}
        \caption{Training (w/o label noise)}~\label{fig:zero-shot-data-temp-cka-b}
    \end{subfigure} 
    \begin{subfigure}{0.48\linewidth}
        \includegraphics[width=\linewidth]{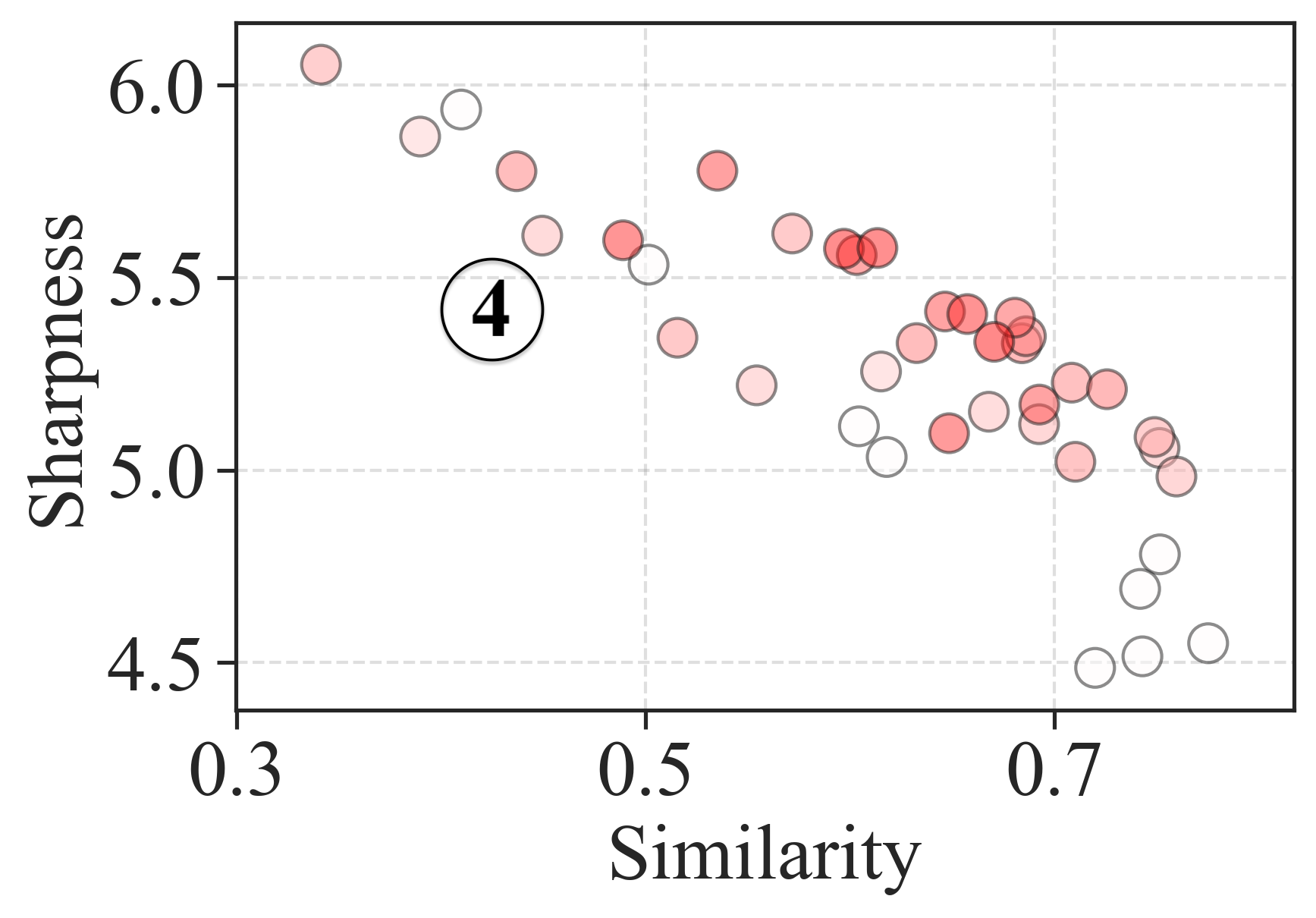}
        \caption{Training (w/o label noise)}~\label{fig:zero-shot-data-temp-cka-c}
    \end{subfigure} \\
    \begin{subfigure}{0.48\linewidth}
        \includegraphics[width=\linewidth]{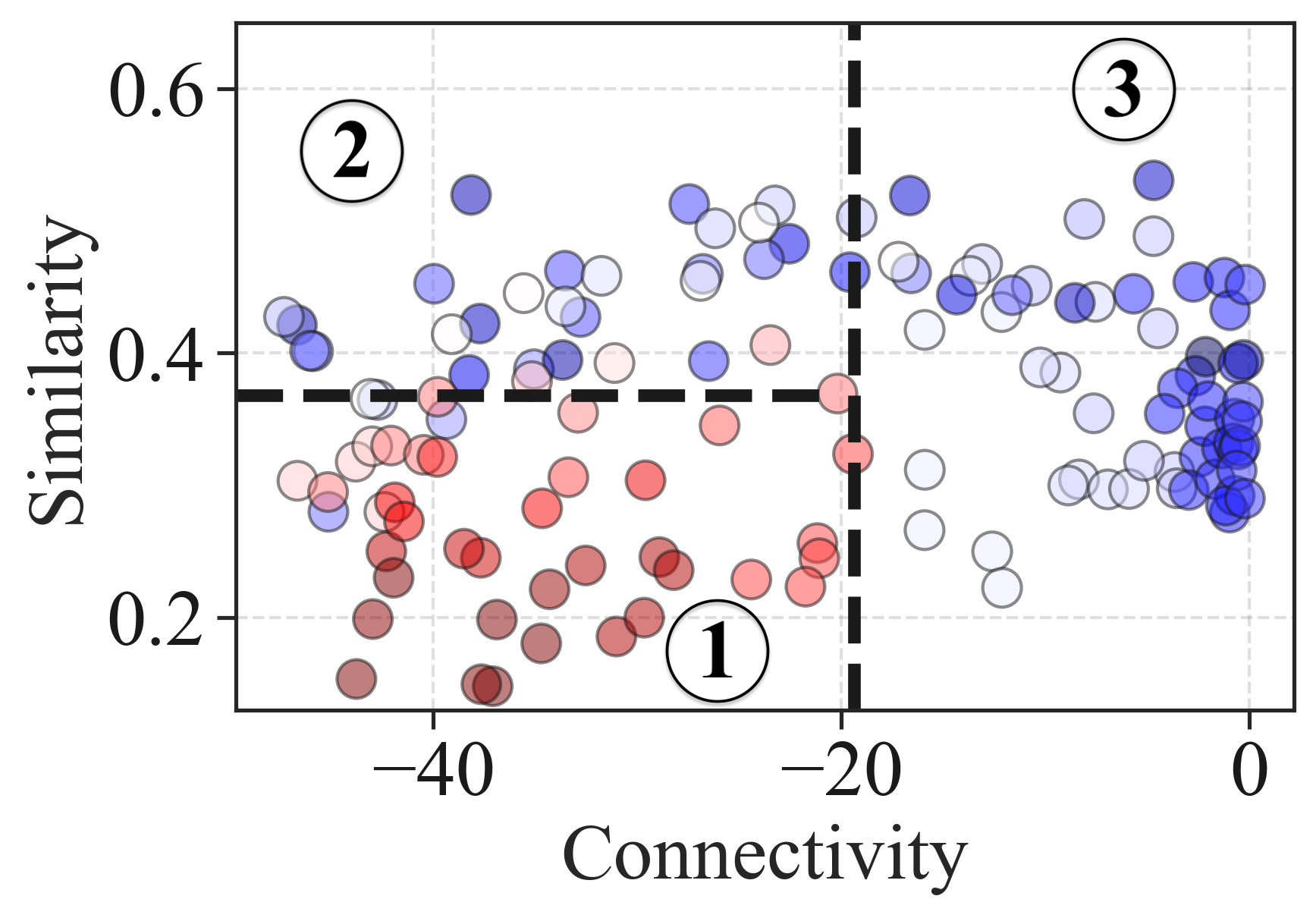}
        \caption{Test (w/ label noise)}~\label{fig:zero-shot-data-temp-cka-d}
    \end{subfigure} 
    \begin{subfigure}{0.48\linewidth}
        \includegraphics[width=\linewidth]{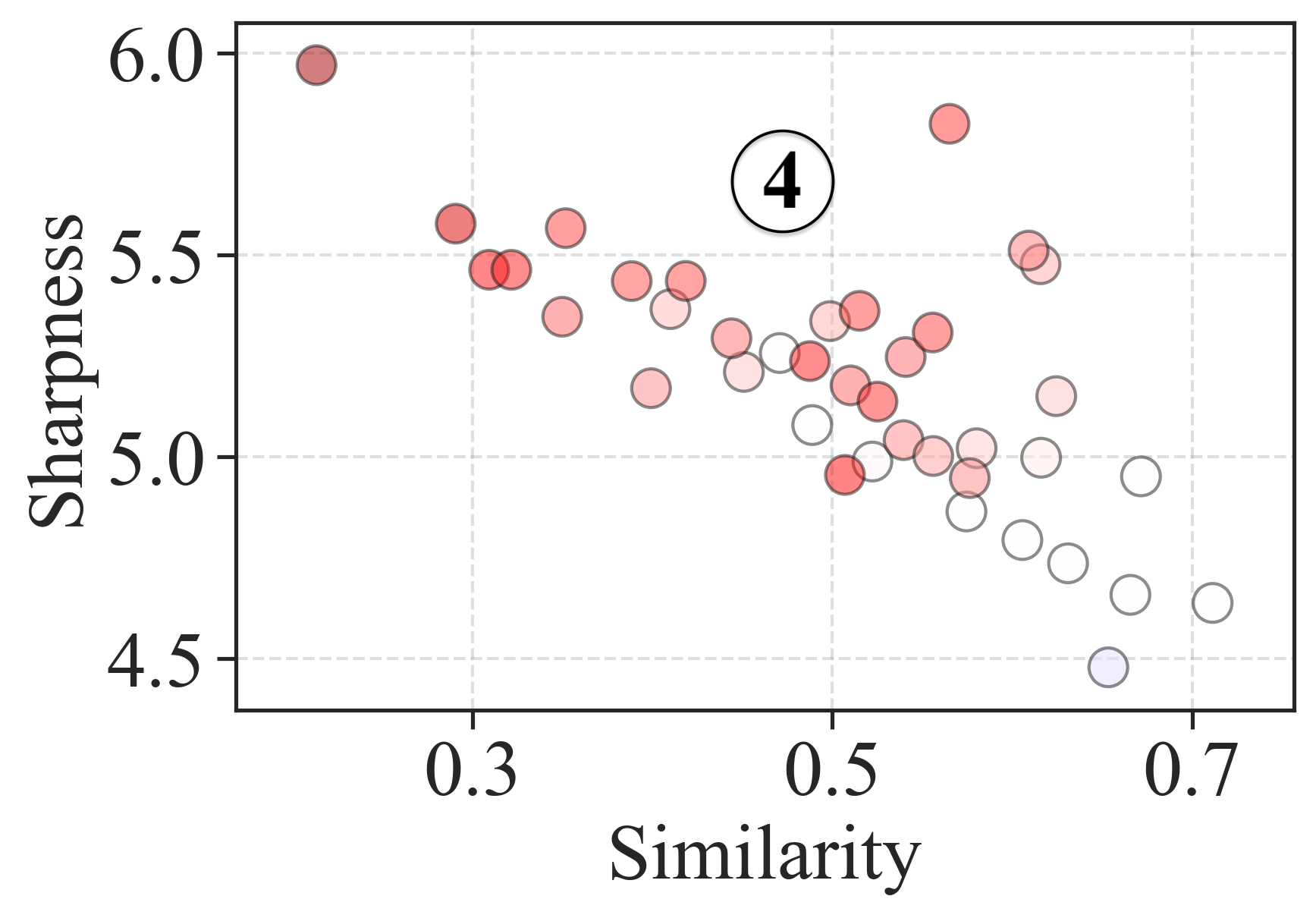}
        \caption{Test (w/ label noise)}~\label{fig:zero-shot-data-temp-cka-e}
    \end{subfigure} 
    \end{minipage} \vspace{-2mm}
    \caption{\textbf{(Ablation study of \ourmethod structure by using similarity instead of sharpness for Q1).}
    \emph{Left}: structure of the tree. The color of the leaf node indicates the predicted class.
    The threshold values are learned from the training set.
    \emph{Right}: 
    The first row represents training samples, and the second row represents test samples.
    Each colored circle represents one sample (which is one pre-trained model), and the color represents the ground-truth label: blue means the hyperparameter is too large, while red means small.
    The black dashed line indicates the decision boundary of \ourmethod.
    Each numbered regime on the right corresponds to the leaf node with the same number on the tree.
    The samples in~\ref{fig:zero-shot-data-temp-cka-b} and those in~\ref{fig:zero-shot-data-temp-cka-c} are separated by training error. The same applies to~\ref{fig:zero-shot-data-temp-cka-d} and~\ref{fig:zero-shot-data-temp-cka-e}.
    }~\label{fig:zero-shot-data-temp-cka}
\end{figure*}

\section{Ablation Study on Structure of \ourmethod}\label{app:abl-cka}

\begin{figure}[!h]
    \centering
    \begin{subfigure}{0.33\linewidth}
        \includegraphics[width=\linewidth]{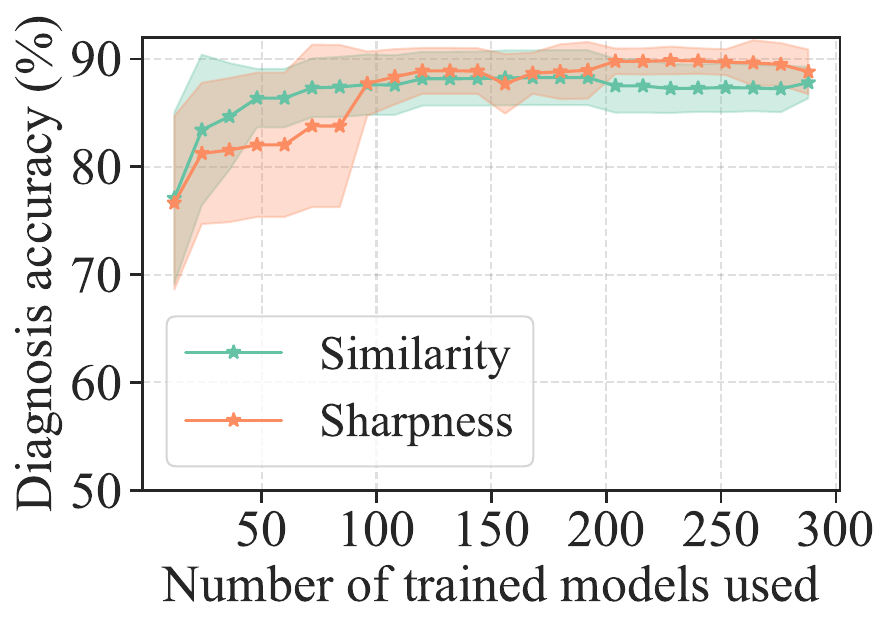} \hspace{1mm}
        \caption{Dataset transfer}~\label{fig:temp-a-abl-cka} 
    \end{subfigure} 
    \begin{subfigure}{0.30\linewidth}
        \includegraphics[width=\linewidth]{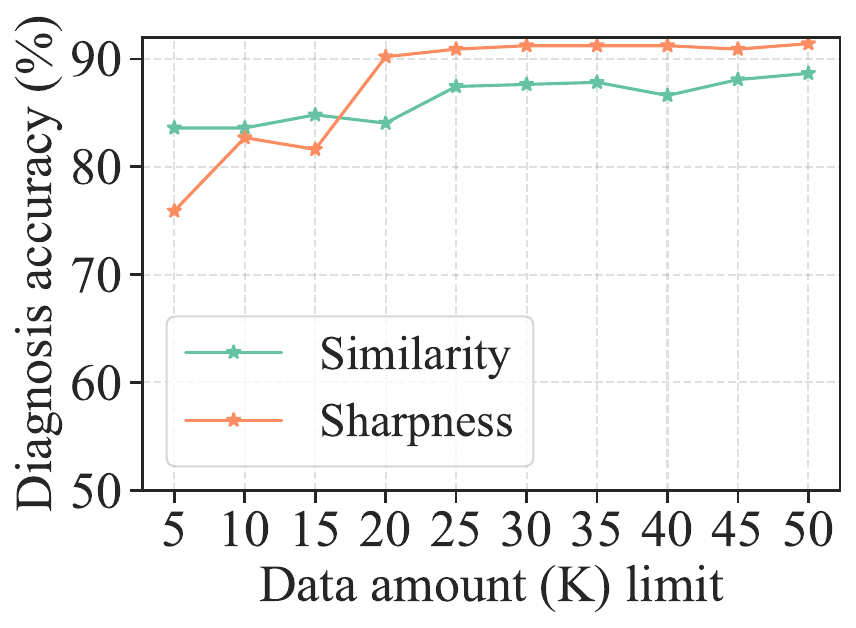}
        \caption{Scale transfer: limiting model's training data amount}~\label{fig:temp-b-abl-cka}
    \end{subfigure} 
    \begin{subfigure}{0.30\linewidth}
        \includegraphics[width=\linewidth]{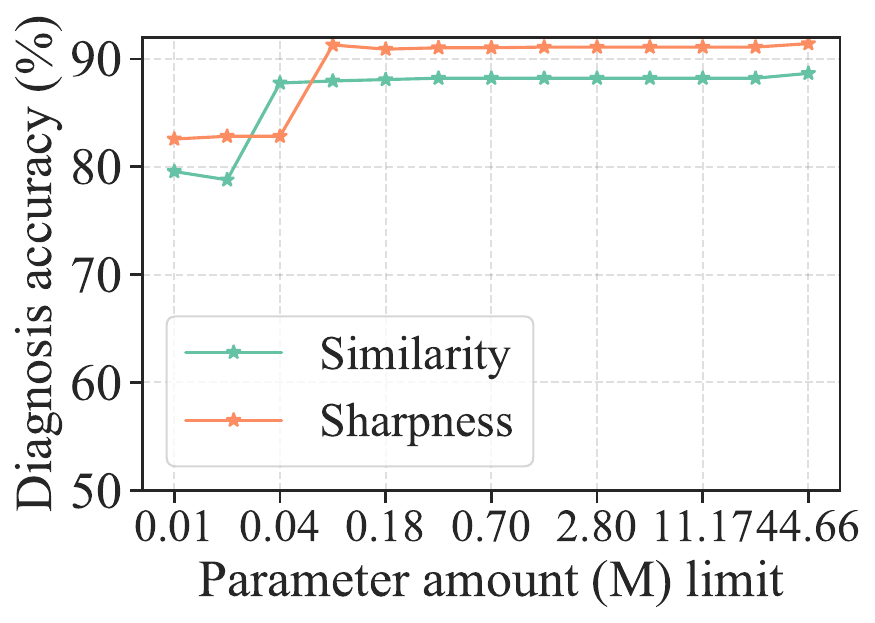}
        \caption{Scale transfer: limiting model's parameter amount}~\label{fig:temp-c-abl-cka}
    \end{subfigure} \vspace{-3mm}
    \caption{\textbf{(Varying \ourmethod structure by using similarity or sharpness in splitting the node in the deepest level for task Q1).} 
    $y$-axis indicates the diagnosis accuracy.
    (a) $x$-axis indicates the number of pre-trained models used for building the training set.
    (b) $x$-axis indicates the maximum amount of training (image) data for training models in the training set for fitting the classifier.
    (c) $x$-axis indicates the maximum number of parameters of the models in the training set.
    } ~\label{fig:temp-abl-cka} 
\end{figure}

In this section, we present an ablation study that uses a similarity metric for splitting nodes at deeper levels of the tree, as opposed to employing sharpness as outlined in the main paper, for diagnostic task Q1. 
We demonstrate that similarity, much like sharpness, is a useful metric for constructing the decision tree, offering an effective alternative for node splitting.

In the main paper, we mainly use training error $\mathcal{E}_\text{tr}$, connectivity $\mathcal{C}$, and sharpness $\mathcal{H}_t$ to construct the tree structure referred to as \ourmethod.
The tree structure to solve Q1 is shown in Figure~\ref{fig:zero-shot-temp-tree}. 
In this ablation study, we replace the sharpness metric at the leftmost deep internal node with the similarity metric $\mathcal{S}$.
The resulting tree structure is shown in Figure~\ref{fig:zero-shot-data-tree-cka}.
We follow the same experimental setup in Section~\ref{app:exp-setup}.
From Figure~\ref{fig:zero-shot-data-temp-cka-b} and Figure~\ref{fig:zero-shot-data-temp-cka-d}, we can see that the similarity metric separates the left space into regime~\textcircled{\raisebox{-0.9pt}{1}} and regime~\textcircled{\raisebox{-0.9pt}{2}}. In each regime, models have the same failure source.
This achieves a similar effect as sharpness in Figure~\ref{fig:temp-vis-a} and \ref{fig:temp-vis-c}. 
This means that the alternative structure of \ourmethod using similarity also provides precise and interpretable classification results.

In Figure~\ref{fig:temp-abl-cka}, we compare two tree structures: one using similarity and the other using sharpness, across dataset transfer and scale transfer for Q1.
Both structures achieve comparably high diagnosis accuracy in all three settings. 
The structure employing sharpness demonstrates marginally superior performance in scale transfer scenarios.
Based on these findings, we conclude that both sharpness and similarity are valuable metrics for node splitting at deeper levels of the tree, offering practical utility in constructing diagnostic tree structures.\looseness-1

\end{document}